\documentclass{article}

\usepackage[final]{neurips_2022}
\bibpunct[:]{(}{)}{;}{a}{,}{,}
\usepackage{subfiles}

\usepackage[utf8]{inputenc} % allow utf-8 input
\usepackage[T1]{fontenc}    % use 8-bit T1 fonts
\usepackage{url}            % simple URL typesetting
\usepackage{booktabs}       % professional-quality tables
\usepackage{amsfonts}       % blackboard math symbols
\usepackage{nicefrac}       % compact symbols for 1/2, etc.
\usepackage{microtype}      % microtypography
\usepackage{xcolor}         % colors
\usepackage{amsmath,amsthm,amssymb}
\usepackage{mathtools}
\usepackage[titlenumbered,linesnumbered,ruled]{algorithm2e}
\usepackage{dsfont}
\usepackage{hyperref}       % hyperlinks
\usepackage{xr}
\usepackage{wrapfig}
\newtheorem{theorem}{Theorem}
\newtheorem{corollary}{Corollary}[theorem]
\newtheorem{assumption}{Assumption}

\newtheorem{remark}{Remark}

\newtheorem{definition}{Definition}

\newtheorem{theoremA}{Theorem}[section]
\newtheorem{corollaryA}{Corollary}[theoremA]

\newtheorem{lemmaA}{Lemma}[section]
\newtheorem{remarkA}{Remark}[section]
\newtheorem{notationA}{Notation}[section]
\newtheorem{definitionA}{Definition}[section]
\newtheorem{propA}{Property}[section]

\newcommand{\e}{\mathrm{e}}
\renewcommand{\d}{\mathrm{d}}
\newcommand{\E}{\mathbb{E}}
\hypersetup{
    colorlinks,
    linkcolor={red!50!black},
    citecolor={blue!50!black},
    urlcolor={blue!80!black}
}

\title{Improved Convergence Rate of Stochastic Gradient Langevin Dynamics with Variance Reduction and its Application to Optimization}

\author{%
  Yuri Kinoshita\\
  Department of Mathematical Informatics,\\
  Graduate School of Information Science and Technology,\\
   The University of Tokyo, Tokyo, Japan\\
  \texttt{yuri-kinoshita111@g.ecc.u-tokyo.ac.jp} \\
   \And
  Taiji Suzuki \\
  Department of Mathematical Informatics,\\
 Graduate School of Information Science and Technology,\\
  The University of Tokyo, Tokyo, Japan\\
 Center for Advanced Intelligence Project, RIKEN, Tokyo, Japan\\
 \texttt{taiji@mist.i.u-tokyo.ac.jp} \\
}

\begin{document}

\maketitle

\begin{abstract}
  The stochastic gradient Langevin Dynamics is one of the most fundamental algorithms to solve sampling problems and non-convex optimization appearing in several machine learning applications. Especially, its variance reduced versions have nowadays gained particular attention. In this paper, we study two variants of this kind, namely, the Stochastic Variance Reduced Gradient Langevin Dynamics and the Stochastic Recursive Gradient Langevin Dynamics. We prove their convergence to the objective distribution in terms of KL-divergence under the sole assumptions of smoothness and Log-Sobolev inequality which are weaker conditions than those used in prior works for these algorithms. With the batch size and the inner loop length set to $\sqrt{n}$, the gradient complexity to achieve an $\epsilon$-precision is $\tilde{O}((n+dn^{1/2}\epsilon^{-1})\gamma^2 L^2\alpha^{-2})$, which is an improvement from any previous analyses. We also show some essential applications of our result to non-convex optimization.
\end{abstract}

\section{Introduction}
\label{c1}

\subsection{Background and Organization}
Over the past decade, the gradient Langevin Dynamics (GLD) has gained particular attention for providing an effective tool for sampling from a Gibbs distribution, a fundamental task omnipresent in the field of machine learning and statistics, and for non-convex optimization, which is nowadays witnessing an unignorable empirical success. Notably, GLD is a stochastic differential equation (SDE) that can be viewed as the steepest descent flow of the Kullback-Leibler (KL) divergence towards the stationary Gibbs distribution in the space of measures endowed with the 2-Wasserstein metric \citep{JKO1998}. As a consequence of the unique properties of GLD, its implementable discrete schemes and their ability to suitably track it have been the subject of a large number of studies.\par
The Euler-Maruyama scheme of GLD gives rise to an algorithm known as the Langevin Monte Carlo method (LMC). This algorithm is biased \citep{W2018}: that is, the distribution of the discrete scheme does not converge to the same as GLD. Nonetheless, it has been shown that this bias could be made arbitrarily small under certain assumptions by taking a sufficiently small step size \citep{D2017a,VW2019}. \citet{D2017b,D2017a} provided one of the first non-asymptotic rates of convergence of LMC for smooth log-concave distributions. Assumptions to obtain a non-asymptotic analysis and this controllable bias have been relaxed by further research to dissipativity and smoothness \citep{RRT2017,XCZG2017}, and recently to Log-Sobolev inequality (LSI) and smoothness \citep{VW2019}. This relaxation of conditions is especially meaningful as the objective distribution nowadays tends to become more and more complicated beyond the classical assumption of log-concavity. \par
However, in the field of machine learning, the main function can often be formulated as the average of the loss function of an enormous number of training data points \citep{WT2011}, which subsequently makes it difficult to calculate its full gradient. As a result, research on stochastic algorithms has been also conducted to avoid this computational burden \citep{CLX2021,DJWPSX2016,RRT2017,WT2011,XCZG2017,ZXG2018,ZXG2019a,ZXG2019b,ZXG2020}. \citet{WT2011} introduced the concept of Stochastic Gradient Langevin Dynamics (SGLD) which combines the Stochastic Gradient Descent with LMC. This has been the subject of successful studies \citep{RRT2017,WT2011,XCZG2017}. Nevertheless, the variance of its stochastic gradient is too large, which leads to a slow convergence compared to LMC. Therefore, stochastic gradient Langevin Dynamics algorithms with variance reduction, such as the Stochastic Variance Reduced Gradient Langevin Dynamics (SVRG-LD), have been considered and their convergence has been thoroughly analyzed for both sampling \citep{DJWPSX2016,ZXG2018,ZXG2019a,ZXG2020} and optimization \citep{HB2021,XCZG2017}. \par
\citet{DJWPSX2016} first united SGLD with variance reduction techniques and proposed two new algorithms, namely, SVRG-LD and SAGA-LD. \citet{CFMBJ2018} and \citet{ZXG2018} proved the convergence rate of SVRG-LD to the target distribution in 2-Wasserstein distance for smooth log-concave distributions. \citet{XCZG2017} showed the weak convergence of SVRG-LD under the smoothness and dissipativity conditions. They expanded the non-asymptotic analysis of \citet{RRT2017} to LMC and SVRG-LD, and improved the result for SGLD. Few years ago, \citet{ZXG2019a} provided the gradient complexity of SVRG-LD to converge to the stationary distribution in 2-Wasserstein distance under the smoothness and dissipativity assumptions. This convergence can be even improved if we make a warm-start \citep{ZXG2020}. While these works investigated algorithms with fixed hyperparameters, \citet{HB2021} additionally assumed a strict saddle and some other minor conditions to study SVRG-LD with a decreasing step size and improved its convergence in high probability to the second order stationary point. \citet{ZXG2019b} also applied variance reduction techniques to the Hamiltonian Langevin Dynamics, or underdamped Langevin Dynamics in opposition to GLD also known as overdamped Langevin Dynamics. As we can observe, the current convergence analyses of the stochastic schemes with variance reduction are mostly restricted to log-concavity and dissipativity, and do not enjoy the same broad convergence guarantee with a concrete gradient complexity as LMC does under LSI and smoothness in terms of KL-divergence.\par
Therefore, in order to bridge this theoretical gap between LMC and  stochastic gradient Langevin Dynamics with variance reduction, we study in this paper the convergence of the latter under the relaxed assumptions of smoothness and LSI. In Section \ref{c3}, we study the convergence to the Gibbs distribution of SVRG-LD and the Stochastic Recursive Gradient Langevin Dynamics (SARAH-LD), another variant of stochastic gradient Langevin Dynamics with variance reduction inspired by the Stochastic Recursive Gradient algorithm (SARAH) of \citet{NLST2017a,NLST2017b}. On the other hand, optimization and sampling are only two sides of the same coin for GLD. That is why, in Section \ref{c4}, we also investigate implications of Section \ref{c3} for non-convex optimization. We prove the convergence of SVRG-LD and SARAH-LD to the global minimum of dissipative functions and we provide their non-asymptotic rate of convergence. We also consider the additional weak Morse assumption and study its effect. Finally, we illustrate our main result with a simple experiment.
\subsection{Contributions}
The major contributions of this paper can be summarized as follows.
\begin{table}
  \centering
  \caption{Comparison of our main result with prior works (sampling). The first three works are about LMC. Compared to Vempala et al. \citeyearpar{VW2019}, with the same assumptions and criterion, the order of gradient complexity is improved from $n$ to $\sqrt{n}$. The others are about SVRG-LD except the last one which is about the Stochastic Gradient Hamiltonian Monte Carlo Methods with Recursive Variance Reduction. $\epsilon$ is the accuracy required on the criterion, $d$ is the dimension of the input of the main function, $n$ is the number of data points, and $L$ is the smoothness constant. ${}^\ast$ 2-Wass. stands for ``2-Wasserstein'', and conv. stands for ``convergence''. ${}^{\ast\ast}$ $\mathrm{poly}(M,L)$ stands for a polynomial of $M$ and $L$.}
  \label{tab1}
  \begin{tabular}{cccccc}
    \toprule
    \footnotesize{Method} & \footnotesize{Major Assumptions} & \footnotesize{Criterion${}^\ast$} & \footnotesize{Gradient Complexity${}^{\ast\ast}$} \\ \midrule
    \footnotesize{Dalalyan \citeyearpar{D2017b}} & \footnotesize{Smooth, Log-concave ($M$)}& \footnotesize{2-Wass.} & $\scriptstyle{\tilde{O}\left(\frac{nd}{\epsilon^2}\cdot \mathrm{poly}(M,L)\right)}$\\[0.2cm]
    \footnotesize{Xu et al. \citeyearpar{XCZG2017}} & \footnotesize{Smooth, Dissipative}& \footnotesize{Weak conv.} & $\scriptstyle{\tilde{O}\left(\frac{nd}{\epsilon}\right)\cdot \e^{\tilde{O}(d)}}$\\[0.2cm]
    \footnotesize{Vempala et al. \citeyearpar{VW2019}} & \footnotesize{Smooth, Log-Sobolev ($\alpha$)}& \footnotesize{KL} &  $\scriptstyle{\tilde{O}\left(\frac{n}{\epsilon}\cdot d\gamma^2 L^2\alpha^{-2}\right)}$ \\[0.2cm]
    \footnotesize{Zou et al. \citeyearpar{ZXG2018}} & \footnotesize{Smooth, Log-concave ($M$)}& \footnotesize{2-Wass.} & $\scriptstyle{\tilde{O}\left(n+\frac{L^{3/2}n^{1/2}d^{1/2}}{M^{3/2}\epsilon}\right)}$\\[0.2cm]
    \footnotesize{Zou et al. \citeyearpar{ZXG2019a}} & \footnotesize{Smooth, Dissipative}& \footnotesize{2-Wass.} & $\scriptstyle{\tilde{O}\left(n+\frac{n^{3/4}}{\epsilon^2}+\frac{n^{1/2}}{\epsilon^4}\right)\cdot \e^{\tilde{O}(\gamma+d)}}$\\[0.2cm]
    \footnotesize{Zou et al. \citeyearpar{ZXG2020}} & \footnotesize{Smooth, Dissipative, Warm-start}& \footnotesize{TV} &      $\scriptstyle{\tilde{O}\left(\frac{\gamma^2}{\epsilon^2}\right)\cdot \e^{\tilde{O}(d)}}$\\[0.2cm]
    \footnotesize{Zou et al. \citeyearpar{ZXG2019b}} & \footnotesize{Smooth, Dissipative}& \footnotesize{2-Wass.} & $\scriptstyle{\tilde{O}\left((n+\frac{n^{1/2}}{\epsilon^{2}\mu_\ast^{3/2}})\wedge\frac{\mu_\ast^{-2}}{\epsilon^{4}}\right)}$\\[0.2cm]
    \footnotesize{\textbf{This paper}} & \footnotesize{Smooth, Log-Sobolev ($\alpha$)}& \footnotesize{KL} &      $\scriptstyle{\tilde{O}\left(\left(n+\frac{dn^{1/2}}{\epsilon}\right)\cdot \gamma^2 L^2\alpha^{-2}\right)}$ \\
    \bottomrule
  \end{tabular}
\end{table}
We provide a non-asymptotic analysis of the convergence of SVRG-LD and SARAH-LD to the Gibbs distribution in terms of KL-divergence under smoothness and LSI which are weaker conditions than those used in prior works for these algorithms. KL-divergence is generally a stronger convergence criterion than both total variation (TV) and 2-Wasserstein distance as they can be controlled by KL-divergence under the LSI condition. Notably, we prove that, with the batch size and inner loop length set to $\sqrt{n}$, the gradient complexity to achieve an $\epsilon$-precision in terms of KL-divergence is $\tilde{O}((n+dn^{1/2}\epsilon^{-1})\gamma^2 L^2\alpha^{-2})$, which is better than any previous analyses. See Table \ref{tab1} for a comparison with previous research in terms of assumptions, criterion and gradient complexity. We also prove the convergence of SVRG-LD and SARAH-LD to the global minimum under an additional assumption of dissipativity with a gradient complexity of $\tilde{O}((n+n^{1/2}\epsilon^{-1}d L\alpha^{-1})\gamma^2 L^2\alpha^{-2})$ which is better than previous work since it has almost all the time a dependence on $n$ of $O(\sqrt{n})$ and does not require the batch size and the inner loop length to depend on the accuracy $\epsilon$. On the other hand, we import the idea of \citet{LE2020} from product manifolds of spheres to the Euclidean space in order to show that under the additional assumption of weak Morse, the convergence in the Euclidean space can be accelerated by eliminating the exponential dependence on $1/\epsilon$.
\subsection{Other Related Works}
The theoretical study of GLD goes back to \citet{CHS1987} who showed that global convergence could be achieved with a proper annealing schedule. This work did not specify how to implement this SDE, but \citet{GM1991} filled this gap. Later, \citet{BM1999} proved an asymptotic convergence in terms of relative entropy for the discrete scheme of gradient Langevin Dynamics when the inverse temperature and the step size are kept constant. \par
The variance reduction technique, introduced to Langevin Dynamics by \citet{DJWPSX2016}, was originally presented by \citet{JZ2013} as Stochastic Variance Reduced Gradient (SVRG) to improve the convergence speed of Stochastic Gradient Descent. Other variance reduction techniques were also considered such as the Stochastic Recursive Gradient Langevin Dynamics (SARAH) from \citet{NLST2017a,NLST2017b} which outperforms SVRG in non-convex optimization \citep{PNPT2020} and is used in many algorithms such as SSRGD \citep{L2019} and SpiderBoost \citep{WJZLT2019}. \par
\citet{LE2020} extended Vempala and Wibisono's result to Riemannian manifolds. One of the highlights of their work is that they showed the Log-Sobolev constant of the Gibbs distribution for a product manifold of spheres only depends on a polynomial of the inverse temperature under some particular conditions including weak Morse. We will adapt this result to our situation.
\par In the concurrent work of \citet{BCESZ2022} (especially Section 6), they also studied the convergence of stochastic schemes of GLD with more relaxed conditions than prior analyses. However, our contributions are not overshadowed by theirs, and we clarify the reasons. In Subsection 6.1 of their paper, \citet{BCESZ2022} focused on stochastic discrete schemes with finite variance and bias (which is not the case for SVRG-LD) and provided a first-order convergence guarantee in the space of measures equipped with the 2-Wasserstein distance. Subsection 6.2 proved a global convergence under some other conditions but most of these two analyses did not consider in particular the usual case in machine learning when $F$ is the average of some other functions, which leads to a generally worse gradient complexity than ours. Concerning this finite sum setting, \citet{BCESZ2022} investigated the Variance Reduced LMC algorithm (slightly different from SVRG-LD in this paper) in Subsection 6.3 and gave a first-order convergence under the sole assumption of smoothness. When restrained in our problem setting, the gradient complexity of SVRG-LD and SARAH-LD we provide is still considerably better (see Section \ref{c3} for more details). %In some extent, our work can be interpreted as an acceleration of their result under some particular conditions.
\subsection{Notation}
We denote deterministic vectors by a lower case symbol (e.g., $x$) and random variables by an upper case symbol (e.g., $X$). The Euclidean norm is denoted by $\|\cdot\|$ for vectors and the inner product by $\langle\cdot,\cdot\rangle$. For matrices, $\|\cdot\|$ is the norm induced by the Euclidean norm for vectors. We only treat distributions absolutely continuous with respect to the Lebesgue measure in $\mathbb{R}^d$ for simplicity. Especially, throughout the paper, $\nu$ refers to the probability measure with the density function $\d \nu\propto\e^{-\gamma F}\d x$, where $F$ is a function introduced below. $a\vee b$ is equivalent to $\max\{a,b\}$ and $a\wedge b$ to $\min\{a,b\}$. We also use the shorthand $\tilde{O}$ to hide logarithmic polynomials.

\section{Preliminaries}
\label{c2}

In this section, we briefly explain the problem setting, necessary mathematical background and assumptions used in this paper.
\subsection{Problem Setting and GLD}
In Section \ref{c3}, we consider sampling from a distribution written in the form $\d\nu\propto \e^{-\gamma F}\d x$ where $\gamma$ is a positive constant (which corresponds to the inverse temperature) and $F:\mathbb{R}^d\to \mathbb{R}$ is formulated as $F(x)\vcentcolon =\frac{1}{n}\sum_{i=1}^n f_i(x)$, the average of the loss function of $n$ training data points $\{x^{(i)}\}_{i=1}^n$. Here, $f_i(x)\vcentcolon=f(x,x^{(i)})$ can be regarded as the loss of data $x^{(i)}$. For instance, $F$ can be the average of the negative log likelihood of $n$ training data points. In Section \ref{c4}, we consider the non-convex optimization (minimization) of the same $F$ as above.
\par GLD can be described as the following stochastic differential equation (SDE):
\begin{align}
  \d X_t^{\footnotesize{\mathrm{GLD}}}=-\nabla F(X_t^{\footnotesize{\mathrm{GLD}}})\d t+\sqrt{2/\gamma}\d  B(t),\label{eq21}
\end{align}
where $\gamma>0$ is called the inverse temperature parameter and $\{B(t)\}_{t\ge 0}$ is the standard Brownian motion in $\mathbb{R}^d$. It can be used for sampling since under some reasonable assumptions of $F$, the distribution $\rho_t^{\footnotesize{\mathrm{GLD}}}$ of $X_t^{\footnotesize{\mathrm{GLD}}}$ governed by SDE \eqref{eq21} converges to the invariant stationary distribution $\d\nu\propto \e^{-\gamma F}\d x$, also known as the Gibbs distribution \citep{CHS1987}. Moreover, as previously mentioned, this convergence is efficient in the sense that SDE \eqref{eq21} corresponds to the steepest descent flow of the Kullback-Leibler (KL) divergence towards the stationary distribution in the space of measures endowed with the 2-Wasserstein metric \citep{JKO1998}. Alternatively, GLD can be interpreted as the composite optimization problem of a negative entropy and an expected function value as follows \citep{W2018}:
\[
\min_{q:\mathrm{density}}\E_q[\gamma F] + \E_q [\log q].
\]
The gradient flow is the well-known Fokker-Planck equation associated to SDE \eqref{eq21}:
\begin{align}
  \frac{\partial \rho_t^{\footnotesize{\mathrm{GLD}}}}{\partial t}&=\nabla\cdot(\rho_t^{\footnotesize{\mathrm{GLD}}}\nabla F)+\frac{1}{\gamma}\Delta \rho_t^{\footnotesize{\mathrm{GLD}}}=\frac{1}{\gamma}\nabla\cdot\left(\rho_t^{\footnotesize{\mathrm{GLD}}}\nabla\log{\frac{\rho_t^{\footnotesize{\mathrm{GLD}}}}{\nu}}\right).\label{eq22}
\end{align}
This will be useful in our analysis. In addition to its potential for sampling, GLD can also be employed for non-convex optimization as the Gibbs distribution concentrates on the global minimum of $F$ for sufficiently large values of $\gamma$ \citep{H1980}.
\subsection{Algorithms of GLD}
\begin{algorithm}
  \hypertarget{al1}{}
  \hypertarget{al2}{}
  input: step size $\eta>0$, batch size $B$, epoch length $m$, inverse temperature $\gamma\ge 1$\\
  initialization: $X_0=0$, $ X^{(0)}=X_0$\\
  \ForEach{$s=0,1,\ldots,(K/m)$}
  {
  $v_{sm}=\nabla F(X^{(s)})$\\
  randomly draw $\epsilon_{sm} \sim N(0,I_{d\times d})$\\
  $X_{sm+1}=X_{sm}-\eta v_{sm}+\sqrt{2\eta/\gamma}\epsilon_{sm}$\\
  \ForEach{$l=1,\ldots,m-1$}
  {
  $k=sm+l$\\
  randomly pick a subset $I_k$ from $\{1,\ldots,n\}$ of size $|I_k|=B$\\
  randomly draw $\epsilon_k \sim N(0,I_{d\times d})$\\
  \uIf{SVRG-LD}{
  $v_k=\frac{1}{B}\sum_{i_k\in I_k}(\nabla f_{i_k}(X_k)-\nabla f_{i_k}(X^{(s)}))+v_{sm}$
  }
  \ElseIf{SARAH-LD}{
  $v_k=\frac{1}{B}\sum_{i_k\in I_k}\left(\nabla f_{i_k}(X_k)-\nabla f_{i_k}(X_{k-1})\right)+v_{k-1}$
  }
  $X_{k+1}=X_k-\eta v_k+\sqrt{2\eta/\gamma}\epsilon_k$\\
  }
  $X^{(s+1)}=X_{(s+1)m}$\\
  }
  \caption{SVRG-LD / SARAH-LD}
\end{algorithm}
Applying the Euler-Maruyama scheme to \eqref{eq21}, we obtain the Langevin Monte Carlo (LMC)
\begin{align*}
    X_{k+1}&=X_k-\eta \nabla F(X_k)+\sqrt{2\eta/\gamma}\epsilon_k,
\end{align*}
where $\eta$ is called the step size. This is similar to the gradient descent except the additional Gaussian noise $\sqrt{2\eta/\gamma}\epsilon_k$, where $\epsilon_k\sim N(0,I_{d\times d})$ and $I_{d\times d}$ is the $d\times d$ unit matrix. In the case $n$ is huge and the computation of $\nabla F$ is too difficult, we are incited to use stochastic gradient methods in analogy to stochastic gradient optimization. This gives
\begin{align*}
    X_{k+1}&=X_k-\eta v(X_k)+\sqrt{2\eta/\gamma}\epsilon_k,
\end{align*}
where $v(X_k)$ is the stochastic gradient. When $v(X_k)$ is defined as $\frac{1}{B}\sum_{i_k\in I_k}\nabla f_{i_k}(X_k)$, where $B$ is called the batch size and $I_k$ is a random subset uniformly chosen from $\{1,\ldots,n\}$ such that $|I_k|=B$, we obtain the Stochastic Gradient Langevin Dynamics (SGLD). As this method exhibits a slow convergence, it has been popular to use variance reduction methods such as the Stochastic Variance Reduced Gradient Langevin Dynamics (SVRG-LD) where $v(X_k)=\frac{1}{B}\sum_{i_k\in I_k}(\nabla f_{i_k}(X_k)-\nabla f_{i_k}(X^{(s)}))+\nabla F(X^{(s)})$. Details of this algorithm is stated in Algorithm \hyperlink{al1}{1}. $X^{(s)}$ is a reference point updated every $m$ steps so that $X_{sm}=X^{(s)}$. As we can observe in Lemma \ref{l4}, around the optimal point, the variance of the stochastic gradient is indeed decreased as $X^{(s)}$ and $X_k$ are both close to each other. We can also easily extend some successful stochastic gradient algorithms to Langevin Dynamics. Hence, we are motivated to extend the Stochastic Recursive Gradient Algorithm (SARAH) to Langevin Dynamics since we can expect that some bottlenecks of the analysis of SVRG-LD can be removed in that of SARAH-LD as subtracting the previous stochastic gradient enables a stabler performance than SVRG-LD. This algorithm can be described as Algorithm \hyperlink{al2}{1} with $v(X_k)=\frac{1}{B}\sum_{i_k\in I_k}\left(\nabla f_{i_k}(X_k)-\nabla f_{i_k}(X_{k-1})\right)+v(X_{k-1})$.
\begin{definition}
  We define $\rho_k$ as the distribution of $X_k$ generated at the $k$th step of SVRG-LD, and similarly $\phi_k$ for SARAH-LD.
\end{definition}
\subsection{Assumptions}\label{ss23}
The assumptions used throughout this paper can be summarized as follows.\par
\begin{assumption}\label{as1}
For all $i=1,\ldots,n$, $\nabla f_i$ is twice differentiable, and $\forall x,y\in \mathbb{R}^d$, $\|\nabla^2 f_i(x)\|\le L$. In other words, $f_i$ ($i=1,\ldots,n$) and $F$ are $L$-smooth.
\end{assumption}
\begin{assumption}\label{as2}
Distribution $\nu$ satisfies the Log-Sobolev inequality (LSI) with a constant $\alpha$. That is, for all probability density functions $\rho$ absolutely continuous with respect to $\nu$, the following holds:
\begin{align*}
  H_{\nu}(\rho)\le \frac{1}{2\alpha}J_{\nu}(\rho),
\end{align*}
where $H_{\nu}(\rho)\vcentcolon=\E_{\rho}\left[ \log{\frac{\rho}{\nu}}\right]$ is the KL-divergence of $\rho$ with respect to $\nu$, and $J_{\nu}(\rho)\vcentcolon=\E_{\rho}\left[ \left\|\nabla\log{\frac{\rho}{\nu}}\right\|^2\right]$ is the relative Fisher information of $\rho$ with respect to $\nu$.
\end{assumption}
The recent work of \citet{VW2019} motivates us to use the combination of smoothness and LSI for the analysis of SVRG-LD and SARAH-LD. Indeed, they showed that these conditions were enough to assure for the Euler-Maruyama scheme an exponentially fast convergence and a bias controllable by the step size. Under smoothness, LSI is not only the necessary condition of log-concavity and dissipativity, but is also robust to bounded perturbation and Lipschitz mapping, contrary to log-concavity \citep{VW2019}. For example, for any distribution $\d\nu$ that satisfies LSI and bounded function $B:\mathbb{R}^d\to\mathbb{R}$, $\d\tilde{\nu}\propto \e^{B}\d\nu$ satisfies LSI as well \citep{HS1986}. Moreover, while KL-divergence is not in general convex with regard to the Wasserstein geodesic, thanks to LSI, the Polyak-Łojaciewicz condition is satisfied. It is well-known that LSI suffices to realize an exponential convergence for the case of continuous time Langevin Dynamics \citep{VW2019}. That is why, it is actually both useful and natural to suppose LSI in this context. Note that under $L$-smoothness of $F$ and LSI with constant $\alpha$ for $\d\nu\propto \e^{-\gamma F}\d x$, it holds that $\alpha\le \gamma L$ \citep{VW2019}.
\par As for optimization, we additionally use the following conditions.
\begin{assumption}\label{as3}
  $F$ is $(M,b)$-dissipative. That is, there exist constants $M>0$ and $b>0$ such that for all $x\in\mathbb{R}^d$ the following holds: $\left\langle \nabla F(x),x\right\rangle\ge M\|x\|^2-b$.
\end{assumption}
\begin{assumption}[\citet{LE2020}, Assumption 3.3]\label{as5} $F$ satisfies the weak Morse condition. That is, for all non-zero eigenvalues of the Hessian of stationary points, there exists a constant $\lambda^\dagger\in (0,1]$ such that
 \[
\lambda^\dagger\le \inf\left\{\left|\lambda_i\left(\nabla^2F(x)\right)\right|\mid \nabla F(x)=0,\ i\in {1,\ldots,d},\ \lambda_i\left(\nabla^2F(x)\right)\ne 0\right\}.
 \]
 Furthermore, for the set $\mathcal{S}$ of stationary points that are not a global minimum, $\sup_{x\in\mathcal{S}}\lambda_{\mathrm{min}}\left(\nabla^2F(x)\right)\le -\lambda^\dagger$.
\end{assumption}
\begin{assumption}\label{as6}
  $\nabla^2 f_i$ is $L'$-Lipschitz and without loss of generality, we let $\min_{x\in\mathbb{R}^d} F(x)=0$.
\end{assumption}
\begin{assumption}\label{as7}
  $F$ has a unique global minimum.
\end{assumption}
\par Smoothness and dissipativity are a classical combination of assumptions for this kind of problem setting \citep{RRT2017,XCZG2017,ZXG2019a}. We assume dissipativity instead of LSI for non-convex optimization in order to obtain an explicit value of the Log-Sobolev constant of $\d\nu\propto\e^{-\gamma F}\d x$ in function of the inverse temperature parameter $\gamma$ (see Property \ref{p1}), making a non-asymptotic analysis possible. Furthermore, Assumptions \ref{as5} to \ref{as7} can ameliorate the exponential dependence of the inverse of the Log-Sobolev constant on the inverse temperature parameter to a polynomial one (see Property \ref{p2}).

\section{Main Results}
\label{c3}

In this section, we state our main results which prove that SVRG-LD and SARAH-LD (Algorithm \hyperlink{al2}{1}) achieve an exponentially fast convergence to the Gibbs distribution and a controllable bias in terms of KL-divergence under the sole assumptions of LSI and smoothness. We provide their gradient complexity as well. The proofs can be found in Appendix \ref{apa} and \ref{apb} respectively.
\subsection{Improved Convergence of SVRG-LD}
Our analysis shows that the convergence of SVRG-LD to the stationary distribution $\d\nu\propto\e^{-\gamma F}\d x$ can be formulated as the theorem below.
\begin{theorem}\label{mth1}
  Under Assumptions \ref{as1} and \ref{as2}, $0<\eta<\frac{\alpha}{16\sqrt{6}L^2m\gamma}$, $\gamma\ge 1$ and $B\ge m$, for all $k=1,2,\ldots$, the following holds in the update of SVRG-LD where $\Xi =\frac{(n-B)}{B(n-1)}$ :
  \begin{align*}
    H_{\nu}(\rho_{k})&\le \e^{-\frac{\alpha\eta}{\gamma}k}H_\nu(\rho_0)+\frac{224\eta\gamma dL^2}{3\alpha}\left(2+3\Xi +2m\Xi \right).
  \end{align*}
\end{theorem}
We observe that the bias term of the upper bound, which is the second term linearly dependent on $\eta$, can be easily controlled while the first term exponentially converges to 0 with $k\to \infty$. This is more precisely formulated in the following corollary.
\begin{corollary}\label{mcor11}
  Under the same assumptions as Theorem \ref{mth1}, for all $\epsilon\ge 0$, if we choose step size $\eta$ such that $\eta\le \frac{3\alpha\epsilon}{448\gamma dL^2}$, then a precision $H_{\nu}(\rho_{k})\le\epsilon$ is reached after $k\ge \frac{\gamma}{\alpha\eta}\log{\frac{2H_{\nu}(\rho_0)}{\epsilon}}$ steps. Especially, if we take $B=m=\sqrt{n}$ and the largest permissible step size $\eta=\frac{\alpha}{16\sqrt{6}L^2\sqrt{n}\gamma}\wedge \frac{3\alpha\epsilon}{448dL^2\gamma}$, then the gradient complexity becomes
  \[
  \tilde{O}\left(\left(n+\frac{dn^{\frac{1}{2}}}{\epsilon}\right)\cdot \frac{\gamma^2 L^2}{\alpha^2}\right).
  \]
\end{corollary}
\par This gradient complexity is an improvement compared with prior works for three reasons. First of all, we provide a non-asymptotic analysis of the convergence of SVRG-LD under smoothness and Log-Sobolev inequality which are conditions weaker than those (e.g., log-concavity or dissipativity) used in prior works for these algorithms. Moreover, we prove it in terms of KL-divergence which is generally a stronger convergence criterion than both total variation (TV) and 2-Wasserstein distance as they can both be controlled by KL-divergence under the LSI condition. For instance, TV was used by \citet{ZXG2020} and 2-Wasserstein distance by \citet{D2017b} and \citet{ZXG2019a}. KL-divergence makes it possible to unify these two different criteria. Finally, while prior research generally used Girsanov's theorem which generates a bias term that accumulates through the iteration (see for example \citet{RRT2017} and \citet{XCZG2017}), we solve this issue by taking benefit of the exponential convergence of GLD to the Gibbs distribution under LSI and smoothness that enables us to forget about past bias. That way, with the batch size and inner loop set to $\sqrt{n}$, the gradient complexity to achieve an $\epsilon$-precision in terms of KL-divergence becomes $\tilde{O}((n+dn^{1/2}\epsilon^{-1})\gamma^2 L^2\alpha^{-2})$, which is better than previous analyses. For example, \citet{VW2019} provided a gradient complexity of $\tilde{O}\left(n\epsilon^{-1}d\gamma^2 L^2\alpha^{-2}\right)$ for LMC under Assumptions \ref{as1} and \ref{as2}, and \citet{ZXG2019a} a gradient complexity of $\tilde{O}(n+n^{3/4}\epsilon^{-2}+n^{1/2}\epsilon^{-4})\cdot \e^{\tilde{O}(\gamma+d)}$ for SVRG-LD under Assumptions \ref{as1} and \ref{as3}. Note that the dependence on the dimension $d$ is not improved since $\alpha^{-1}$ may exponentially depend on $d$. Recently, \citet{ZXG2019b} proposed the Stochastic Gradient Hamiltonian Monte Carlo Methods with Recursive Variance Reduction with a gradient complexity of $\tilde{O}((n+n^{1/2}\epsilon^{-2}\mu_\ast^{-3/2})\wedge\mu_\ast^{-2}\epsilon^{-4})$ in terms of 2-Wasserstein distance. Even though their algorithm is based on the underdamped Langevin Dynamics whose discrete schemes use to perform better than those of the overdamped Langevin Dynamics such as SVRG-LD, our gradient complexity, which applies to a broader family of distributions, is almost the same except for a small interval of $\epsilon$, but we do not require the batch size $B$ and the inner loop length $m$ to depend on $\epsilon$ while \citet{ZXG2019b} do, i.e., $B\lesssim B_0^{1/2}$, $m=O(B_0/B)$, where $B_0=\tilde{O}\left(\epsilon^{-4}\mu_\ast^{-1}\wedge n\right)$. This strengthens the importance of our result since it shows that adapting this analysis to other stochastic schemes of GLD is promising and could lead to tighter bounds and relaxation of conditions. See Table \ref{tab1} for a summary. Concerning the concurrent work of \citet{BCESZ2022}, under the sole assumption of smoothness, they provided a gradient complexity of $O(L^2d^2n/\epsilon^2)$ for the Variance Reduced LMC algorithm that updates the stochastic gradient differently as SVRG-LD and SARAH-LD. This is almost the square of our result, and in some extent, our work can be interpreted as an acceleration of their result with a slightly stronger additional condition than Poincaré inequality.
\paragraph{Proof Sketch} Proceeding in a similar way as  \citet{VW2019}, we evaluate how $H_\nu(\rho_k)$ decreases at each step as shown in Theorem \ref{th1} of Appendix \ref{apa}. This is realized by comparing the evolution of the continuous-time GLD for time $\eta$ and one step of SVRG-LD. Since we use a stochastic gradient, we need at the same time to evaluate the variance of the stochastic gradient. Theorem \ref{mth1} can be obtained by recursively solving the inequality derived in Theorem \ref{th1}.
%\begin{theorem}\label{mth1bis}
%  Under Assumptions \ref{as1} and \ref{as2}, $0<\eta<\frac{\alpha}{16\sqrt{6}L^2m\gamma}$, $\gamma\ge 1$ and $B\ge m$, for all $k=sm+r$, where $s\in\mathbb{N}\cup\{0\}$ and $r=0,\ldots,m-1$, the following holds in the update of SVRG-LD where $\Xi =\frac{(n-B)}{B(n-1)}$ and $\Upsilon =\left(2+3\Xi +2m\Xi \right)$ :
%  \begin{align*}
%    H_\nu(\rho_{k+1})\le&\ \e^{-\frac{3\alpha}{2\gamma}\eta}\left(1+\frac{\alpha}{4\gamma}\eta\right)H_\nu(\rho_{sm+r})+\e^{-\frac{3\alpha}{2\gamma}\eta}\sum_{i=0}^{r-1}\frac{\alpha}{4m\gamma}\eta \e^{-\frac{\alpha m}{\gamma}\eta} H_\nu(\rho_{sm+i})+8\eta^2 dL^2\Upsilon.
%  \end{align*}
%\end{theorem}
\subsection{Convergence Analysis of SARAH-LD}
As for SARAH-LD, its convergence to the stationary distribution $\d\nu\propto\e^{-\gamma F}\d x$ can be formulated as the theorem below. Interestingly, we obtain the same result as SVRG-LD (Theorem \ref{mth1}) but we do not require $B\ge m$ anymore.
\begin{theorem}\label{mth2}
  Under Assumptions \ref{as1} and \ref{as2}, $0<\eta<\frac{\alpha}{16\sqrt{2}L^2m\gamma}$ and  $\gamma\ge 1$, for all $k=1,2,\ldots$, the following holds in the update of SARAH-LD where $\Xi =\frac{(n-B)}{B(n-1)}$ :
  \begin{align*}
    H_{\nu}(\phi_{k})&\le \e^{-\frac{\alpha\eta}{\gamma}k}H_\nu(\phi_0)+\frac{32\eta\gamma dL^2}{3\alpha}\left(2+\Xi +2m\Xi \right).
  \end{align*}
\end{theorem}
This is the first convergence guarantee of SARAH-LD in this problem setting so far, and it leads to the following gradient complexity.
\begin{corollary}\label{mcor21}
  Under the same assumptions as Theorem \ref{mth2}, for all $\epsilon\ge 0$, if we choose step size $\eta$ such that $\eta\le \frac{3\alpha\epsilon}{64\gamma dL^2}\left(2+\Xi +2m\Xi \right)^{-1}$, then a precision $H_{\nu}(\phi_{k})\le\epsilon$ is reached after $k\ge \frac{\gamma}{\alpha\eta}\log{\frac{2H_{\nu}(\phi_0)}{\epsilon}}$ steps. Especially, if we take $B=m=\sqrt{n}$ and the largest permissible step size $\eta=\frac{\alpha}{16\sqrt{2}L^2\sqrt{n}\gamma}\wedge \frac{3\alpha\epsilon}{320dL^2\gamma}$, then the gradient complexity becomes
  \[
  \tilde{O}\left(\left(n+\frac{dn^{\frac{1}{2}}}{\epsilon}\right)\cdot \frac{\gamma^2 L^2}{\alpha^2}\right).
  \]
\end{corollary}
\par The reason why we obtain the same gradient complexity for both SARAH-LD and SVRG-LD (except better coefficients for SARAH-LD) is that in our analysis, the Brownian noise added at each step of the Langevin Dynamics plays the role of a fundamental bottleneck that even SARAH-LD could not eliminate, and we still need to set $B=m=\sqrt{n}$. We can hypothesize that this order of gradient complexity might be tight for variance-reduced stochastic gradient Langevin Dynamics algorithms.

\section{Some Applications to Non-Convex Optimization}
\label{c4}

Here, we apply our main results to non-convex optimization. Thanks to our analysis applicable to a broader family of probability distributions satisfying LSI, the additional conditions we pose in this section are mainly reflected in the concrete formulation of the Log-Sobolev constant, which keeps our study simple and clear. The proofs can be found in Appendix \ref{apc}. Since SVRG-LD and SARAH-LD exhibited almost the same performance in sampling, we can simultaneously analyse them. We first prove the convergence to the global minimum of SVRG-LD and SARAH-LD without clarifying the explicit formulation of the Log-Sobolev constant in function of $\gamma$.
\begin{theorem}\label{mth3}
  Using SVRG-LD or SARAH-LD, under Assumptions \ref{as1} to \ref{as3}, $0<\eta<\frac{\alpha}{16\sqrt{6}L^2m\gamma}$, $\gamma\ge \frac{4d}{\epsilon}\log{\left(\frac{\e L}{M}\right)}\vee \frac{8db}{\epsilon^2}\vee 1\vee \frac{2}{M}$ and $B\ge m$, if we take $B=m=\sqrt{n}$ and the largest permissible step size $\eta=\frac{\alpha}{16\sqrt{6}L^2\sqrt{n}\gamma}\wedge \frac{3}{1792}\frac{\alpha^2\epsilon}{L^2d\gamma}$, the gradient complexity to reach a precision of
  \[
  \E_{X_k}[F(X_k)]-F(X^\ast)\le \epsilon
  \]
  is
  \[
  \tilde{O}\left(\left(n+\frac{n^\frac{1}{2}}{\epsilon}\cdot \frac{d L}{\alpha}\right)\frac{\gamma^2 L^2}{\alpha^2}\right),
  \]
  where $\alpha$ is a function of $\gamma$, and $X^\ast$ is the global minimum of $F$.
\end{theorem}
\begin{remark}
  Under Assumptions \ref{as1} and \ref{as3}, Assumption \ref{as2} is negligible as shown in Property \ref{p01}.
\end{remark}
Under Assumptions \ref{as1} to \ref{as3} only, this leads to a gradient complexity which exponentially depends on the inverse of the precision level $\epsilon$ as shown in the next corollary since the inverse of the Log-Sobolev constant exponentially depends on $\gamma$.
\begin{corollary}\label{mcor31}
  Under the same assumptions as Theorem \ref{mth3}, taking $\gamma=i(\epsilon)\vcentcolon=\frac{4d}{\epsilon}\log{\left(\frac{\e L}{M}\right)}\vee \frac{8db}{\epsilon^2}\vee 1\vee \frac{2}{M}$, we obtain a gradient complexity of
  \[
  \tilde{O}\left(\left(n+\frac{n^\frac{1}{2}}{\epsilon}\cdot \frac{d L}{C_1 i(\epsilon)}\e^{C_2i(\epsilon)}\right) L^2\e^{2C_2i(\epsilon)}\right)
  \]
  since $\alpha=\gamma C_1\e^{-C_2\gamma}$ (Property \ref{p1}).
\end{corollary}
\par
The second term with $n^{1/2}$ is almost all the time dominant since it has a factor that exponentially depends on $1/\epsilon$ and the first term not. This dependence on $n$ of $O(n^{1/2})$ is the best so far for these algorithms. Moreover, comparing with the gradient complexity $\tilde{O}\left(n^{1/2}\lambda^{-4}\epsilon^{-5/2}\right)\cdot\e^{\tilde{O}(d)}$, also of order $n^{1/2}$, provided by \citet{XCZG2017} who used SVRG-LD and the same assumptions, our gradient complexity is an improvement since their analysis required a batch size $B$ and an inner loop length $m$ that strongly depend on $\epsilon$ (i.e., $B=\sqrt{n}\epsilon^{-3/2}$, $m=\sqrt{n}\epsilon^{3/2}$) and ours does not. Note that the dependence of the gradient complexity of \citet{XCZG2017} on $1/\epsilon$ is not necessarily better than ours as $\lambda$ is actually the spectral gap of the discrete-time Markov chain generated by \eqref{eq21} and its inverse exponentially depends on $1/\epsilon$ as well. Although \citet{XCZG2017} did not investigate the explicit nature of $\lambda$, this is supported by \citet{RRT2017} who proved this exponential dependence for the spectral gap of the continuous-time SDE and by \citet{MSH2002} who showed the spectral gap of continuous-time SDE and that of discrete-time version are almost the same in this context.
 \paragraph{Analysis under the weak Morse condition}
  Now, under the additional Assumptions \ref{as5} to \ref{as7}, it is interesting to note that a \emph{polynomial dependence} on $1/\epsilon$ is achieved as the following corollary shows.
 \begin{corollary}\label{mcor32}
   Under the same assumptions as Theorem \ref{mth3} and Assumptions \ref{as5} to \ref{as7}, taking $\gamma=j(\epsilon)\vcentcolon=\frac{4d}{\epsilon}\log{\left(\frac{\e L}{M}\right)}\vee \frac{8db}{\epsilon^2}\vee 1\vee \frac{2}{M}\vee  C_\gamma$, where $C_\gamma$ is a constant independent of $\epsilon$ defined in Property \ref{p2}, we obtain a gradient complexity of
   \[
   \tilde{O}\left(\left(n+\frac{n^\frac{1}{2}}{\epsilon}\cdot \frac{d L}{C_3}j(\epsilon)\right)C_3^2j(\epsilon)^4 L^2\right),
   \]
   since $\alpha=C_3/\gamma$ (Property \ref{p2}).
 \end{corollary}
The crux of this corollary is Property \ref{p2}. To prove this, we show like \citet{LE2020} that $\nu$ satisfies the Poincaré inequality with a constant independent of $\gamma$. Since it is not hard to show this around the global minimum, we can step by step extend the set where this inequality holds by a Lyapunov argument (Theorems \ref{tha1} and \ref{tha1bis}). The essential difference between this analysis and that of \citet{LE2020} is that we do not work on compact manifolds anymore. Some rather minor difficulties emerge as we cannot employ the compactness but they can be addressed by supposing dissipativity which assures a quadratic growth for large $x$.
 \begin{remark}
   These results do not definitively assert that SARAH-LD and SVRG-LD show the exact same performance in terms of optimization. Indeed, suppose we are close enough to the global optimum. Then, a big noise is not necessary anymore since it is more important to stably converge to the global minimum. Here, we should be able to significantly decrease the noise $\epsilon_k$, and the bottleneck from the noise should disappear. In this case, SARAH-LD would perform better than SVRG-LD as we approach the original non-convex optimization setting where SARAH outperforms SVRG.
 \end{remark}
 \begin{remark}
   We also investigated an annealed version of SVRG-LD and SARAH-LD but could not ameliorate the gradient complexity. The detailed analysis can be found in Appendix \ref{ape}.
 \end{remark}

\section{Experiment}
\label{c4bis}

In this section, we illustrate our main result with a simple experiment.\footnote{Source code can be found in \url{https://github.com/yuri-k111/NeurIPS2022_code}} We follow the same problem setting as that of \citet{WT2011} in Section 5.1. That is, we aim to sample from a non-log-concave posterior distribution $p(\theta|x)\propto p(\theta)\prod_{i=1}^np(x_i|\theta)$ where $\{x_i\}_{i=1}^n$ is sampled from $p(x|\theta)$, a distribution parameterized by $\theta=(\theta_1,\theta_2)$. The prior $p(\theta)$ and the distribution of $x$ parametrized by $\theta$ are respectively defined as $\theta_1\sim N(0,10)$, $\theta_2\sim N(0,1)$ and $x\sim 1/2N(\theta_1,2)+1/2N(\theta_1+\theta_2,2)$. Here, we set $n=10000$, $\theta_1=0$ and $\theta_2=1$. Using the obtained 10000 samples, we simulated 1000 points of SVRG-LD with the inner loop length $m=n/B$ and different batch sizes $B$, namely, 100, 1000 and 10000 so that $B\ge m$ as required in Theorem \ref{mth1}. Evolution of KL-divergence between the true posterior, estimated by the Metropolis-adjusted Langevin algorithm, and that simulated by SVRG-LD is plotted in Figure \hyperlink{figexp1}{1}. KL-divergence was approximated following \citet{P2008}.
 \begin{figure}
   \hypertarget{figexp1}{}
   \centering
   \includegraphics[width=65mm]{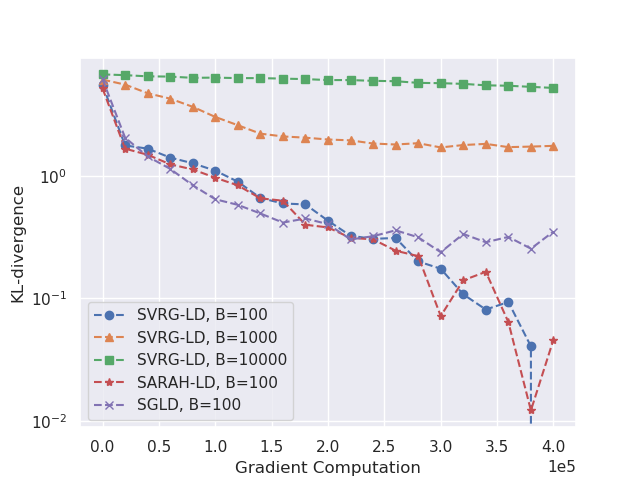}
   \caption{KL-divergence between the true and the simulated posterior. 1000 points were simulated for each algorithm with step size $\eta=0.00001$. The inner loop length $m$ for SVRG-LD was defined as $n/B$, and initial points were randomly drawn from the standard normal distribution. 1 gradient computation refers to one computation of $\nabla f_i$.}
 \end{figure}
 \par As we can observe, Figure \hyperlink{figexp1}{1} correctly reproduces the theoretical bound of Theorem \ref{mth1}, with an exponential convergence in the beginning and a persistent bias due to the use of a discrete scheme and mini-batches. The fastest convergence in terms of gradient complexity under the condition $B\ge m$ is achieved by SVRG-LD with $B=\sqrt{n}$, which confirms our main theorem. Furthermore, with this best batch size, we also simulated 1000 points of SGLD and SARAH-LD as shown in Figure \hyperlink{figexp1}{1} as well. While SGLD and SVRG-LD have similar convergence speed in the beginning, the latter eventually achieves a higher precision thanks to the variance reduction method adopted in this scheme. SARAH-LD exhibits a similar performance as SVRG-LD, which agrees with Theorem \ref{mth2}.

\section{Discussion and Conclusion}
\label{c5}
The main limitations of our work reside in the gap between practice and theory. Indeed, while our paper supposes assumptions quite standard in the literature of GLD, it cannot explain the whole empirical success that machine learning is currently experiencing. Some choices of parameters may also seem different than the practical use. However, compared to previous work, we succeeded in proving convergence of GLD with the popular stochastic gradient with relaxed conditions, and deleting the dependence of batch size and inner loop length on epsilon, which are all more realistic situations than prior work. The theoretical study in machine learning and deep learning precisely plays the role of filling as much as possible this large gap, and our work could be regarded as a further step forward to achieve this goal. Furthermore, in this paper, we focused on the pure sampling and optimization performance of the algorithms, and some of the drawbacks are simply due to this fact. For example, another limitation is that we did not investigate the generalization error in Section 4, but this was only outside the scope of this work.\par
In conclusion, we analysed the convergence rate of stochastic gradient Langevin Dynamics with variance reduction under smoothness and LSI and its application to optimization. In Section \ref{c3}, we proved the convergence of SVRG-LD in terms of KL-divergence with more relaxed conditions (LSI and smoothness) and with a better gradient complexity than previous works. We also expanded SARAH to SARAH-LD and showed that this algorithm enjoyed the same advantages as SVRG-LD with only an improvement in the coefficients of the gradient complexity. These results led us to apply SVRG-LD and SARAH-LD to non-convex optimization in Section \ref{c4}. We provided the global convergence and a non-asymptotic analysis of SVRG-LD and SARAH-LD. We obtained better conditions than prior works. Furthermore, we showed that under the additional assumption including weak Morse and Hessian Lipschitzness, the gradient complexity could be ameliorated, eliminating the exponential dependence on the inverse of the required error.
\begin{ack}
This study was partially supported by Japan Digital Design and JST CREST. We would like to thank Mufan Li and Murat A. Erdogdu for their valuable comments to fix some issues in the proof in Appendix \ref{apd} and their suggestion to improve it. We also thank anonymous reviewers for their feedback.
\end{ack}

\bibliography{biblio}
\bibliographystyle{abbrvnat}

%%%%%%%%%%%%%%%%%%%%%%%%%%%%%%%%%%%%%%%%%%%%%%%%%%%%%%%%%%%%
%%%%%%%%%%%%%%%%%%%%%%%%%%%%%%%%%%%%%%%%%%%%%%%%%%%%%%%%%%%%
%\subfile{supplementary.tex}
%\newpage
\appendix

\section{Proof of Theorem \ref{mth1} and Corollary \ref{mcor11}}\label{apa}

In this Section, to clearly differentiate from SARAH-LD, we redefine the random variable generated at the $k$-th step of SVRG-LD (Algorithm \hyperlink{al1}{1}) as $Y_k$ and the stochastic gradient as $v_k^{\scriptscriptstyle{(\mathrm{Y})}}$. The distribution of $Y_k$ is $\rho_k$.
\subsection{Preparation for the Proof}
We first prepare some lemmas.
\begin{lemmaA}\label{l1}
 Under Assumption \ref{as1},
 \[
 \E_\nu[\|\nabla F\|^2]\le dL/\gamma.
 \]
\end{lemmaA}
\begin{proof}
  As $\d\nu=\e^{-\gamma F}\d x$, using integration by parts, we obtain
  \[
  \E_\nu[\|\nabla (\gamma F)\|^2]=\E_\nu[\|\Delta (\gamma F)\|].
  \]
  Now, since $F$ is $L$-smooth by Assumption \ref{as1}, $\nabla^2 F\preceq LI$ holds, which implies $\Delta F\le dL$. As a result,
  \[
  \E_\nu[\|\nabla F\|^2]=\frac{1}{\gamma}\E_\nu[\|\Delta F\|]\le \frac{dL}{\gamma}.
  \]
\end{proof}
The relation between 2-Wasserstein distance and KL-divergence is given by the following inequality.
\begin{lemmaA}\label{l2}
  Under Assumption \ref{as2}, $\nu$ satisfies the following Talagrand's inequality with the same Log-Sobolev constant $\alpha$:
  \begin{align}\label{eq31}
    \frac{\alpha}{2}W_2(\rho,\nu)^2\le H_\nu(\rho).
  \end{align}
\end{lemmaA}
\begin{remarkA}
  See Theorem 1 of \citet{OV2000} for a proof of Lemma \ref{l2}.
\end{remarkA}
The following two lemmas that bound $\E[\|{v}_{k}^{\scriptscriptstyle{(\mathrm{Y})}}\|^2]$ and the variance of the stochastic gradient ${v}_k^{\scriptscriptstyle{(\mathrm{Y})}}$ with the KL-divergences $H_\nu(\rho_{k}),H_\nu(\rho_{k-1}),\ldots$ are crucial in our proof.
\begin{lemmaA}\label{l3}
  Under Assumption \ref{as1}, suppose Talagrand's inequality \eqref{eq31} holds for $\nu$ with a constant $\alpha$, then for all $k=sm+r$, where $s\in\mathbb{N}\cup\{0\}$ and $r=0,\ldots,m-1$, the following holds in the update of SVRG-LD:
  \[
  \E_{Y_k,I_k,Y^{(s)}}[\|{v}_{k}^{\scriptscriptstyle{(\mathrm{Y})}}\|^2]\le \Lambda 'H_\nu(\rho_{sm+r})+T+\sum_{i=0}^{r-1}S(S+1)^{r-i-1}\left(\Lambda 'H_\nu(\rho_{sm+i})+T\right),
  \]
  where $\Lambda =\left(1+\frac{2(n-B)}{B(n-1)}\right)$, $\Xi =\frac{(n-B)}{B(n-1)}$,
  \[
  \Lambda '=\frac{4L^2}{\alpha}\Lambda ,
  \]
  \[
  S=4L^2m\eta^2\Xi,
  \]
  and
  \[
  T=\frac{2dL}{\gamma}\Lambda +\frac{8\eta mdL^2}{\gamma}\Xi .
  \]
\end{lemmaA}
\begin{proof}
  Let $v_i^{(1)}(Y_k)\vcentcolon=\nabla f_{i}(Y_k)-\nabla f_{i}(Y^{(s)})+\nabla F(Y^{(s)})$, then
  \begin{align*}
    \E_{Y_k,I_k,Y^{(s)}}[\|{v}_k^{\scriptscriptstyle{(\mathrm{Y})}}\|^2]=&\E_{Y_k,I_k,Y^{(s)}}\left[\left\|\frac{1}{B}\sum_{i\in I_k}v_i^{(1)}(Y_k)\right\|^2\right]\\
    =&\frac{1}{B^2}\E_{Y_k,I_k,Y^{(s)}}\left[\sum_{i\ne i',\{i,i'\}\in I_k}\left\langle v_i^{(1)}(Y_k),v_{i'}^{(1)}(Y_k)\right\rangle\right]\\
    &+\frac{1}{B^2}\E_{Y_k,I_k,Y^{(s)}}\left[\sum_{i\in I_k}\|v_i^{(1)}(Y_k)\|^2\right]\\
    =&\frac{B-1}{Bn(n-1)}\E_{Y_k,Y^{(s)}}\left[\sum_{i\ne i'}\left\langle v_i^{(1)}(Y_k),v_{i'}^{(1)}(Y_k)\right\rangle\right]\\
    &+\frac{1}{B}\E_{Y_k,i,Y^{(s)}}\left[\|v_i^{(1)}(Y_k)\|^2\right]\\
    &\ \  (\text{$i$ follows the uniform distribution under $\{1,\ldots,n\}$})\\
    =&\frac{B-1}{Bn(n-1)}\E_{Y_k,Y^{(s)}}\left[\sum_{i, i'}\left\langle v_i^{(1)}(Y_k),v_{i'}^{(1)}(Y_k)\right\rangle\right]\\
    &-\frac{B-1}{B(n-1)}\E_{Y_k,i,Y^{(s)}}\left[\|v_i^{(1)}(Y_k)\|^2\right]\\
    &  +\frac{1}{B}\E_{Y_k,i,Y^{(s)}}\left[\|v_i^{(1)}(Y_k)\|^2\right]\\
    =&\frac{(B-1)n}{B(n-1)}\E_{Y_k}[\|\nabla F(Y_k)\|^2]+\frac{n-B}{B(n-1)}\E_{Y_k,i,Y^{(s)}}[\|v_i^{(1)}(Y_k)\|^2],
  \end{align*}
  where we used $\frac{1}{n}\sum_{i=1}^n v_i^{(1)}(Y_k)=\nabla F(Y_k)$ for the last equality.\par
  As a result, taking into account $\frac{(B-1)n}{B(n-1)}-1=\frac{B-n}{B(n-1)}\le 0$,
   \begin{equation}\label{eq32}
     \E_{Y_k,I_k,Y^{(s)}}[\|{v}_k^{\scriptscriptstyle{(\mathrm{Y})}}\|^2]=\E_{Y_k}[\|\nabla F(Y_k)\|^2]+\frac{n-B}{B(n-1)}\E_{Y_k,i,Y^{(s)}}[\|v_i^{(1)}(Y_k)\|^2].
   \end{equation}
  Choosing an optimal coupling $Y_k\sim \rho_k$ and $Y^{\ast}\sim \nu$ so that $\E[\|Y_k-Y^{\ast}\|^2]=W_2(\rho_k,\nu)^2$, we obtain
   \begin{align}
     \E_{Y_k}[\|\nabla F(Y_k)\|^2]&\le 2\E_{Y_k,Y^{\ast}}[\|\nabla F(Y_k)-\nabla F(Y^{\ast})\|^2]+2\E_{Y^{\ast}}[\|\nabla F(Y^{\ast})\|^2]\nonumber \\
     &\le 2L^2\E[\|Y_k-Y^{\ast}\|^2]+2dL/\gamma\nonumber  \\
     &=2L^2W_2(\rho_k,\nu)^2+2dL/\gamma\nonumber \\
     &\le \frac{4L^2}{\alpha}H_\nu(\rho_k)+2dL/\gamma,\label{eq33}
   \end{align}
   where we used Lemma \ref{l1} and the smoothness of $F$ for the second inequality, the definition of $W_2$ for the equality and Talagrand's inequality (Lemma \ref{l2}) for the last inequality.
   \par Moreover,
    \begin{align*}
      \E_{Y_k,i,Y^{(s)}}[\|v_i^{(1)}(Y_k)\|^2]=&\E_{Y_k,i,Y^{(s)}}\left[\left\|\nabla f_{i}(Y_k)-\nabla f_{i}(Y^{(s)})+\nabla F(Y^{(s)})\right\|^2\right]\\
      \le& 2\E\left[\left\|(\nabla f_{i}(Y_k)-\nabla f_{i}(Y^{(s)}))-\left(\nabla F(Y_k)-\nabla F(Y^{(s)})\right)\right\|^2\right]\\
      &+2\E[\|\nabla F(Y_k)\|^2]\\
      \le& 2\E\left[\left\|\nabla f_{i}(Y_k)-\nabla f_{i}(Y^{(s)})\right\|^2\right]+2\E[\|\nabla F(Y_k)\|^2]\\
      \le& 2L^2\E\left[\left\|Y_k-Y^{(s)}\right\|^2\right]+\frac{8L^2}{\alpha}H_\nu(\rho_k)+4dL/\gamma\\
      =& 2L^2\E\left[\left\|\sum_{i=1}^r\left(Y_{sm+i}-Y_{sm+i-1}\right)\right\|^2\right]\\
      &+\frac{8L^2}{\alpha}H_\nu(\rho_k)+4dL/\gamma\\
      =& 2L^2\E\left[\left\|\sum_{i=1}^r\left(-\eta{v}_{sm+i-1}^{\scriptscriptstyle{(\mathrm{Y})}}+\sqrt{\frac{2\eta}{\gamma}}\epsilon_{sm+i-1}\right)\right\|^2\right]\\
      &+\frac{8L^2}{\alpha}H_\nu(\rho_k)+4dL/\gamma\\
      \le& 4L^2\E\left[\left\|\sum_{i=1}^r\eta{v}_{sm+i-1}^{\scriptscriptstyle{(\mathrm{Y})}}\right\|^2\right]+4L^2\E\left[\left\|\sum_{i=1}^r\left(\sqrt{\frac{2\eta}{\gamma}}\epsilon_{sm+i-1}\right)\right\|^2\right]\\
      &+\frac{8L^2}{\alpha}H_\nu(\rho_k)+4dL/\gamma\\
      \le& 4r\eta^2L^2\sum_{i=1}^r\E[\|{v}_{sm+i-1}^{\scriptscriptstyle{(\mathrm{Y})}}\|^2]+\frac{8\eta L^2}{\gamma}\sum_{i=1}^r\E[\|\epsilon_{sm+i-1}\|^2]\\
      &+\frac{8L^2}{\alpha}H_\nu(\rho_k)+4dL/\gamma\\
      \le& 4m\eta^2L^2\sum_{i=1}^r\E[\|{v}_{sm+i-1}^{\scriptscriptstyle{(\mathrm{Y})}}\|^2]+\frac{8\eta mL^2 d}{\gamma}+\frac{8L^2}{\alpha}H_\nu(\rho_k)+4dL/\gamma.
    \end{align*}
    We used $\E[\|y-\E[y]\|^2]\le \E[\|y\|^2]$ for the second inequality, smoothness of $F$ and equation \eqref{eq33} for the third inequality and $r<m$ for the last inequality.\par
    Plugging these to equation \eqref{eq32}, we conclude
    \begin{align*}
      \E_{Y_k,I_k,Y^{(s)}}[\|{v}_k^{\scriptscriptstyle{(\mathrm{Y})}}\|^2]\le&\ \left(1+\frac{2(n-B)}{B(n-1)}\right)\left(\frac{4L^2}{\alpha}H_\nu(\rho_k)+\frac{2dL}{\gamma}\right)\\
      &+\frac{(n-B)}{ B(n-1)}\left(4m\eta^2L^2\sum_{i=1}^r\E[\|{v}_{sm+i-1}^{\scriptscriptstyle{(\mathrm{Y})}}\|^2]+\frac{8\eta mL^2 d}{\gamma}\right).
    \end{align*}
Therefore, setting
\[
\Lambda '=\frac{4L^2}{\alpha}\left(1+\frac{2(n-B)}{B(n-1)}\right),
\]
\[
S=4L^2m\eta^2\frac{(n-B)}{B(n-1)},
\]
and
\[
T=\frac{2dL}{\gamma}\left(1+\frac{2(n-B)}{B(n-1)}\right)+\frac{8\eta mdL^2}{\gamma}\frac{(n-B)}{B(n-1)},
\]
 we can rearrange this so that
\begin{align}\label{eq34}
  \E_{Y_k,I_k,Y^{(s)}}[\|{v}_k^{\scriptscriptstyle{(\mathrm{Y})}}\|^2]&\le \sum_{i=1}^rS\E[\|{v}_{sm+i-1}^{\scriptscriptstyle{(\mathrm{Y})}}\|^2]+\Lambda 'H_\nu(\rho_{sm+r})+T.
\end{align}
Now, we are ready to prove by mathematical induction that the inequality of the statement holds for all $r=0,\ldots, m-1$. When $r=0$, the inequality holds from equation \eqref{eq33} as follows:
\begin{align*}
  \E_{Y_k,I_k,Y^{(s)}}[\|{v}_k^{\scriptscriptstyle{(\mathrm{Y})}}\|^2]&=\E_{Y_{sm}}[\|{v}_{sm}^{\scriptscriptstyle{(\mathrm{Y})}}\|^2]\\
  &\le \E_{Y_{sm}}[\|\nabla F(Y_{sm})\|^2]+\frac{n-B}{B(n-1)}\E_{Y_{sm}}[\|v_i^{(1)}(Y_{sm})\|^2]\\
  &=\left(1+\frac{n-B}{B(n-1)}\right)\E_{Y_{sm}}[\|\nabla F(Y_{sm})\|^2]\\
  &\le \left(1+\frac{n-B}{B(n-1)}\right)\left(\frac{4L^2}{\alpha}H_\nu(\rho_{sm})+\frac{2dL}{\gamma}\right)\\
  &\le \Lambda 'H_\nu(\rho_{sm})+T,
\end{align*}
where for the second equality we used $v_i^{(1)}(Y_{sm})=\nabla F(Y_{sm})$.
\par Next, let us assume that the inequality of the lemma holds for $r\le l$. Then, from equation \eqref{eq34}, we obtain
\begin{align*}
  \E[\|{v}_{sm+l+1}^{\scriptscriptstyle{(\mathrm{Y})}}\|^2]\le& \sum_{i=0}^lS\E[\|{v}_{sm+i}^{\scriptscriptstyle{(\mathrm{Y})}}\|^2]+\Lambda 'H_\nu(\rho_{sm+l+1})+T\\
  \le& \sum_{i=0}^lS
  \left(\Lambda 'H_\nu(\rho_{sm+i})+T+\sum_{j=0}^{i-1}S(S+1)^{i-j-1}\left(\Lambda 'H_\nu(\rho_{sm+j})+T\right)\right)\\
  &+\Lambda 'H_\nu(\rho_{sm+l+1})+T\\
  =&\sum_{i=0}^lS\left(\Lambda 'H_\nu(\rho_{sm+i})+T\right)\left(1+\sum_{j=0}^{l-i-1}S(S+1)^j\right)\\
  &+\Lambda 'H_\nu(\rho_{sm+l+1})+T\\
  =&\sum_{i=0}^lS\left(\Lambda 'H_\nu(\rho_{sm+i})+T\right)\left(1+S\frac{(S+1)^{l-i}-1}{(S+1)-1}\right)\\
  &+\Lambda 'H_\nu(\rho_{sm+l+1})+T\\
  =&\Lambda 'H_\nu(\rho_{sm+l+1})+T+\sum_{i=0}^lS(S+1)^{l+1-i-1}\left(\Lambda 'H_\nu(\rho_{sm+i})+T\right).
\end{align*}
In the second inequality, we used the hypothesis of mathematical induction. This is equivalent to using Gronwall's lemma. This concludes the proof.\\
\end{proof}
\begin{lemmaA}\label{l4}
  Under Assumption \ref{as1}, for all $k=sm+r$, where $s\in\mathbb{N}\cup\{0\}$ and $r=0,\ldots,m-1$, the following holds in the update of SVRG-LD:
  \[
  \E_{Y_k,I_k,Y^{(s)}}[\|{v}_{k}^{\scriptscriptstyle{(\mathrm{Y})}}-\nabla F(Y_k)\|^2]\le \frac{L^2(n-B)}{B(n-1)}\E_{Y_k,I_k,Y^{(s)}}[\|Y_k-Y^{(s)}\|^2].
  \]
\end{lemmaA}
\begin{proof}
  Let $v_i^{(2)}(Y_k)=\nabla f_{i}(Y_k)-\nabla f_{i}(Y^{(s)})+\nabla F(Y^{(s)})-\nabla F(Y_k)$. Then,
  \begin{align*}
    \E_{Y_k,I_k,Y^{(s)}}[\|{v}_k^{\scriptscriptstyle{(\mathrm{Y})}}-\nabla F(Y_k)\|^2]=&\E_{Y_k,I_k,Y^{(s)}}\left[\left\|\frac{1}{B}\sum_{i\in I_k}v_i^{(2)}(Y_k)\right\|^2\right]\\
    =&\frac{1}{B^2}\E_{Y_k,I_k,Y^{(s)}}\left[\sum_{i\ne i',\{i,i'\}\in I_k}\left\langle v_i^{(2)}(Y_k),v_{i'}^{(2)}(Y_k)\right\rangle\right]\\
    &+\frac{1}{B^2}\E_{Y_k,I_k,Y^{(s)}}\left[\sum_{i\in I_k}\|v_i^{(2)}(Y_k)\|^2\right]\\
    =&\frac{B-1}{Bn(n-1)}\E_{Y_k,Y^{(s)}}\left[\sum_{i\ne i'}\left\langle v_i^{(2)}(Y_k),v_{i'}^{(2)}(Y_k)\right\rangle\right]\\
    &+\frac{1}{B}\E_{Y_k,i,Y^{(s)}}\left[\|v_i^{(2)}(Y_k)\|^2\right]\\
    & (\text{$i$ follows the uniform distribution under $\{1,\ldots,n\}$})\\
    =&\frac{B-1}{Bn(n-1)}\E_{Y_k,Y^{(s)}}\left[\sum_{i, i'}\left\langle v_i^{(2)}(Y_k),v_{i'}^{(2)}(Y_k)\right\rangle\right]\\
    &-\frac{B-1}{B(n-1)}\E_{Y_k,i,Y^{(s)}}\left[\|v_i^{(2)}(Y_k)\|^2\right]\\
    &  +\frac{1}{B}\E_{Y_k,i,Y^{(s)}}\left[\|v_i^{(2)}(Y_k)\|^2\right]\\
    =&\frac{n-B}{B(n-1)}\E_{Y_k,i,Y^{(s)}}[\|v_i^{(2)}(Y_k)\|^2].
  \end{align*}
  In the last equality, we used $\frac{1}{n}\sum_{i=1}^n v_i^{(2)}(Y_k)=0$.
  \par Now, since
  \begin{align*}
    \E_{Y_k,i,Y^{(s)}}[\|v_i^{(2)}(Y_k)\|^2]=&\ \E_{Y_k,i,Y^{(s)}}[\|\nabla f_{i}(Y_k)-\nabla f_{i}(Y^{(s)})+\nabla F(Y^{(s)})-\nabla F(Y_k)\|^2]\\
    =&\ \E[\|\nabla f_{i}(Y_k)-\nabla f_{i}(Y^{(s)})-\E[\nabla f_{i}(Y_k)-\nabla f_{i}(Y^{(s)})]\|^2]\\
    \le &\  \E[\|\nabla f_{i}(Y_k)-\nabla f_{i}(Y^{(s)})\|^2]\\
    \le &\ L^2\E[\|Y_k-Y^{(s)}\|^2],
  \end{align*}
  we obtain the desired result.\\
\end{proof}
\subsection{Main Proof}
We are now ready to prove the main results. The main idea of the following proofs is due to \citet{VW2019}. We first evaluate how $H_\nu(\rho_{k})$ decreases compared with the previous steps.
\begin{theoremA}\label{th1}
  Under Assumptions \ref{as1} and \ref{as2}, $0<\eta<\frac{\alpha}{16\sqrt{6}L^2m\gamma}$, $\gamma\ge 1$ and $B\ge m$, for all $k=sm+r$, where $s\in\mathbb{N}\cup\{0\}$ and $r=0,\ldots,m-1$, the following holds in the update of SVRG-LD:
  \begin{align*}
    H_\nu(\rho_{k+1})\le&\ \e^{-\frac{3\alpha}{2\gamma}\eta}\left(1+\frac{\alpha}{4\gamma}\eta\right)H_\nu(\rho_{sm+r})+\e^{-\frac{3\alpha}{2\gamma}\eta}\sum_{i=0}^{r-1}\frac{\alpha}{4m\gamma}\eta \e^{-\frac{\alpha m}{\gamma}\eta} H_\nu(\rho_{sm+i})\\
    &+8\eta^2 dL^2\Upsilon ,
  \end{align*}
  where $\Lambda =\left(1+\frac{2(n-B)}{B(n-1)}\right)$, $\Xi =\frac{(n-B)}{B(n-1)}$ and $\Upsilon =\left(\Lambda +\Xi +1+2m\Xi \right)$.
\end{theoremA}
\begin{proof}
Note that from Lemma \ref{l2}, Talagrand's inequality is satisfied with constant $\alpha$.
\par One step of SVRG-LD can be formulated as follows:
\[
Y_{sm+r+1}\leftarrow Y_{sm+r}-\eta {v}_{sm+r}^{\scriptscriptstyle{(\mathrm{Y})}}+\sqrt{2\eta/\gamma}\epsilon_{sm+r}.
\]
This can be further interpreted as the output at time $t=\eta$ of the following SDE:
\begin{equation}\label{eq35}
  \d\tilde{Y}_t=-{v}_{sm+r}^{\scriptscriptstyle{(\mathrm{Y})}}\d t+\sqrt{2/\gamma}\d B_t,\ \tilde{Y}_0=Y_{sm+r}.
\end{equation}
In this context, the distribution $\tilde{\rho}_t$ of $\tilde{Y}_t$ depends on both $Y_{sm+r}$ and
\[
\beta_{sm+r}^{\scriptscriptstyle{(\mathrm{Y})}}\vcentcolon=(I_{sm+r},Y^{(s)}).
\]Let us define their joint distribution $\tilde{\rho}_{rt\beta_{sm+r}^{\scriptscriptstyle{(\mathrm{Y})}}}$ as follows:
\begin{align*}
  \d\tilde{\rho}_{rt\beta_{sm+r}^{\scriptscriptstyle{(\mathrm{Y})}}}(Y_{sm+r},\tilde{Y}_t,\beta_{sm+r}^{\scriptscriptstyle{(\mathrm{Y})}})&=\d\tilde{\rho}_{r\beta_{sm+r}^{\scriptscriptstyle{(\mathrm{Y})}}}(Y_{sm+r},\beta_{sm+r}^{\scriptscriptstyle{(\mathrm{Y})}})\d\tilde{\rho}_{t|r\beta_{sm+r}^{\scriptscriptstyle{(\mathrm{Y})}}}(\tilde{Y}_t|Y_{sm+r},\beta_{sm+r}^{\scriptscriptstyle{(\mathrm{Y})}})\\
  &=\d\tilde{\rho}_{t\beta_{sm+r}^{\scriptscriptstyle{(\mathrm{Y})}}}(\tilde{Y}_t,\beta_{sm+r}^{\scriptscriptstyle{(\mathrm{Y})}})\d\tilde{\rho}_{r|t\beta_{sm+r}^{\scriptscriptstyle{(\mathrm{Y})}}}(Y_{sm+r}|\tilde{Y}_t,\beta_{sm+r}^{\scriptscriptstyle{(\mathrm{Y})}}).
\end{align*}
Then, the Fokker-Planck equation \eqref{eq22} when $Y_{sm+r}$ and $\beta_{sm+r}^{\scriptscriptstyle{(\mathrm{Y})}}$ are fixed becomes
\begin{align}
  \frac{\partial \tilde{\rho}_{t|r\beta_{sm+r}^{\scriptscriptstyle{(\mathrm{Y})}}}(\tilde{Y}_t|Y_{sm+r},\beta_{sm+r}^{\scriptscriptstyle{(\mathrm{Y})}})}{\partial t}=&\ \nabla\cdot(\tilde{\rho}_{t|r\beta_{sm+r}^{\scriptscriptstyle{(\mathrm{Y})}}}(\tilde{Y}_t|Y_{sm+r},\beta_{sm+r}^{\scriptscriptstyle{(\mathrm{Y})}}){v}_{sm+r}^{\scriptscriptstyle{(\mathrm{Y})}})\nonumber \\
  &+\frac{1}{\gamma}\Delta \tilde{\rho}_{t|r\beta_{sm+r}^{\scriptscriptstyle{(\mathrm{Y})}}}(\tilde{Y}_t|Y_{sm+r},\beta_{sm+r}^{\scriptscriptstyle{(\mathrm{Y})}}).\label{eq36}
\end{align}
Therefore, the following holds about the distribution $\tilde{\rho}_t$ of $\tilde{Y}_t$ governed by equation \eqref{eq35}:
\begin{align*}
  \frac{\partial \tilde{\rho}_t(y)}{\partial t}=&\int \frac{\partial \tilde{\rho}_{t|r\beta_{sm+r}^{\scriptscriptstyle{(\mathrm{Y})}}}(y|Y_{sm+r},\beta_{sm+r}^{\scriptscriptstyle{(\mathrm{Y})}})}{\partial t}\tilde{\rho}_{r\beta_{sm+r}^{\scriptscriptstyle{(\mathrm{Y})}}}(Y_{sm+r},\beta_{sm+r}^{\scriptscriptstyle{(\mathrm{Y})}})\d Y_{sm+r}\d \beta_{sm+r}^{\scriptscriptstyle{(\mathrm{Y})}}\\
  =&\int \left(\nabla\cdot(\tilde{\rho}_{t|r\beta_{sm+r}^{\scriptscriptstyle{(\mathrm{Y})}}}(y|Y_{sm+r},\beta_{sm+r}^{\scriptscriptstyle{(\mathrm{Y})}}){v}_{sm+r}^{\scriptscriptstyle{(\mathrm{Y})}})+\frac{1}{\gamma}\Delta \tilde{\rho}_{t|r\beta_{sm+r}^{\scriptscriptstyle{(\mathrm{Y})}}}(y|Y_{sm+r},\beta_{sm+r}^{\scriptscriptstyle{(\mathrm{Y})}})\right)\\
  &\ \ \ \ \ \ \ \ \ \ \ \ \ \ \ \ \ \ \ \ \ \ \ \ \ \ \ \ \ \ \ \ \ \ \ \ \ \ \ \ \ \ \ \ \ \ \ \ \ \ \ \cdot \tilde{\rho}_{r\beta_{sm+r}^{\scriptscriptstyle{(\mathrm{Y})}}}(Y_{sm+r},\beta_{sm+r}^{\scriptscriptstyle{(\mathrm{Y})}})\d Y_{sm+r}\d \beta_{sm+r}^{\scriptscriptstyle{(\mathrm{Y})}}\\
  =&\int \nabla\cdot(\tilde{\rho}_{rt\beta_{sm+r}^{\scriptscriptstyle{(\mathrm{Y})}}}(Y_{sm+r},y,\beta_{sm+r}^{\scriptscriptstyle{(\mathrm{Y})}}){v}_{sm+r}^{\scriptscriptstyle{(\mathrm{Y})}})\d Y_{sm+r}\d \beta_{sm+r}^{\scriptscriptstyle{(\mathrm{Y})}}\\
  &+\int\frac{1}{\gamma}\Delta \tilde{\rho}_{rt\beta_{sm+r}^{\scriptscriptstyle{(\mathrm{Y})}}}(Y_{sm+r},y,\beta_{sm+r}^{\scriptscriptstyle{(\mathrm{Y})}})\d Y_{sm+r}\d \beta_{sm+r}^{\scriptscriptstyle{(\mathrm{Y})}}\\
  =&\nabla\cdot\left(\tilde{\rho}_t(y)\int \tilde{\rho}_{r\beta_{sm+r}^{\scriptscriptstyle{(\mathrm{Y})}}|t}{v}_{sm+r}^{\scriptscriptstyle{(\mathrm{Y})}}\d Y_{sm+r}\d \beta_{sm+r}^{\scriptscriptstyle{(\mathrm{Y})}}\right)+\frac{1}{\gamma}\Delta\tilde{\rho}_{t}(y)\\
  =&\nabla\cdot\left(\tilde{\rho}_t(y)\E_{\tilde{\rho}_{r\beta_{sm+r}^{\scriptscriptstyle{(\mathrm{Y})}}|t}}[{v}_{sm+r}^{\scriptscriptstyle{(\mathrm{Y})}}|\tilde{Y}_t=y]\right)+\frac{1}{\gamma}\Delta\tilde{\rho}_{t}(y),
\end{align*}
where for the second equation we used equation \eqref{eq36}.\par
Plugging this to
\[
\frac{\d}{\d t}H_\nu(\tilde{\rho}_t)=\frac{\d}{\d t}\int_{\mathbb{R}^n}\tilde{\rho}_t\log{\frac{\tilde{\rho}_t}{\nu}}\d y=\int_{\mathbb{R}^n}\frac{\partial \tilde{\rho}_t}{\partial t}\log{\frac{\tilde{\rho}_t}{\nu}}\d y,
\]
we obtain
\begin{align*}
  \frac{\d}{\d t}H_\nu(\tilde{\rho}_t)=&\int_{\mathbb{R}^d}\left(\nabla\cdot\left(\tilde{\rho}_t(y)\E_{\tilde{\rho}_{r\beta_{sm+r}^{\scriptscriptstyle{(\mathrm{Y})}}|t}}[{v}_{sm+r}^{\scriptscriptstyle{(\mathrm{Y})}}|\tilde{Y}_t=y]\right)+\frac{1}{\gamma}\Delta\tilde{\rho}_{t}(y)\right)\log{\frac{\tilde{\rho}_t}{\nu}}\d y\\
  =&\int\left(\nabla\cdot\left(\tilde{\rho}_t(y)\left(\frac{1}{\gamma}\nabla \log{\frac{\tilde{\rho}_t(y)}{\nu(y)}}+\E_{\tilde{\rho}_{r\beta_{sm+r}^{\scriptscriptstyle{(\mathrm{Y})}}|t}}[{v}_{sm+r}^{\scriptscriptstyle{(\mathrm{Y})}}|\tilde{Y}_t=y]-\nabla  F(y)\right)\right)\right)\\
  &\ \ \ \ \ \ \ \ \ \ \ \ \ \ \ \ \ \ \ \ \ \ \ \ \ \ \ \ \ \ \ \ \ \ \ \ \ \ \ \ \ \ \ \ \ \ \ \ \ \ \ \ \ \ \ \ \ \ \ \ \ \ \cdot\log{\frac{\tilde{\rho}_t(y)}{\nu(y)}}\d y\\
  =&-\int\tilde{\rho}_t(y)\left\langle\frac{1}{\gamma}\nabla\log{\frac{\tilde{\rho}_t(y)}{\nu(y)}}+\E_{\tilde{\rho}_{r\beta_{sm+r}^{\scriptscriptstyle{(\mathrm{Y})}}|t}}[{v}_{sm+r}^{\scriptscriptstyle{(\mathrm{Y})}}|\tilde{Y}_t=y]-\nabla F,\nabla\log{\frac{\tilde{\rho}_t}{\nu}}\right\rangle \d y\\
  =&-\int\tilde{\rho}_t(y)\frac{1}{\gamma}\left\|\log{\frac{\tilde{\rho}_t(y)}{\nu(y)}}\right\|^2\d y\\
  &+\int\tilde{\rho}_t(y)\left\langle \nabla F(y) -\E_{\tilde{\rho}_{r\beta_{sm+r}^{\scriptscriptstyle{(\mathrm{Y})}}|t}}[{v}_{sm+r}^{\scriptscriptstyle{(\mathrm{Y})}}|\tilde{Y}_t=y],\nabla\log{\frac{\tilde{\rho}_t(y)}{\nu(y)}}\right\rangle \d y\\
  =&-\frac{1}{\gamma}J_\nu(\tilde{\rho}_t)\\
  &+\int\tilde{\rho}_{rt\beta_{sm+r}^{\scriptscriptstyle{(\mathrm{Y})}}}\left\langle \nabla F-{v}_{sm+r}^{\scriptscriptstyle{(\mathrm{Y})}},\nabla\log{\frac{\tilde{\rho}_t}{\nu}}\right\rangle \d Y_{sm+r}\d y\d \beta_{sm+r}^{\scriptscriptstyle{(\mathrm{Y})}}\\
  =&-\frac{1}{\gamma}J_\nu(\tilde{\rho}_t)+\E_{\tilde{\rho}_{rt\beta_{sm+r}^{\scriptscriptstyle{(\mathrm{Y})}}}}\left[\left\langle \nabla F(\tilde{Y}_t) -{v}_{sm+r}^{\scriptscriptstyle{(\mathrm{Y})}},\nabla\log{\frac{\tilde{\rho}_t(\tilde{Y}_t)}{\nu(\tilde{Y}_t)}}\right\rangle\right].
\end{align*}
Now, let us define the second term of the right-hand side of the very last equality as \raise0.2ex\hbox{\textcircled{\scriptsize{A}}}. Applying $\langle a,b\rangle \le \gamma\|a\|^2+\frac{1}{4\gamma}\|b\|^2$ to this, we obtain
\begin{align*}
  \raise0.2ex\hbox{\textcircled{\scriptsize{A}}}\le&\ \gamma \E_{\tilde{\rho}_{rt\beta_{sm+r}^{\scriptscriptstyle{(\mathrm{Y})}}}}\left[\|\nabla F(\tilde{Y}_t) -{v}_{sm+r}^{\scriptscriptstyle{(\mathrm{Y})}}\|^2\right]+\frac{1}{4\gamma}\E_{\tilde{\rho}_{rt\beta_{sm+r}^{\scriptscriptstyle{(\mathrm{Y})}}}}\left[\left\|\nabla\log{\frac{\tilde{\rho}_t(\tilde{Y}_t)}{\nu(\tilde{Y}_t)}}\right\|^2\right] \\
  \le&\ 2\gamma \E_{\tilde{\rho}_{rt\beta_{sm+r}^{\scriptscriptstyle{(\mathrm{Y})}}}}\left[\|\nabla F(\tilde{Y}_t) -\nabla F(Y_{sm+r})\|^2\right]\\
  &+2\gamma \E_{\tilde{\rho}_{rt\beta_{sm+r}^{\scriptscriptstyle{(\mathrm{Y})}}}}\left[\|\nabla F(Y_{sm+r}) -{v}_{sm+r}^{\scriptscriptstyle{(\mathrm{Y})}}\|^2\right]\\
  &+\frac{1}{4\gamma}J_\nu(\tilde{\rho}_t)\\
  \le&\ 2 \gamma L^2\E_{\tilde{\rho}_{rt\beta_{sm+r}^{\scriptscriptstyle{(\mathrm{Y})}}}}[\|\tilde{Y}_t-Y_{sm+r}\|^2]+\frac{2\gamma L^2(n-B)}{B(n-1)} \E_{\tilde{\rho}_{rt\beta_{sm+r}^{\scriptscriptstyle{(\mathrm{Y})}}}}\left[\|Y_{sm+r} -Y_{sm}\|^2\right]\\
  &+\frac{1}{4\gamma}J_\nu(\tilde{\rho}_t),
\end{align*}
where for the last inequality we used the smoothness of $F$ and Lemma \ref{l4}.\par
As $\tilde{Y}_t=Y_{sm+r}-t{v}_{sm+r}^{\scriptscriptstyle{(\mathrm{Y})}}+\sqrt{2t/\gamma}\epsilon_{sm+r}\ (\epsilon_{sm+r}\sim N(0,I))$, from Lemma \ref{l3}, we have
\begin{align*}
  \E[\|\tilde{Y}_t-Y_{sm+r}\|^2]=&\E[\|-t{v}_{sm+r}^{\scriptscriptstyle{(\mathrm{Y})}}+\sqrt{2t/\gamma}\epsilon_{sm+r}\|^2]\\
  =&t^2\E[\|{v}_{sm+r}^{\scriptscriptstyle{(\mathrm{Y})}}\|^2]+2td/\gamma\\
  \le& t^2\left(\Lambda 'H_\nu(\rho_{sm+r})+T+\sum_{i=0}^{r-1}S(S+1)^{r-i-1}\left(\Lambda 'H_\nu(\rho_{sm+i})+T\right)\right)\\
  &+2td/\gamma.
\end{align*}
Furthermore, by the proof of Lemma \ref{l3} we know that the following holds:
\begin{align*}
  \E\left[\|Y_{sm+r} -Y_{sm}\|^2\right]\le& 2m\eta^2\sum_{i=0}^{r-1}\E[\|{v}_{sm+i}^{\scriptscriptstyle{(\mathrm{Y})}}\|^2]+4\eta md/\gamma\\
  \le&2m\eta^2 \sum_{i=0}^{r-1}(S+1)^{r-i-1}(\Lambda 'H_\nu(\rho_{sm+i})+T)+4\eta md/\gamma.
\end{align*}
As a result, taking into account that we are only concerned about the time interval $0\le t\le\eta$, applying $t\le\eta$, we conclude
\begin{align*}
  \raise0.2ex\hbox{\textcircled{\scriptsize{A}}}\le&\ 2\gamma L^2\eta^2\left(\Lambda 'H_\nu(\rho_{sm+r})+T+\sum_{i=0}^{r-1}S(S+1)^{r-i-1}\left(\Lambda 'H_\nu(\rho_{sm+i})+T\right)\right)\\
  &+4\eta dL^2+4\gamma L^2\eta^2 m\Xi \sum_{i=0}^{r-1}(S+1)^{r-i-1}(\Lambda 'H_\nu(\rho_{sm+i})+T)+8\eta mdL^2 \Xi \\
  & +\frac{1}{4\gamma}J_\nu(\rho_t)\\
  \le&\ 2\gamma L^2\eta^2\Lambda 'H_\nu(\rho_{sm+r})+\sum_{i=0}^{r-1}4\gamma L^2\eta^2(S+1)^{r-i}\Lambda 'H_\nu(\rho_{sm+i})\\
  &+4\gamma L^2\eta^2\sum_{i=0}^{r}(S+1)^{r-i}T+4\eta dL^2(1+2m\Xi )+\frac{1}{4\gamma}J_\nu(\rho_t)\\
  \le&\ 2\gamma L^2\eta^2\Lambda 'H_\nu(\rho_{sm+r})+\sum_{i=0}^{r-1}4\gamma L^2\eta^2(S+1)^{r}\Lambda 'H_\nu(\rho_{sm+i})\\
  &+4\gamma L^2\eta^2\sum_{i=0}^{r}(S+1)^{r}T+4\eta dL^2(1+2m\Xi )+\frac{1}{4\gamma}J_\nu(\rho_t)\\
  \le&\ 2\gamma L^2\eta^2\Lambda 'H_\nu(\rho_{sm+r})+\sum_{i=0}^{r-1}4\gamma L^2\eta^2(S+1)^{m}\Lambda 'H_\nu(\rho_{sm+i})\\
  &+4\gamma L^2\eta^2m(S+1)^{m}T+4\eta dL^2(1+2m\Xi )+\frac{1}{4\gamma}J_\nu(\rho_t),
\end{align*}
where for the second inequality we used $m\Xi \le  1$ and for the last inequality $r< m$. Here, as $\Xi \le 1$ and $\eta\le \frac{1}{4 mL}$ by $\alpha\le \gamma L$,
\begin{align*}
  (S+1)^m\le \e^{Sm}=\e^{4L^2m^2\eta^2\Xi }\le \e^{1/4}\le 2.
\end{align*}
Therefore,
\begin{align*}
  \raise0.2ex\hbox{\textcircled{\scriptsize{A}}}\le &\ 2\gamma L^2\eta^2\Lambda 'H_\nu(\rho_{sm+r})+\sum_{i=0}^{r-1}8\gamma L^2\eta^2\Lambda 'H_\nu(\rho_{sm+i})\\
  &+8\gamma L^2\eta^2mT+4\eta dL^2(1+2m\Xi )+\frac{1}{4\gamma}J_\nu(\rho_t)\\
  =&\ \frac{8\gamma L^4\eta^2}{\alpha}\Lambda H_\nu(\rho_{sm+r})+\sum_{i=0}^{r-1}\frac{32\gamma L^4\eta^2}{\alpha}\Lambda H_\nu(\rho_{sm+i})\\
  &+8\gamma L^2\eta^2m\left(\frac{2dL}{\gamma}\Lambda +\frac{8\eta mdL^2}{\gamma}\Xi \right)+4\eta dL^2(1+2m\Xi )+\frac{1}{4\gamma}J_\nu(\rho_t)\\
  =&\ \frac{8\gamma L^4\eta^2}{\alpha}\Lambda H_\nu(\rho_{sm+r})+\sum_{i=0}^{r-1}\frac{32\gamma L^4\eta^2}{\alpha}\Lambda H_\nu(\rho_{sm+i})\\
  &+4\eta dL^2\left(4\eta mL\Lambda +16\eta^2m^2L^2\Xi +1+2m\Xi \right)+\frac{1}{4\gamma}J_\nu(\rho_t)\\
  \le&\ \frac{8\gamma L^4\eta^2}{\alpha}\Lambda H_\nu(\rho_{sm+r})+\sum_{i=0}^{r-1}\frac{32\gamma L^4\eta^2}{\alpha}\Lambda H_\nu(\rho_{sm+i})\\
  &+4\eta dL^2\left(\Lambda +\Xi +1+2m\Xi \right)+\frac{1}{4\gamma}J_\nu(\rho_t),
 \end{align*}
 where for the first equality, we used $\Lambda '=\frac{4L^2}{\alpha}\Lambda $ and $T=\left(\frac{2dL}{\gamma}\Lambda +\frac{8\eta mdL^2}{\gamma}\Xi \right)$, and for the last inequality $\eta\le \frac{1}{4mL}$. Thus, setting $\Upsilon =\Lambda +\Xi +1+2m\Xi $, we obtain
\begin{align*}
  \frac{\d}{\d t}H_\nu(\tilde{\rho}_t)\le&\ -\frac{3}{4\gamma}J_\nu(\tilde{\rho}_t)+\frac{8\gamma L^4\eta^2}{\alpha}\Lambda H_\nu(\rho_{sm+r})+\sum_{i=0}^{r-1}\frac{32\gamma L^4\eta^2}{\alpha}\Lambda H_\nu(\rho_{sm+i})\\
  &+4\eta dL^2\Upsilon .
\end{align*}
According to Assumption \ref{as2},
\begin{align*}
  \frac{\d}{\d t}H_\nu(\tilde{\rho}_t)\le&\ -\frac{3\alpha}{2\gamma}H_\nu(\tilde{\rho}_t)+\frac{8\gamma L^4\eta^2}{\alpha}\Lambda H_\nu(\rho_{sm+r})+\sum_{i=0}^{r-1}\frac{32\gamma L^4\eta^2}{\alpha}\Lambda H_\nu(\rho_{sm+i})\\
  &+4\eta dL^2\Upsilon .
\end{align*}
Grouping the second to fourth terms as $U_{sm+r}^{\scriptscriptstyle{\mathrm{(Y)}}}$ and multiplying both sides by $\e^{\frac{3\alpha}{2\gamma}t}$, we can write the above equation as
\begin{align*}
  \frac{\d}{\d t}\left(\e^{\frac{3\alpha}{2\gamma}t}H_\nu(\tilde{\rho}_t)\right)&\le \e^{\frac{3\alpha}{2\gamma}t}U_{sm+r}^{\scriptscriptstyle{\mathrm{(Y)}}}.
\end{align*}
Integrating both sides from $t=0$ to $t=\eta$ and using $\tilde{\rho}_\eta=\rho_{sm+r+1}$, we obtain
\begin{align*}
  \e^{\frac{3\alpha}{2\gamma}\eta}H_\nu(\rho_{sm+r+1})-H_\nu(\rho_{sm+r})&\le \frac{2\gamma(\e^{\frac{3\alpha}{2\gamma}\eta}-1)}{3\alpha}U_{sm+r}^{\scriptscriptstyle{\mathrm{(Y)}}}\\
  &\le 2\eta U_{sm+r}^{\scriptscriptstyle{\mathrm{(Y)}}}.
\end{align*}
Here, for the last inequality, we used that $\e^c\le 1+2c\  (0<c=\frac{3\alpha}{2\gamma}\eta\le1)$ holds since $0<\eta\le\frac{\alpha}{16\sqrt{6}L^2m^\gamma}\le\frac{2\gamma }{3\alpha}$, where we used $1/L\le \gamma/\alpha$ and $m\ge 1$. Rearranging this, we obtain
\begin{align}\label{eq37}
  H_\nu(\rho_{sm+r+1})\le&\ \e^{-\frac{3\alpha}{2\gamma}\eta}\left(1+\frac{16\gamma L^4\eta^3}{\alpha}\Lambda \right)H_\nu(\rho_{sm+r})+\e^{-\frac{3\alpha}{2\gamma}\eta}\sum_{i=0}^{r-1}\frac{64\gamma L^4\eta^3}{\alpha}\Lambda H_\nu(\rho_{sm+i})\nonumber \\
  &+\e^{-\frac{3\alpha}{2\gamma}\eta}8\eta^2 dL^2\Upsilon .
\end{align}
Furthermore, since $\eta\le\frac{\alpha}{16\sqrt{6}mL^2\gamma}\le\frac{\alpha}{8\sqrt{3}L^2\gamma}$, $\e^{-\frac{3\alpha}{2\gamma}\eta}\le 1$ and $\Lambda \le 3$
\begin{align*}
  H_\nu(\rho_{sm+r+1})&\le \e^{-\frac{3\alpha}{2\gamma}\eta}\left(1+\frac{\alpha}{4\gamma}\eta\right)H_\nu(\rho_{sm+r})+\e^{-\frac{3\alpha}{2\gamma}\eta}\sum_{i=0}^{r-1}\frac{\alpha}{8\gamma m}\eta H_\nu(\rho_{sm+i})+8\eta^2 dL^2\Upsilon .
\end{align*}
On the other hand, since $\eta\le\frac{\alpha}{8mL^2\gamma}$ and $\alpha\le \gamma L$ holds,
\[
\e^{-\frac{\alpha m}{ \gamma}\eta}\ge \e^{-\frac{\alpha m}{\gamma}\cdot\frac{\alpha}{8mL^2\gamma }}=\e^{-\frac{\alpha^2}{8L^2\gamma^2}}\ge \e^{-1/8}\ge 0.88\ge \frac{1}{2},
\]
which further implies
\begin{align*}
  H_\nu(\rho_{sm+r+1})\le&\ \e^{-\frac{3\alpha}{2\gamma}\eta}\left(1+\frac{\alpha}{4\gamma}\eta\right)H_\nu(\rho_{sm+r})+\e^{-\frac{3\alpha}{2\gamma}\eta}\sum_{i=0}^{r-1}\frac{\alpha}{4m\gamma}\eta \e^{-\frac{\alpha m}{\gamma}\eta} H_\nu(\rho_{sm+i})\\
  &+8\eta^2 dL^2\Upsilon .
\end{align*}
\end{proof}
Finally, let us prove Theorem \ref{mth1} and Corollary \ref{mcor11}.
\begin{theoremA}[Theorem \ref{mth1} restated]\label{th2}
  Under Assumptions \ref{as1} and \ref{as2}, $0<\eta<\frac{\alpha}{16\sqrt{6}L^2m\gamma}$, $\gamma\ge 1$ and $B\ge m$, for all $k\ge 1$, the following holds in the update of SVRG-LD:
  \begin{align*}
    H_{\nu}(\rho_{k})&\le \e^{-\frac{\alpha\eta}{\gamma}k}H_\nu(\rho_0)+\frac{224\eta\gamma dL^2}{3\alpha}\Upsilon ,
  \end{align*}
  where $\Xi =\frac{(n-B)}{B(n-1)}$ and $\Upsilon =\left(\Lambda +\Xi +1+2m\Xi \right)$.
\end{theoremA}
\begin{proof}
  Let us first prove by mathematical induction that the following inequality holds for all $k=1,2\ldots$:
  \begin{align*}
    H_{\nu}(\rho_{k})&\le \e^{-\frac{\alpha\eta}{\gamma}k}H_\nu(\rho_0)+8\eta^2 dL^2\Upsilon \cdot\left(1-\e^{-\frac{\alpha\eta}{\gamma}}\right)^{-1}.\ \ \ \ \ \ \ \ \ldots\ \ \hypertarget{ast}{}(\ast)
  \end{align*}
  (I) When $k=1$, from Theorem \ref{th1}, since $Y^{(s)}=Y_0$,
    \begin{align*}
      H_\nu(\rho_{1})&\le \e^{-\frac{3\alpha}{2\gamma}\eta}\left(1+\frac{\alpha}{4\gamma}\eta\right)H_\nu(\rho_{0})+\e^{-\frac{3\alpha}{2\gamma}\eta}\frac{\alpha}{4m\gamma}\eta \e^{-\frac{\alpha m}{\gamma}\eta} H_\nu(\rho_{0})+8\eta^2 dL^2\Upsilon \\
      &\le \e^{-\frac{3\alpha}{2\gamma}\eta}\left(1+\frac{\alpha}{4\gamma}\eta+\frac{\alpha}{4m\gamma}\eta\right)H_\nu(\rho_{0})+8\eta^2 dL^2\Upsilon \\
      &\le \e^{-\frac{3\alpha}{2\gamma}\eta}\left(1+\frac{\alpha}{2\gamma}\eta\right)H_\nu(\rho_{0})+8\eta^2 dL^2\Upsilon \\
      &\le \e^{-\frac{3\alpha}{2\gamma}\eta}\e^{\frac{\alpha}{2\gamma}\eta}H_\nu(\rho_{0})+8\eta^2 dL^2\Upsilon \\
      &= \e^{-\frac{\alpha}{\gamma}\eta}H_\nu(\rho_{0})+8\eta^2 dL^2\Upsilon \\
      &\le \e^{-\frac{\alpha}{\gamma}\eta}H_\nu(\rho_{0})+8\eta^2 dL^2\Upsilon \cdot\left(1-\e^{-\frac{\alpha\eta}{\gamma}}\right)^{-1}.
    \end{align*}
Here, for the second and last inequality, we used $\e^{-\frac{\alpha m\eta}{\gamma}}\le \e^{-\frac{\alpha \eta}{\gamma}}\le 1$. Thus, \hyperlink{ast}{$(\ast)$} holds for $k=1$.\\
(II) Now, let us assume that \hyperlink{ast}{$(\ast)$} holds for all $k\le l$. Letting $r$ and $s$ the remainder and quotient of the Euclidian division of $l$ by $m$ respectively, when $k=l+1$ we obtain from Theorem \ref{th1},
  \begin{align*}
    H_\nu(\rho_{sm+r+1})&\le\ \e^{-\frac{3\alpha}{2\gamma}\eta}\left(1+\frac{\alpha}{4\gamma}\eta\right)H_\nu(\rho_{sm+r})+\e^{-\frac{3\alpha}{2\gamma}\eta}\sum_{i=0}^{r-1}\frac{\alpha}{4m\gamma}\eta \e^{-\frac{\alpha m}{\gamma}\eta} H_\nu(\rho_{sm+i})\\
    &+8\eta^2 dL^2\Upsilon .
  \end{align*}
   From the hypothesis of mathematical induction,
   \begin{align*}
     H_\nu(\rho_{sm+r+1}&)\\
     \le&\ \e^{-\frac{3\alpha}{2\gamma}\eta}\left(1+\frac{\alpha}{4\gamma}\eta\right)\left(\e^{-\frac{\alpha\eta}{\gamma}(sm+r)}H_\nu(\rho_0)+8\eta^2 dL^2\Upsilon \cdot\left(1-\e^{-\frac{\alpha\eta}{\gamma}}\right)^{-1}\right)\\
     &+\e^{-\frac{3\alpha}{2\gamma}\eta}\sum_{i=0}^{r-1}\frac{\alpha}{4m\gamma}\eta \e^{-\frac{\alpha m}{\gamma}\eta} \left( \e^{-\frac{\alpha\eta}{\gamma}(sm+i)}H_\nu(\rho_0)+8\eta^2 dL^2\Upsilon \cdot\left(1-\e^{-\frac{\alpha\eta}{\gamma}}\right)^{-1}\right)\\
     &+8\eta^2 dL^2\Upsilon .
   \end{align*}
   Since $\e^{-\frac{\alpha}{\gamma}\eta m}\le \e^{-\frac{\alpha}{\gamma}\eta r }\le \e^{-\frac{\alpha}{\gamma}\eta(r-i)}$ when $0\le i<r<m$,
   \begin{align*}
     H_\nu(\rho_{sm+r+1})\le&\ \e^{-\frac{3\alpha}{2\gamma}\eta}\left(1+\frac{\alpha}{4\gamma}\eta\right)\left(\e^{-\frac{\alpha\eta}{\gamma}(sm+r)}H_\nu(\rho_0)+8\eta^2 dL^2\Upsilon \cdot\left(1-\e^{-\frac{\alpha\eta}{\gamma}}\right)^{-1}\right)\\
     &+\e^{-\frac{3\alpha}{2\gamma}\eta}\sum_{i=0}^{r-1}\frac{\alpha}{4m\gamma}\eta \e^{-\frac{\alpha \eta}{\gamma}(r-i)} \left( \e^{-\frac{\alpha\eta}{\gamma}(sm+i)}H_\nu(\rho_0)\right)\\
     &+\e^{-\frac{3\alpha}{2\gamma}\eta}\sum_{i=0}^{r-1}\frac{\alpha}{4m\gamma}\eta \e^{-\frac{\alpha \eta}{\gamma}(r-i)} \left(8\eta^2 dL^2\Upsilon \cdot\left(1-\e^{-\frac{\alpha\eta}{\gamma}}\right)^{-1}\right)\\
     &+8\eta^2 dL^2\Upsilon \\
     \le&\ \e^{-\frac{3\alpha}{2\gamma}\eta}\left(1+\frac{\alpha}{4\gamma}\eta\right)\left(\e^{-\frac{\alpha\eta}{\gamma}(sm+r)}H_\nu(\rho_0)+8\eta^2 dL^2\Upsilon \cdot\left(1-\e^{-\frac{\alpha\eta}{\gamma}}\right)^{-1}\right)\\
     &+\e^{-\frac{3\alpha}{2\gamma}\eta}\sum_{i=0}^{r-1}\frac{\alpha}{4m\gamma}\eta \left( \e^{-\frac{\alpha\eta}{\gamma}(sm+r)}H_\nu(\rho_0)+8\eta^2 dL^2\Upsilon \cdot\left(1-\e^{-\frac{\alpha\eta}{\gamma}}\right)^{-1}\right)\\
     &+8\eta^2 dL^2\Upsilon \\
     \le&\ \e^{-\frac{3\alpha}{2\gamma}\eta}\left(1+\frac{\alpha}{2\gamma}\eta\right)\left(\e^{-\frac{\alpha\eta}{\gamma}(sm+r)}H_\nu(\rho_0)+8\eta^2 dL^2\Upsilon \cdot\left(1-\e^{-\frac{\alpha\eta}{\gamma}}\right)^{-1}\right)\\
     &+8\eta^2 dL^2\Upsilon \\
     \le&\ \e^{-\frac{3\alpha}{2\gamma}\eta}\e^{\frac{\alpha}{2\gamma}\eta}\left(\e^{-\frac{\alpha\eta}{\gamma}(sm+r)}H_\nu(\rho_0)+8\eta^2 dL^2\Upsilon \cdot\left(1-\e^{-\frac{\alpha\eta}{\gamma}}\right)^{-1}\right)\\
     &+8\eta^2 dL^2\Upsilon \\
     =&\ \e^{-\frac{\alpha\eta}{\gamma}(sm+r+1)}H_\nu(\rho_0)+\left(1+\e^{-\frac{\alpha\eta}{\gamma}}\cdot\left(1-\e^{-\frac{\alpha\eta}{\gamma}}\right)^{-1}\right)8\eta^2 dL^2\Upsilon \\
     =&\ \e^{-\frac{\alpha\eta}{\gamma}(sm+r+1)}H_\nu(\rho_0)+8\eta^2 dL^2\Upsilon \cdot\left(1-\e^{-\frac{\alpha\eta}{\gamma}}\right)^{-1}.
   \end{align*}
Therefore, \hyperlink{ast}{$(\ast)$} holds for all $k\ge 1$.\par
Now, using the inequality $1-\e^{-c}\ge \frac{3}{4}c$ for $0<c=\frac{\alpha\eta}{\gamma}\le\frac{1}{4}$ (since $y=1-\e^{-x}$ and $y=\frac{3}{4}x$ are both concave increasing functions intersecting at $x=0$ and $1-\e^{-1/4}\ge\frac{3}{4}\times \frac{1}{4}$), which holds here because $\eta\le\frac{\alpha}{16\sqrt{6}L^2\gamma}\le\frac{\gamma}{4\alpha}$ since $1/L\le \gamma/\alpha$ and $m\ge 1$, we conclude
   \begin{align*}
     H_{\nu}(\rho_{k})&\le \e^{-\frac{\alpha\eta}{\gamma}k}H_\nu(\rho_0)+\frac{32\eta\gamma dL^2}{3\alpha}\Upsilon \\
     &\le \e^{-\frac{\alpha\eta}{\gamma}k}H_\nu(\rho_0)+\frac{224\eta\gamma dL^2}{3\alpha},
   \end{align*}
which is the desired result. Here, for the last inequality, we used $\Upsilon =\Lambda +\Xi +1+2m\Xi \le 3+1+1+2=7$.\\
\end{proof}
\begin{corollaryA}[Corollary \ref{mcor11} restated]\label{cor21}
  Under the same assumptions as Theorem \ref{th2}, for all $\epsilon\ge 0$, if we choose step size $\eta$ such that
  \[
\eta\le \frac{3\alpha\epsilon}{448\gamma dL^2},
  \] then a precision $H_{\nu}(\rho_{k})\le\epsilon$ is reached after
  \[
  k\ge \frac{\gamma}{\alpha\eta}\log{\frac{2H_{\nu}(\rho_0)}{\epsilon}}
  \]
  steps. Especially, if we take $B=m=\sqrt{n}$ and the largest permissible step size $\eta=\frac{\alpha}{16\sqrt{6}L^2\sqrt{n}\gamma}\wedge \frac{3\alpha\epsilon}{448dL^2\gamma}$, then the gradient complexity becomes
  \[
  \tilde{O}\left(\left(n+\frac{dn^{\frac{1}{2}}}{\epsilon}\right)\frac{\gamma^2 L^2}{\alpha^2}\right).
  \]
\end{corollaryA}
\begin{proof}
  Let $\epsilon>0$. Then, by additionally requiring
  \[
  \eta\le \frac{3\alpha\epsilon}{448\gamma dL^2},
  \]
we obtain
  \[
  H_\nu(\rho_k)\le \e^{-\frac{\alpha\eta}{\gamma}k}H_\nu(\rho_0)+\frac{\epsilon}{2}.
  \]
Thus, $H_\nu(\rho_k)\le \epsilon$ can be reached for
  \[
  k\ge \frac{\gamma}{\alpha\eta}\log{\frac{2H_{\nu}(\rho_0)}{\epsilon}}.
  \]
  As a result, if $0<\epsilon\le\frac{28d}{3\sqrt{6}m}$ and we select the largest permissible step size, the gradient complexity becomes
  \[
    O\left(k\cdot B+\frac{k}{m}\cdot n\right)=\tilde{O}\left(\left(\frac{B+n/m}{\epsilon}\right)\frac{d\gamma^2 L^2}{\alpha^2}\right),
  \]
  and the optimal complexity is
  \[
  \tilde{O}\left(\frac{dn^{1/2}\gamma^2 L^2}{\epsilon\alpha^2}\right)
  \]
  with $B=\sqrt{n}$ and $m=\sqrt{n}$.
  \par On the other hand, if $\epsilon\ge\frac{28d}{3\sqrt{6}m}$ and we select the largest permissible step size, the gradient complexity becomes
  \[
    O\left(k\cdot B+\frac{k}{m}\cdot n\right)=\tilde{O}\left(\left(mB+n\right)\frac{\gamma^2 L^2}{\alpha^2}\right),
  \]
  and the optimal complexity is
  \[
  \tilde{O}\left(\frac{n \gamma^2 L^2}{\alpha^2}\right)
  \]
  with $B=\sqrt{n}$ and $m=\sqrt{n}$
  \par Therefore, for all $\epsilon\ge 0$, the gradient complexity is
  \[
  \tilde{O}\left(\left(n+\frac{dn^{\frac{1}{2}}}{\epsilon}\right)\frac{\gamma^2 L^2}{\alpha^2}\right).
  \]
\end{proof}

\section{Proof of Theorem \ref{mth2} and Corollary \ref{mcor21}}\label{apb}

In this Section, to clearly differentiate from SVRG-LD, we redefine the random variable generated at the $k$-th step of SARAH-LD (Algorithm \hyperlink{al1}{1}) as $Z_k$ and the stochastic gradient as $v_k^{\scriptscriptstyle{(\mathrm{Z})}}$. The distribution of $Z_k$ is $\phi_k$.
\subsection{Preparation for the Proof}
Let us first provide an upper bound of $\E[\|v^{\scriptscriptstyle{(\mathrm{Z})}}_{k}\|^2]$ and the variance of the stochastic gradient $v^{\scriptscriptstyle{(\mathrm{Z})}}_k$ using the KL-divergences $H_\nu(\phi_{k}),H_\nu(\phi_{k-1}),\ldots$.
\begin{lemmaA}\label{l5}
  Under Assumption \ref{as1}, for all $k=sm+r$, where $s\in\mathbb{N}\cup\{0\}$ and $r=0,\ldots,m-1$, the following holds in the update of SARAH-LD:
  \[
  \E[\|\nabla F(Z_k)-v^{\scriptscriptstyle{(\mathrm{Z})}}_k\|^2]= \sum_{i=1}^r\E[\|v^{\scriptscriptstyle{(\mathrm{Z})}}_{sm+i}-v^{\scriptscriptstyle{(\mathrm{Z})}}_{sm+i-1}\|^2]-\sum_{i=1}^r\E[\|\nabla F(Z_{sm+i})-\nabla F(Z_{sm+i-1})\|^2].
  \]
\end{lemmaA}
\begin{proof}
 Let us define  \[\mathcal{F}_r=\sigma\left(Z^{(s)},\epsilon_{sm},I_{sm+1},\epsilon_{sm+1},I_{sm+2},\epsilon_{sm+2},\ldots,I_{sm+r-1},\epsilon_{sm+r-1}\right),\]
  which is the $\sigma$-$algebra$ generated by
  \[
  Z^{(s)},\epsilon_{sm},I_{sm+1},\epsilon_{sm+1},I_{sm+2},\epsilon_{sm+2},\ldots,I_{sm+r-1},\ \mathrm{and}\ \epsilon_{sm+r-1}.
  \]
  When $r=0$, the statement clearly holds. In the remainder of the proof, we assume $r\ge 1$. Then,
  \begin{align*}
    \E[\|\nabla F(Z_k)-v^{\scriptscriptstyle{(\mathrm{Z})}}_k\|^2\mid\mathcal{F}_r]=&\  \E[\|\nabla F(Z_{k-1})-v^{\scriptscriptstyle{(\mathrm{Z})}}_{k-1}+\nabla F(Z_k)-\nabla F(Z_{k-1})\\
    & \ \ \ \ \ \ \ \ \ \ \ \ \ \ \ \ \ \ \ \ \ \ \ \ \ \ \ \ \ \ \ \ \ \ \ \ \ \ \ \ \ \ \ \ \ \ \ \ \ \ \ \ \ \ \ \ -(v^{\scriptscriptstyle{(\mathrm{Z})}}_k-v^{\scriptscriptstyle{(\mathrm{Z})}}_{k-1})\|^2\mid\mathcal{F}_r]\\
    =&\ \|\nabla F(Z_{k-1})-v^{\scriptscriptstyle{(\mathrm{Z})}}_{k-1}\|^2+\|\nabla F(Z_k)-\nabla F(Z_{k-1})\|^2\\
    &+\E[\|v^{\scriptscriptstyle{(\mathrm{Z})}}_k-v^{\scriptscriptstyle{(\mathrm{Z})}}_{k-1}\|^2\mid\mathcal{F}_r]\\
    &+ 2\left\langle \nabla F(Z_{k-1})-v^{\scriptscriptstyle{(\mathrm{Z})}}_{k-1},\nabla F(Z_k)-\nabla F(Z_{k-1})\right\rangle \\
    &-2\left\langle \nabla F(Z_{k-1})-v^{\scriptscriptstyle{(\mathrm{Z})}}_{k-1},\E[v^{\scriptscriptstyle{(\mathrm{Z})}}_k-v^{\scriptscriptstyle{(\mathrm{Z})}}_{k-1}\mid\mathcal{F}_r]\right\rangle\\
    &-2\left\langle \nabla F(Z_k)-\nabla F(Z_{k-1}),\E[v^{\scriptscriptstyle{(\mathrm{Z})}}_k-v^{\scriptscriptstyle{(\mathrm{Z})}}_{k-1}\mid\mathcal{F}_r]\right\rangle\\
    =&\ \|\nabla F(Z_{k-1})-v^{\scriptscriptstyle{(\mathrm{Z})}}_{k-1}\|^2-\|\nabla F(Z_k)-\nabla F(Z_{k-1})\|^2\\
    &+\E[\|v^{\scriptscriptstyle{(\mathrm{Z})}}_k-v^{\scriptscriptstyle{(\mathrm{Z})}}_{k-1}\|^2\mid\mathcal{F}_r].
  \end{align*}
  Here in the last equality, we used that the following holds:
  \begin{align*}
    \E[v^{\scriptscriptstyle{(\mathrm{Z})}}_k-v^{\scriptscriptstyle{(\mathrm{Z})}}_{k-1}\mid\mathcal{F}_r]=&\E\left[\frac{1}{B}\sum_{i\in I_k}\nabla f_i(Z_k)-\nabla f_i(Z_{k-1})\mid\mathcal{F}_r\right]\\
    =&\nabla F(Z_k)-\nabla F(Z_{k-1}).
  \end{align*}
  Taking expectation, we obtain
  \begin{align*}
    \E[\|\nabla F(Z_k)-v^{\scriptscriptstyle{(\mathrm{Z})}}_k\|^2]=&\ \E[\|\nabla F(Z_{k-1})-v^{\scriptscriptstyle{(\mathrm{Z})}}_{k-1}\|^2]-\E[\|\nabla F(Z_k)-\nabla F(Z_{k-1})\|^2]\\
    &+\E[\|v^{\scriptscriptstyle{(\mathrm{Z})}}_k-v^{\scriptscriptstyle{(\mathrm{Z})}}_{k-1}\|^2].
  \end{align*}
  Since this equation holds for all $k=sm+r\ (r=1,\ldots m-1)$, recalling that
  \[
  \E[\|\nabla F(Z_{sm})-v^{\scriptscriptstyle{(\mathrm{Z})}}_{sm}\|^2]=0,
  \]
  and recursively applying this, we conclude that
  \[
  \E[\|\nabla F(Z_k)-v^{\scriptscriptstyle{(\mathrm{Z})}}_k\|^2]= \sum_{i=1}^r\E[\|v^{\scriptscriptstyle{(\mathrm{Z})}}_{sm+i}-v^{\scriptscriptstyle{(\mathrm{Z})}}_{sm+i-1}\|^2]-\sum_{i=1}^r\E[\|\nabla F(Z_{sm+i})-\nabla F(Z_{sm+i-1})\|^2].
  \]
\end{proof}
\begin{lemmaA}\label{l6}
  Under Assumption \ref{as1}, for all $k=sm+r$, where $s\in\mathbb{N}\cup\{0\}$ and $r=0,\ldots,m-1$, the following holds in the update of SARAH-LD:
  \[
  \E[\|\nabla F(Z_k)-v^{\scriptscriptstyle{(\mathrm{Z})}}_k\|^2]\le  \sum_{i=1}^r\Xi L^2\eta^2\E[\|v^{\scriptscriptstyle{(\mathrm{Z})}}_{sm+i-1}\|^2]+\frac{2\eta m dL^2}{\gamma}\Xi ,
  \]
  where $\Xi =\frac{n-B}{B(n-1)}$.
\end{lemmaA}
\begin{proof}
When $r=0$, the statement clearly holds. In the remainder of the proof, we assume $r\ge 1$.\par
Since $v^{\scriptscriptstyle{(\mathrm{Z})}}_k-v^{\scriptscriptstyle{(\mathrm{Z})}}_{k-1}=\frac{1}{B}\sum_{j\in I_k}\left(\nabla f_{j}(Z_k)-\nabla f_{j}(Z_{k-1})\right)$, defining
\[
w_j\vcentcolon=\nabla f_{j}(Z_k)-\nabla f_{j}(Z_{k-1}),
\]
we obtain
\begin{align*}
  \E[\|v^{\scriptscriptstyle{(\mathrm{Z})}}_k-v^{\scriptscriptstyle{(\mathrm{Z})}}_{k-1}\|^2\mid\mathcal{F}_k]=&\ \E\left[\left\|\frac{1}{B}\sum_{j\in I_k}w_j\right\|^2\mid\mathcal{F}_k\right]\\
  =&\ \frac{1}{B^2}\E\left[\sum_{j\ne j',\{j,j'\}\in I_k}\left\langle w_j,w_{j'}\right\rangle\mid\mathcal{F}_k\right]+\frac{1}{B^2}\E\left[\sum_{j\in I_k}\|w_j\|^2\mid\mathcal{F}_k\right]\\
  =&\ \frac{B-1}{Bn(n-1)}\E\left[\sum_{j\ne j'}\left\langle w_j,w_{j'}\right\rangle\mid\mathcal{F}_k\right]+\frac{1}{B}\E\left[\|w_j\|^2\mid\mathcal{F}_k\right]\\
  &\ \ (\text{$j$ follows a uniform distribution under $\{1,\ldots,n\}$})\\
  =&\ \frac{B-1}{Bn(n-1)}\E\left[\sum_{j, j'}\left\langle w_j,w_{j'}\right\rangle\mid\mathcal{F}_k\right]-\frac{B-1}{B(n-1)}\E\left[\|w_j\|^2\mid\mathcal{F}_k\right]\\
  &  +\frac{1}{B}\E\left[\|w_j\|^2\mid\mathcal{F}_k\right]\\
  =&\ \frac{(B-1)n}{B(n-1)}\E[\|\nabla F(Z_k)-\nabla F(Z_{k-1})\|^2\mid\mathcal{F}_k]\\
  &+\frac{n-B}{B(n-1)}\E[\|\nabla f_{j}(Z_k)-\nabla f_{j}(Z_{k-1})\|^2\mid\mathcal{F}_k]\\
  \le& \ \E[\|\nabla F(Z_k)-\nabla F(Z_{k-1})\|^2|\mathcal{F}_k]\\
  &+\frac{n-B}{B(n-1)}\E[\|\nabla f_{j}(Z_k)-\nabla f_{j}(Z_{k-1})\|^2\mid\mathcal{F}_k],
\end{align*}
 where for the fifth equation we used $\frac{1}{n}\sum_{j=1}^n w_j=\nabla F(Z_k)-\nabla F(Z_{k-1})$ and for the inequality, $\frac{(B-1)n}{B(n-1)}\le 1$.
 \par As a result,
 \begin{align*}
   \E[\|v^{\scriptscriptstyle{(\mathrm{Z})}}_k-v^{\scriptscriptstyle{(\mathrm{Z})}}_{k-1}\|^2\mid\mathcal{F}_k]\le&\ \|\nabla F(Z_k)-\nabla F(Z_{k-1})\|^2\\
   &+ \frac{n-B}{B(n-1)}\E[\|\nabla f_{j}(Z_k)-\nabla f_{j}(Z_{k-1})\|^2\mid\mathcal{F}_k]\\
   \le &\ \|\nabla F(Z_k)-\nabla F(Z_{k-1})\|^2+ L^2\Xi \E[\|Z_k-Z_{k-1}\|^2\mid\mathcal{F}_k]\\
   = & \ \|\nabla F(Z_k)-\nabla F(Z_{k-1})\|^2+L^2 \Xi  \left\|-\eta v^{\scriptscriptstyle{(\mathrm{Z})}}_{k-1}+\sqrt{\frac{2\eta}{\gamma}}\epsilon_{k-1}\right\|^2.
 \end{align*}
 Taking expectation, we obtain
 \begin{align*}
   \E[\|v^{\scriptscriptstyle{(\mathrm{Z})}}_k-v^{\scriptscriptstyle{(\mathrm{Z})}}_{k-1}\|^2]-\E[\|\nabla F(Z_k)-\nabla F(Z_{k-1})\|^2]\le &\  L^2 \Xi  \E\left[\left\|-\eta v^{\scriptscriptstyle{(\mathrm{Z})}}_{k-1}+\sqrt{\frac{2\eta}{\gamma}}\epsilon_{k-1}\right\|^2\right]\\
   = &\ L^2 \Xi \left(\eta^2 \E[\|v^{\scriptscriptstyle{(\mathrm{Z})}}_{k-1}\|^2]+\frac{2\eta d}{\gamma}\right).
 \end{align*}
 Since this equation holds for all $k=sm+r\ (r=1,\ldots m-1)$, from Lemma \ref{l5},
 \begin{align*}
   \E[\|\nabla F(Z_k)-v^{\scriptscriptstyle{(\mathrm{Z})}}_k\|^2]\le& \sum_{i=1}^r\E[\|v^{\scriptscriptstyle{(\mathrm{Z})}}_{sm+i}-v^{\scriptscriptstyle{(\mathrm{Z})}}_{sm+i-1}\|^2]\\
   &-\sum_{i=1}^r\E[\|\nabla F(Z_{sm+i})-\nabla F(Z_{sm+i-1})\|^2]\\
   \le &\ \Xi L^2\eta^2 \sum_{i=1}^r\E[\|v^{\scriptscriptstyle{(\mathrm{Z})}}_{sm+i-1}\|^2]+\frac{2\eta r dL^2}{\gamma}\Xi \\
   \le &\ \Xi L^2\eta^2 \sum_{i=1}^r\E[\|v^{\scriptscriptstyle{(\mathrm{Z})}}_{sm+i-1}\|^2]+\frac{2\eta m dL^2}{\gamma}\Xi .
 \end{align*}
\end{proof}
\begin{lemmaA}\label{l7}
  Under Assumption \ref{as1}, suppose Talagrand's inequality holds for $\nu$ with a constant $\alpha$, then for all $k=sm+r$, where $s\in\mathbb{N}\cup\{0\}$ and $r=0,\ldots,m-1$, the following holds in the update of SARAH-LD:
  \[
  \E[\|v^{\scriptscriptstyle{(\mathrm{Z})}}_{k}\|^2]\le \frac{8L^2}{\alpha}H_\nu(\phi_{sm+r})+P+\sum_{i=0}^{r-1}Q(Q+1)^{r-i-1}\left(\frac{8L^2}{\alpha}H_\nu(\phi_{sm+i})+P\right),
  \]
  where
  \[
  \Xi =\frac{(n-B)}{B(n-1)},
  \]
  \[
  P=\frac{4dL}{\gamma}+\frac{4\eta mdL^2}{\gamma}\Xi,
  \]
  and
  \[
  Q=2\Xi L^2\eta^2.
  \]
\end{lemmaA}
\begin{proof}
  First, from Lemma \ref{l6}, we have
  \begin{align*}
    \E[\|v^{\scriptscriptstyle{(\mathrm{Z})}}_{k}\|^2]\le &\ 2\E[\|v^{\scriptscriptstyle{(\mathrm{Z})}}_{k}-\nabla F(Z_k)\|^2]+2\E[\|\nabla F(Z_k)\|^2]\\
    \le &\ 2\left(\sum_{i=1}^r\Xi L^2\eta^2\E[\|v^{\scriptscriptstyle{(\mathrm{Z})}}_{sm+i-1}\|^2]+\frac{2\eta m dL^2}{\gamma}\Xi \right)+2\E[\|\nabla F(Z_k)\|^2].
  \end{align*}
  Choosing an optimal coupling $Z_k\sim \phi_k$ and $Z^{\ast}\sim \nu$ so that $\E[\|Z_k-Z^{\ast}\|^2]=W_2(\phi_k,\nu)^2$, we obtain
     \begin{align}
       \E_{Z_k}[\|\nabla F(Z_k)\|^2]&\le 2\E_{Z_k,Z^{\ast}}[\|\nabla F(Z_k)-\nabla F(Z^{\ast})\|^2]+2\E_{Z^{\ast}}[\|\nabla F(Z^{\ast})\|^2]\nonumber \\
       &\le 2L^2\E[\|Z_k-Z^{\ast}\|^2]+2dL/\gamma\nonumber \\
       &=2L^2W_2(\phi_k,\nu)^2+2dL/\gamma\nonumber \\
       &\le \frac{4L^2}{\alpha}H_\nu(\phi_k)+2dL/\gamma,\label{eq41}
     \end{align}
     where, we used the smoothness of $F$ and Lemma \ref{l1} for the second inequality, the definition of $W_2$ for the equality and Talagrand's inequality for the last inequality.
     \par As a result,
  \begin{align}
    \E[\|v^{\scriptscriptstyle{(\mathrm{Z})}}_{k}\|^2]\le &2\left(\sum_{i=1}^r\Xi L^2\eta^2\E[\|v^{\scriptscriptstyle{(\mathrm{Z})}}_{sm+i-1}\|^2]+\frac{2\eta m dL^2}{\gamma}\Xi \right)+\frac{8L^2}{\alpha}H_\nu(\phi_k)+4dL/\gamma\nonumber\\
    =& \sum_{i=1}^rQ\E[\|v^{\scriptscriptstyle{(\mathrm{Z})}}_{sm+i-1}\|^2]+\frac{8L^2}{\alpha}H_\nu(\phi_k)+P.\label{eq42}
  \end{align}
  Here, we set
  \[
  P=\frac{4dL}{\gamma}+\frac{4\eta mdL^2}{\gamma}\Xi,
  \]
  and
  \[
  Q=2\Xi L^2\eta^2.
  \]
  \par Now, let us prove by mathematical induction that the inequality of the statement holds for all $r=0,\ldots, m-1$. When $r=0$, the inequality holds from equation \eqref{eq41} as follows:
  \begin{align*}
    \E[\|v^{\scriptscriptstyle{(\mathrm{Z})}}_{sm}\|^2]&=\E[\|\nabla F(Z_{sm})\|^2]\\
    &\le \frac{4L^2}{\alpha}H_\nu(\phi_{sm})+2dL/\gamma\\
    &\le \frac{8L^2}{\alpha}H_\nu(\phi_{sm})+P.
  \end{align*}
  Next, let us assume that the inequality of the lemma holds for $r\le l$. Then, from equation \eqref{eq42}, we obtain
  \begin{align*}
    \E[\|&v^{\scriptscriptstyle{(\mathrm{Z})}}_{sm+l+1} \|^2]\\
    &\le\  \sum_{i=0}^lQ\E[\|v^{\scriptscriptstyle{(\mathrm{Z})}}_{sm+i}\|^2]+\frac{8L^2}{\alpha}H_\nu(\phi_{sm+l+1})+P\\
    &\le\sum_{i=0}^lQ
    \left(\frac{8L^2}{\alpha}H_\nu(\phi_{sm+i})+P+\sum_{j=0}^{i-1}Q(Q+1)^{i-j-1}\left(\frac{8L^2}{\alpha}H_\nu(\phi_{sm+j})+P\right)\right)\\
    &\ \ \   +\frac{8L^2}{\alpha}H_\nu(\phi_{sm+l+1})+P\\
    &= \frac{8L^2}{\alpha}H_\nu(\phi_{sm+l+1})+P+\sum_{i=0}^lQ\left(\frac{8L^2}{\alpha}H_\nu(\phi_{sm+i})+P\right)\left(1+\sum_{j=0}^{l-i-1}Q(Q+1)^j\right)\\
    &= \frac{8L^2}{\alpha}H_\nu(\phi_{sm+l+1})+P+\sum_{i=0}^lQ\left(\frac{8L^2}{\alpha}H_\nu(\phi_{sm+i})+P\right)\left(1+Q\frac{(Q+1)^{l-i}-1}{(Q+1)-1}\right)\\
    &= \frac{8L^2}{\alpha}H_\nu(\phi_{sm+l+1})+P+\sum_{i=0}^lQ(Q+1)^{l+1-i-1}\left(\frac{8L^2}{\alpha}H_\nu(\phi_{sm+i})+P\right).
  \end{align*}
  In the second inequality, we used the hypothesis of mathematical induction. This is equivalent to using Gronwall's lemma. This concludes the proof.\\
\end{proof}
\subsection{Main Proof}
We are now ready to prove the main results. The main idea of the following proofs is due to \citet{VW2019}. We first evaluate how $H_\nu(\phi_{k})$ decreases compared with the previous steps.
\begin{theoremA}\label{th3}
  Under Assumptions \ref{as1} and \ref{as2}, $0<\eta<\frac{\alpha}{16\sqrt{2}L^2m\gamma}$ and $\gamma\ge 1$, for all $k=sm+r$, where $s\in\mathbb{N}\cup\{0\}$ and $r=0,\ldots,m-1$, the following holds in the update of SARAH-LD:
  \begin{align*}
    H_\nu(\phi_{sm+r+1})\le&\ \e^{-\frac{3\alpha}{2\gamma}\eta}\left(1+\frac{\alpha}{4\gamma}\eta\right)H_\nu(\phi_{sm+r})+\e^{-\frac{3\alpha}{2\gamma}\eta}\sum_{i=0}^{r-1}\frac{\alpha}{4m\gamma}\eta \e^{-\frac{\alpha m}{\gamma}\eta} H_\nu(\phi_{sm+i})\\
    &+8\eta^2 dL^2\left(2+\Xi +2m\Xi \right),
  \end{align*}
  where $\Xi =\frac{(n-B)}{B(n-1)}$.
\end{theoremA}
\begin{proof}
  Note that from Lemma \ref{l2}, Talagrand's inequality is satisfied with constant $\alpha$.
  \par One step of SVRG-LD can be formulated as follows:
\[
Z_{sm+r+1}\leftarrow Z_{sm+r}-\eta v^{\scriptscriptstyle{(\mathrm{Z})}}_{sm+r}+\sqrt{2\eta/\gamma}\epsilon_{sm+r}.
\]
This can be further interpreted as the output at time $t=\eta$ of the following SDE:
\begin{equation}\label{eq43}
  \d\tilde{Z}_t=-v^{\scriptscriptstyle{(\mathrm{Z})}}_{sm+r}\d t+\sqrt{2/\gamma}\d B_t,\ \tilde{Z}_0=Z_{sm+r}.
\end{equation}
In this context, the distribution $\tilde{\phi}_t$ of $\tilde{Z}_t$ depends on both $Z_{sm+r}$ and
\[
\beta_{sm+r}^{\scriptscriptstyle{(\mathrm{Z})}}\vcentcolon=(v^{\scriptscriptstyle{(\mathrm{Z})}}_{sm+r-1},I_{sm+r}).
\]
Let us define their joint distribution as follows:
\begin{align*}
  \d\tilde{\phi}_{rt\beta_{sm+r}^{\scriptscriptstyle{(\mathrm{Z})}}}(Z_{sm+r},\tilde{Z}_t,\beta_{sm+r}^{\scriptscriptstyle{(\mathrm{Z})}})=&\ \d\tilde{\phi}_{r\beta_{sm+r}^{\scriptscriptstyle{(\mathrm{Z})}}}(Z_{sm+r},\beta_{sm+r}^{\scriptscriptstyle{(\mathrm{Z})}})\d\tilde{\phi}_{t|r\beta_{sm+r}^{\scriptscriptstyle{(\mathrm{Z})}}}(\tilde{Z}_t|Z_{sm+r},\beta_{sm+r}^{\scriptscriptstyle{(\mathrm{Z})}})\\
  =&\ \d\tilde{\phi}_{t\beta_{sm+r}^{\scriptscriptstyle{(\mathrm{Z})}}}(\tilde{Z}_t,\beta_{sm+r}^{\scriptscriptstyle{(\mathrm{Z})}})\d\tilde{\phi}_{r|t\beta_{sm+r}^{\scriptscriptstyle{(\mathrm{Z})}}}(Z_{sm+r}|\tilde{Z}_t,\beta_{sm+r}^{\scriptscriptstyle{(\mathrm{Z})}}).
\end{align*}
Then, the Fokker-Planck equation \eqref{eq22} when $Z_{sm+r}$ and  $\beta_{sm+r}^{\scriptscriptstyle{(\mathrm{Z})}}$ are fixed becomes
\begin{align}
  \frac{\partial \tilde{\phi}_{t|r\beta_{sm+r}^{\scriptscriptstyle{(\mathrm{Z})}}}(\tilde{Z}_t|Z_{sm+r},\beta_{sm+r}^{\scriptscriptstyle{(\mathrm{Z})}})}{\partial t}=&\ \nabla\cdot(\tilde{\phi}_{t|r\beta_{sm+r}^{\scriptscriptstyle{(\mathrm{Z})}}}(\tilde{Z}_t|Z_{sm+r},\beta_{sm+r}^{\scriptscriptstyle{(\mathrm{Z})}})v^{\scriptscriptstyle{(\mathrm{Z})}}_{sm+r})\nonumber \\
  &+\frac{1}{\gamma}\Delta \tilde{\phi}_{t|r\beta_{sm+r}^{\scriptscriptstyle{(\mathrm{Z})}}}(\tilde{Z}_t|Z_{sm+r},\beta_{sm+r}^{\scriptscriptstyle{(\mathrm{Z})}}).\label{eq44}
\end{align}
Therefore, the following holds about the distribution $\tilde{\phi}_t$ of $\tilde{Z}_t$ governed by equation \eqref{eq43},
\begin{align*}
  \frac{\partial \tilde{\phi}_t(z)}{\partial t}=&\ \int \frac{\partial \tilde{\phi}_{t|r\beta_{sm+r}^{\scriptscriptstyle{(\mathrm{Z})}}}(z|Z_{sm+r},\beta_{sm+r}^{\scriptscriptstyle{(\mathrm{Z})}})}{\partial t}\tilde{\phi}_{r\beta_{sm+r}^{\scriptscriptstyle{(\mathrm{Z})}}}(Z_{sm+r},\beta_{sm+r}^{\scriptscriptstyle{(\mathrm{Z})}})\d Z_{sm+r}\d \beta_{sm+r}^{\scriptscriptstyle{(\mathrm{Z})}}\\
  =&\ \int \left(\nabla\cdot(\tilde{\phi}_{t|r\beta_{sm+r}^{\scriptscriptstyle{(\mathrm{Z})}}}(z|Z_{sm+r},\beta_{sm+r}^{\scriptscriptstyle{(\mathrm{Z})}})v^{\scriptscriptstyle{(\mathrm{Z})}}_{sm+r})+\frac{1}{\gamma}\Delta \tilde{\phi}_{t|r\beta_{sm+r}^{\scriptscriptstyle{(\mathrm{Z})}}}(z|Z_{sm+r},\beta_{sm+r}^{\scriptscriptstyle{(\mathrm{Z})}})\right)\\
  &\ \ \ \ \ \ \ \ \ \ \ \ \ \ \ \ \ \ \ \ \ \ \ \ \ \ \ \ \ \ \ \ \ \ \ \ \ \ \ \ \ \ \ \ \ \ \ \ \cdot \tilde{\phi}_{r\beta_{sm+r}^{\scriptscriptstyle{(\mathrm{Z})}}}(Z_{sm+r},\beta_{sm+r}^{\scriptscriptstyle{(\mathrm{Z})}})\d Z_{sm+r}\d \beta_{sm+r}^{\scriptscriptstyle{(\mathrm{Z})}}\\
  =&\ \int \nabla\cdot(\tilde{\phi}_{rt\beta_{sm+r}^{\scriptscriptstyle{(\mathrm{Z})}}}(Z_{sm+r},z,\beta_{sm+r}^{\scriptscriptstyle{(\mathrm{Z})}})v^{\scriptscriptstyle{(\mathrm{Z})}}_{sm+r})\d  Z_{sm+r}\d \beta_{sm+r}^{\scriptscriptstyle{(\mathrm{Z})}}\\
  &+\int\frac{1}{\gamma}\Delta \tilde{\phi}_{rt\beta_{sm+r}^{\scriptscriptstyle{(\mathrm{Z})}}}(Z_{sm+r},z,\beta_{sm+r}^{\scriptscriptstyle{(\mathrm{Z})}})\d  Z_{sm+r}\d \beta_{sm+r}^{\scriptscriptstyle{(\mathrm{Z})}}\\
  =&\ \nabla\cdot\left(\tilde{\phi}_t(z)\int \tilde{\phi}_{r\beta_{sm+r}^{\scriptscriptstyle{(\mathrm{Z})}}|t}v^{\scriptscriptstyle{(\mathrm{Z})}}_{sm+r}\d Z_{sm+r}\d \beta_{sm+r}^{\scriptscriptstyle{(\mathrm{Z})}}\right)+\frac{1}{\gamma}\Delta\tilde{\phi}_{t}(z)\\
  =&\ \nabla\cdot\left(\tilde{\phi}_t(z)\E_{\tilde{\phi}_{r\beta_{sm+r}^{\scriptscriptstyle{(\mathrm{Z})}}|t}}[v^{\scriptscriptstyle{(\mathrm{Z})}}_{sm+r}|\tilde{Z}_t=z]\right)+\frac{1}{\gamma}\Delta\tilde{\phi}_{t}(z),
\end{align*}
where for the second equation we used equation \eqref{eq44}.\par
Plugging this to
\[
\frac{\d}{\d t}H_\nu(\tilde{\phi}_t)=\frac{\d}{\d t}\int_{\mathbb{R}^n}\tilde{\phi}_t\log{\frac{\tilde{\phi}_t}{\nu}}\d z=\int_{\mathbb{R}^n}\frac{\partial \tilde{\phi}_t}{\partial t}\log{\frac{\tilde{\phi}_t}{\nu}}\d z,
\]
we obtain
\begin{align*}
  \frac{\d}{\d t}H_\nu(\tilde{\phi}_t)=&\ \int_{\mathbb{R}^n}\left(\nabla\cdot\left(\tilde{\phi}_t(z)\E_{\tilde{\phi}_{rZ|t}}[v^{\scriptscriptstyle{(\mathrm{Z})}}_{sm+r}|\tilde{Z}_t=z]\right)+\frac{1}{\gamma}\Delta\tilde{\phi}_{t}(z)\right)\log{\frac{\tilde{\phi}_t}{\nu}}\d z\\
  =&\ \int\left(\nabla\cdot\left(\tilde{\phi}_t\left(\frac{1}{\gamma}\nabla \log{\frac{\tilde{\phi}_t}{\nu}}+\E_{\tilde{\phi}_{r\beta_{sm+r}^{\scriptscriptstyle{(\mathrm{Z})}}|t}}[v^{\scriptscriptstyle{(\mathrm{Z})}}_{sm+r}|\tilde{Z}_t=z]-\nabla  F\right)\right)\right)\log{\frac{\tilde{\phi}_t}{\nu}}\d z\\
  =&\ -\int\tilde{\phi}_t\left\langle\frac{1}{\gamma}\nabla\log{\frac{\tilde{\phi}_t}{\nu}}+\E_{\tilde{\phi}_{r\beta_{sm+r}^{\scriptscriptstyle{(\mathrm{Z})}}|t}}[v^{\scriptscriptstyle{(\mathrm{Z})}}_{sm+r}|\tilde{Z}_t=z]-\nabla F,\nabla\log{\frac{\tilde{\phi}_t}{\nu}}\right\rangle \d z\\
  =&\ -\int\tilde{\phi}_t\frac{1}{\gamma}\left\|\log{\frac{\tilde{\phi}_t}{\nu}}\right\|^2\d z\\
  &+\int_{\mathbb{R}^n}\tilde{\phi}_t\left\langle \nabla F -\E_{\tilde{\phi}_{r\beta_{sm+r}^{\scriptscriptstyle{(\mathrm{Z})}}|t}}[v^{\scriptscriptstyle{(\mathrm{Z})}}_{sm+r}|\tilde{Z}_t=z],\nabla\log{\frac{\tilde{\phi}_t}{\nu}}\right\rangle \d z\\
  =&\ -\frac{1}{\gamma}J_\nu(\tilde{\phi}_t)\\
  &+\int\tilde{\phi}_{rt\beta_{sm+r}^{\scriptscriptstyle{(\mathrm{Z})}}}\left\langle \nabla F-v^{\scriptscriptstyle{(\mathrm{Z})}}_{sm+r},\nabla\log{\frac{\tilde{\phi}_t}{\nu}}\right\rangle \d Z_{sm+r}\d z\d \beta_{sm+r}^{\scriptscriptstyle{(\mathrm{Z})}}\\
  =&\ -\frac{1}{\gamma}J_\nu(\tilde{\phi}_t)+\E_{\tilde{\phi}_{rt\beta_{sm+r}^{\scriptscriptstyle{(\mathrm{Z})}}}}\left[\left\langle \nabla F(\tilde{Z}_t) -v^{\scriptscriptstyle{(\mathrm{Z})}}_{sm+r},\nabla\log{\frac{\tilde{\phi}_t(\tilde{Z}_t)}{\nu(\tilde{Z}_t)}}\right\rangle\right].
\end{align*}
Now, let us define the second term of the right-hand side of the very last equality as \raise0.2ex\hbox{\textcircled{\scriptsize{B}}}. Applying $\langle a,b\rangle \le \gamma\|a\|^2+\frac{1}{4\gamma}\|b\|^2$ to this, we obtain
\begin{align*}
  \raise0.2ex\hbox{\textcircled{\scriptsize{B}}}\le&\ \gamma \E_{\tilde{\phi}_{rt\beta_{sm+r}^{\scriptscriptstyle{(\mathrm{Z})}}}}\left[\|\nabla F(\tilde{Z}_t) -v^{\scriptscriptstyle{(\mathrm{Z})}}_{sm+r}\|^2\right]+\frac{1}{4\gamma}\E_{\tilde{\phi}_{rt\beta_{sm+r}^{\scriptscriptstyle{(\mathrm{Z})}}}}\left[\left\|\nabla\log{\frac{\tilde{\phi}_t(\tilde{Z}_t)}{\nu(\tilde{Z}_t)}}\right\|^2\right] \\
  \le&\ 2\gamma \E_{\tilde{\phi}_{rt\beta_{sm+r}^{\scriptscriptstyle{(\mathrm{Z})}}}}\left[\|\nabla F(\tilde{Z}_t) -\nabla F(Z_{sm+r})\|^2\right]+2\gamma \E_{\tilde{\phi}_{rt\beta_{sm+r}^{\scriptscriptstyle{(\mathrm{Z})}}}}\left[\|\nabla F(Z_{sm+r}) -v^{\scriptscriptstyle{(\mathrm{Z})}}_{sm+r}\|^2\right]\\
  &+\frac{1}{4\gamma}J_\nu(\tilde{\phi}_t)\\
  \le&\ 2 \gamma L^2\E_{\tilde{\phi}_{rt\beta_{sm+r}^{\scriptscriptstyle{(\mathrm{Z})}}}}[\|\tilde{Z}_t-Z_{sm+r}\|^2]+\sum_{i=1}^r2\gamma \Xi L^2\eta^2\E[\|v^{\scriptscriptstyle{(\mathrm{Z})}}_{sm+i-1}\|^2]+4\eta m dL^2\Xi \\
  &+\frac{1}{4\gamma}J_\nu(\tilde{\phi}_t),
\end{align*}
where for the last inequality, we used the smoothness of $F$ and Lemma \ref{l6}.
\par As $\tilde{Z}_t=Z_{sm+r}-tv^{\scriptscriptstyle{(\mathrm{Z})}}_{sm+r}+\sqrt{2t/\gamma}\epsilon_{sm+r}\ (\epsilon_{sm+r}\sim N(0,I))$, from Lemma \ref{l7}, we have
\begin{align*}
  \E[\|\tilde{Z}_t-Z_{sm+r}\|^2]=&\ \E[\|-tv^{\scriptscriptstyle{(\mathrm{Z})}}_{sm+r}+\sqrt{2t/\gamma}\epsilon_{sm+r}\|^2]\\
  =&\ t^2\E[\|v^{\scriptscriptstyle{(\mathrm{Z})}}_{sm+r}\|^2]+2td/\gamma\\
  \le&\  t^2\left(\frac{8L^2}{\alpha}H_\nu(\phi_{sm+r})+P\right)\\
  &+t^2\sum_{i=0}^{r-1}Q(Q+1)^{r-i-1}\left(\frac{8L^2}{\alpha}H_\nu(\phi_{sm+i})+P\right)\\
 &+2td/\gamma.
\end{align*}
Furthermore, by the proof of Lemma \ref{l7}, we know that the following holds:
\begin{align*}
  \sum_{i=1}^r\E[\|v^{\scriptscriptstyle{(\mathrm{Z})}}_{sm+i-1}\|^2]
  &\le \sum_{i=0}^{r-1}(Q+1)^{r-i-1}\left( \frac{8L^2}{\alpha}H_\nu(\phi_{sm+i})+P\right).
\end{align*}
As a result, taking into account that we are only concerned about the time interval $0\le t\le\eta$, applying $t\le\eta$, we conclude
\begin{align*}
  \raise0.2ex\hbox{\textcircled{\scriptsize{B}}}\le&\ 2\gamma L^2\eta^2\left(\frac{8L^2}{\alpha}H_\nu(\phi_{sm+r})+P+\sum_{i=0}^{r-1}Q(Q+1)^{r-i-1}\left(\frac{8L^2}{\alpha}H_\nu(\phi_{sm+i})+P\right)
\right)\\
  &+4\eta dL^2+2\gamma L^2\eta^2 \Xi \sum_{i=0}^{r-1}(Q+1)^{r-i-1}\left( \frac{8L^2}{\alpha}H_\nu(\phi_{sm+i})+P\right)+4\eta mdL^2 \Xi \\
  & +\frac{1}{4\gamma}J_\nu(\tilde{\phi}_t)\\
  \le&\  \frac{16L^4\gamma\eta^2}{\alpha}H_\nu(\phi_{sm+r})+\sum_{i=0}^{r-1}(Q+1)^{r-i}\frac{16\gamma L^4\eta^2}{\alpha}H_\nu(\phi_{sm+i})+2\gamma L^2\eta^2\sum_{i=0}^{r}(Q+1)^{r-i}P\\
  &+4\eta dL^2(1+2m\Xi )+\frac{1}{4\gamma}J_\nu(\tilde{\phi}_t)\\
  \le&\  \frac{16L^4\gamma\eta^2}{\alpha}H_\nu(\phi_{sm+r})+\sum_{i=0}^{r-1}\frac{16L^4\gamma\eta^2}{\alpha}(Q+1)^{r}H_\nu(\phi_{sm+i})+2\gamma L^2\eta^2\sum_{i=0}^{r}(Q+1)^{r}P\\
  &+4\eta dL^2(1+2m\Xi )+\frac{1}{4\gamma}J_\nu(\tilde{\phi}_t)\\
  \le&\  \frac{16L^4\gamma\eta^2}{\alpha}H_\nu(\phi_{sm+r})+\sum_{i=0}^{r-1}\frac{16L^4\gamma\eta^2}{\alpha}(Q+1)^{m}H_\nu(\phi_{sm+i})+2\gamma L^2\eta^2m(Q+1)^{m}P\\
  &+4\eta dL^2(1+2m\Xi )+\frac{1}{4\gamma}J_\nu(\tilde{\phi}_t).
\end{align*}
where for the second inequality we used $\Xi \le  1$ and for the last inequality $r< m$.
\par Here, as $\Xi \le 1$ and $\eta\le \frac{1}{4 mL}$ by $\alpha\le \gamma L$,
\begin{align*}
  (Q+1)^m\le \e^{Qm}=\e^{2L^2m\eta^2\Xi }\le \e^{1/4}\le 2.
\end{align*}
\par Therefore,
\begin{align*}
  \raise0.2ex\hbox{\textcircled{\scriptsize{B}}}\le&\  \frac{16L^4\gamma\eta^2}{\alpha}H_\nu(\phi_{sm+r})+\sum_{i=0}^{r-1}\frac{32L^4\gamma\eta^2}{\alpha}H_\nu(\phi_{sm+i})\\
  &+4\gamma L^2\eta^2mP+4\eta dL^2(1+2m\Xi )+\frac{1}{4\gamma}J_\nu(\tilde{\phi}_t)\\
  \le&\  \frac{16L^4\gamma\eta^2}{\alpha}H_\nu(\phi_{sm+r})+\sum_{i=0}^{r-1}\frac{32L^4\gamma\eta^2}{\alpha}H_\nu(\phi_{sm+i})\\
  &+4\gamma L^2\eta^2m\left(\frac{4dL}{\gamma}+\frac{4\eta mdL^2}{\gamma}\Xi \right)+4\eta dL^2(1+2m\Xi )+\frac{1}{4\gamma}J_\nu(\tilde{\phi}_t)\\
  \le&\  \frac{16L^4\gamma\eta^2}{\alpha}H_\nu(\phi_{sm+r})+\sum_{i=0}^{r-1}\frac{32L^4\gamma\eta^2}{\alpha}H_\nu(\phi_{sm+i})\\
  &+4\eta dL^2\left(2+\Xi +2m\Xi \right)+\frac{1}{4\gamma}J_\nu(\tilde{\phi}_t).
 \end{align*}
 where for the last inequality, we used $\eta\le \frac{1}{4mL}$.
\par Thus,
\begin{align*}
  \frac{\d}{\d t}H_\nu(\tilde{\phi}_t)&\le -\frac{3}{4\gamma}J_\nu(\tilde{\phi}_t)+\frac{16L^4\gamma\eta^2}{\alpha}H_\nu(\phi_{sm+r})+\sum_{i=0}^{r-1}\frac{32L^4\gamma\eta^2}{\alpha}H_\nu(\phi_{sm+i})\\
  &\ \ +4\eta dL^2\left(2+\Xi +2m\Xi \right).
\end{align*}
According to Assumption \ref{as2},
\begin{align*}
  \frac{\d}{\d t}H_\nu(\tilde{\phi}_t)&\le -\frac{3\alpha}{2\gamma}H_\nu(\tilde{\phi}_t)+\frac{16L^4\gamma\eta^2}{\alpha}H_\nu(\phi_{sm+r})+\sum_{i=0}^{r-1}\frac{32L^4\gamma\eta^2}{\alpha}H_\nu(\phi_{sm+i})\\
  &\ \ +4\eta dL^2\left(2+\Xi +2m\Xi \right).
\end{align*}
Grouping the second to fourth terms as $U_{sm+r}^{\scriptscriptstyle{\mathrm{(Z)}}}$ and multiplying both sides by $\e^{\frac{3\alpha}{2\gamma}t}$, we can write the above equation as
\begin{align*}
  \frac{\d}{\d t}\left(\e^{\frac{3\alpha}{2\gamma}t}H_\nu(\tilde{\phi}_t)\right)&\le \e^{\frac{3\alpha}{2\gamma}t}U_{sm+r}^{\scriptscriptstyle{\mathrm{(Z)}}}.
\end{align*}
Integrating both sides from $t=0$ to $t=\eta$ and using $\tilde{\phi}_\eta=\phi_{sm+r+1}$, we obtain
\begin{align*}
  \e^{\frac{3\alpha}{2\gamma}\eta}H_\nu(\phi_{sm+r+1})-H_\nu(\phi_{sm+r})&\le \frac{2\gamma(\e^{\frac{3\alpha}{2\gamma}\eta}-1)}{3\alpha}U_{sm+r}^{\scriptscriptstyle{\mathrm{(Z)}}}\\
  &\le 2\eta U_{sm+r}^{\scriptscriptstyle{\mathrm{(Z)}}}.
\end{align*}
Here, for the last inequality, we used $\e^c\le 1+2c\  (0<c=\frac{3\alpha}{2\gamma}\eta\le1)$ holds since $0<\eta\le\frac{\alpha}{16\sqrt{2}L^2m^\gamma}\le\frac{2\gamma }{3\alpha}$, where we used $1/L\le \gamma/\alpha$ and $m\ge 1$. Rearranging this, we obtain
\begin{align}\label{eq45}
  H_\nu(\phi_{sm+r+1})&\le \e^{-\frac{3\alpha}{2\gamma}\eta}\left(1+\frac{32\gamma L^4\eta^3}{\alpha}\right)H_\nu(\phi_{sm+r})+\e^{-\frac{3\alpha}{2\gamma}\eta}\sum_{i=0}^{r-1}\frac{64\gamma L^4\eta^3}{\alpha}H_\nu(\phi_{sm+i})\nonumber \\
  &\ \ +\e^{-\frac{3\alpha}{2\gamma}\eta}8\eta^2 dL^2\left(2+\Xi +2m\Xi \right).
\end{align}
Furthermore, since $\eta\le\frac{\alpha}{16\sqrt{2}mL^2\gamma}\le\frac{\alpha}{8\sqrt{3}L^2\gamma}$ and $\e^{-\frac{3\alpha}{2\gamma}\eta}\le 1$,
\begin{align*}
  H_\nu(\phi_{sm+r+1})\le&\ \e^{-\frac{3\alpha}{2\gamma}\eta}\left(1+\frac{\alpha}{4\gamma}\eta\right)H_\nu(\phi_{sm+r})+\e^{-\frac{3\alpha}{2\gamma}\eta}\sum_{i=0}^{r-1}\frac{\alpha}{8\gamma m}\eta H_\nu(\phi_{sm+i})\\
  &+8\eta^2 dL^2\left(2+\Xi +2m\Xi \right).
\end{align*}
On the other hand, since $\eta\le\frac{\alpha}{8mL^2\gamma}$ and $\alpha\le \gamma L$ holds,
\[
\e^{-\frac{\alpha m}{ \gamma}\eta}\ge \e^{-\frac{\alpha m}{\gamma}\cdot\frac{\alpha}{8mL^2\gamma }}=\e^{-\frac{\alpha^2}{8L^2\gamma^2}}\ge \e^{-1/8}\ge 0.88\ge \frac{1}{2},
\]
which further implies
\begin{align*}
  H_\nu(\phi_{sm+r+1})\le&\ \e^{-\frac{3\alpha}{2\gamma}\eta}\left(1+\frac{\alpha}{4\gamma}\eta\right)H_\nu(\phi_{sm+r})+\e^{-\frac{3\alpha}{2\gamma}\eta}\sum_{i=0}^{r-1}\frac{\alpha}{4m\gamma}\eta \e^{-\frac{\alpha m}{\gamma}\eta} H_\nu(\phi_{sm+i})\\
  &+8\eta^2 dL^2\left(2+\Xi +2m\Xi \right).
\end{align*}
\end{proof}
Finally, let us prove Theorem \ref{mth2} and Corollary \ref{mcor21}.
\begin{theoremA}[Theorem \ref{mth2} restated]\label{th4}
  Under Assumptions \ref{as1} and \ref{as2}, $0<\eta<\frac{\alpha}{16\sqrt{2}L^2m\gamma}$ and  $\gamma\ge 1$, for all $k$, the following holds in the update of SARAH-LD:
  \begin{align*}
    H_{\nu}(\phi_{k})&\le \e^{-\frac{\alpha\eta}{\gamma}k}H_\nu(\phi_0)+\frac{32\eta\gamma dL^2}{3\alpha}\left(2+\Xi +2m\Xi \right),
  \end{align*}
where $\Xi =\frac{(n-B)}{B(n-1)}$.
\end{theoremA}
\begin{proof}
Same as Theorem \ref{th2}.\\
\end{proof}
\begin{corollaryA}[Corollary \ref{mcor21} restated]\label{cor41}
  Under the same assumptions as Theorem \ref{th4}, for all $\epsilon\ge 0$, if we choose step size $\eta$ such that
  \[
\eta\le \frac{3\alpha\epsilon}{64\gamma dL^2}\left(2+\Xi +2m\Xi \right)^{-1},
  \]then a precision $H_{\nu}(\phi_{k})\le\epsilon$ is reached after
  \[
  k\ge \frac{\gamma}{\alpha\eta}\log{\frac{2H_{\nu}(\phi_0)}{\epsilon}}
  \]
  steps. Especially, if we take $B=m=\sqrt{n}$ and the largest permissible step size $\eta=\frac{\alpha}{16\sqrt{2}L^2\sqrt{n}\gamma}\wedge \frac{3\alpha\epsilon}{320dL^2\gamma}$, then the gradient complexity becomes
  \[
  \tilde{O}\left(\left(n+\frac{dn^{\frac{1}{2}}}{\epsilon}\right)\cdot \frac{\gamma^2 L^2}{\alpha^2}\right).
  \]
\end{corollaryA}
\begin{proof}
  The first half of the statement is the same as Corollary \ref{cor21}.\par
  When $B\ge m$, from Theorem \ref{th4}, we obtain
   \begin{align*}
      H_{\nu}(\phi_{k})&\le \e^{-\frac{\alpha\eta}{\gamma}k}H_\nu(\phi_0)+\frac{32\eta\gamma dL^2}{3\alpha}\left(2+\Xi +2m\Xi \right)\\
      &\le \e^{-\frac{\alpha\eta}{\gamma}k}H_\nu(\phi_0)+\frac{160\eta\gamma dL^2}{3\alpha}.
    \end{align*}
  Proceeding in the same way as Corollary \ref{cor21}, we obtain the optimal gradient complexity of
  \[
  \tilde{O}\left(\left(n+\frac{dn^{\frac{1}{2}}}{\epsilon}\right)\cdot \frac{\gamma^2 L^2}{\alpha^2}\right)
  \]
  with $B=m=\sqrt{n}$ and $\eta=\frac{\alpha}{16\sqrt{2}L^2\sqrt{n}\gamma}\wedge \frac{3\alpha\epsilon}{320dL^2\gamma}$.
  \par Now, when $B\le m$, from Theorem \ref{th4}, we obtain
  \begin{align*}
     H_{\nu}(\phi_{k})&\le \e^{-\frac{\alpha\eta}{\gamma}k}H_\nu(\phi_0)+\frac{32\eta\gamma dL^2}{3\alpha}\left(2+\Xi +2m\Xi \right)\\
     &\le \e^{-\frac{\alpha\eta}{\gamma}k}H_\nu(\phi_0)+\frac{160\eta\gamma dL^2}{3\alpha}\frac{m}{B}.
   \end{align*}
   This leads to a gradient complexity of
   \[
   \tilde{O}\left(\left(n+\frac{d(m+n/B)}{\epsilon}\right)\cdot \frac{\gamma^2 L^2}{\alpha^2}\right)
   \]
   with $\eta=\frac{\alpha}{16\sqrt{2}L^2m\gamma}\wedge \frac{3\alpha\epsilon}{320dL^2\gamma}\frac{B}{m}$, which is optimal with $B=m=\sqrt{n}$ again. \\
\end{proof}

\section{Proof of Theorem \ref{mth3}, Corollaries \ref{mcor31} and \ref{mcor32}}\label{apc}

We define $X_k$ like Algorithm \hyperlink{al1}{1} in order to simultaneously represent $Y_k$ and $Z_k$.
\subsection{Preparation for the Proof}
\subsubsection{Link between Sampling and Optimization}
Since
 \[
\E_{X_k}[F(X_k)]-F(X^\ast)
\]
can be separated into the discretisation error
\[\E_{X_k}[F(X_k)]-\E_{X\sim\nu }[F(X)]\]
and the approximation error due to sampling
\[\E_{X\sim\nu }[F(X)]-F(X^\ast),\]
in this subsection, we analyse the upper bound of these two terms.
\begin{propA}\label{p3}
  Under Assumption \ref{as1}, the following holds:
  \[
  \forall x\in \mathbb{R}^d,\  F(x)-F(x^\ast)\ge \frac{1}{2L}\|\nabla F(x)\|,
  \]
  where $x^\ast$ is the global minimum of $F$.
\end{propA}
\begin{proof}
  Let us define $G(x)\vcentcolon=F(x)-F(x^\ast)$. Since $G$ is also $L$-smooth,
  \[
  G\left(x-\frac{1}{L}\nabla G(x)\right)\le G(x)-\frac{1}{L}\|\nabla G(x)\|^2+\frac{1}{2L}\|\nabla G(x)\|^2=G(x)-\frac{1}{2L}\|\nabla G(x)\|^2,
  \]
  where for the inequality, we used that the following holds for a $L$-smooth function $H$:
  \[
  \forall x,y\in\mathbb{R}^d,\ H(y)\le H(x)+\left\langle\nabla H(x),y-x\right\rangle+\frac{L}{2}\|y-x\|^2.
  \]
  Now, since $G\ge 0$, we obtain
  \[
    F(x)-F(x^\ast)\ge \frac{1}{2L}\|\nabla F(x)\|,
  \]
  which concludes the proof.\\
\end{proof}
\begin{theoremA}\label{th5}
  Under Assumption \ref{as1}, the following holds for distributions $\rho_k$ and $\nu$:
  \[
  \E_{X_k\sim\rho_k }[F(X_k)]-\E_{X\sim\nu }[F(X)]\le LW_2^2(\rho_k,\nu)+\E_{X\sim\nu }[F(X)]-F(X^\ast),
  \]
  where $X^\ast$ is the global minimum of $F$. The same statement holds with $X_k\sim \phi_k$.
\end{theoremA}
\begin{proof}
  Let $X_k\sim \rho_k$ and $X\sim \nu$ be an optimal coupling so that $\E[\|X_k-X\|^2]=W_2(\rho_k,\nu)^2$. The following holds only from the smoothness of $F$:
  \begin{align*}
    F(X_k)-F(X)&=\int_0^1\langle X_k-X,\nabla F\left((1-t)X+tX_k\right)\rangle \d t\\
    &\le \int_0^1\|X_k-X\|\|\nabla F\left((1-t)X+tX_k\right)\|\d t\\
    &\le \int_0^1\|X_k-X\|\|\nabla F\left((1-t)X+tX_k\right)-\nabla F(X)\|\\
    &\ \ \ \ \ \ \ \ \ \ \ \ \ \ \ \ \ \ \ \ \ \ \ \ \ \ \ \ \ \ \ \ \ \ \ \ \ \ \ \ \ \ +\|X_k-X\|\|\nabla F(X)\|\d t\\
    &\le \int_0^1 Lt\|X_k-X\|^2+\frac{L}{2}\|X_k-X\|^2+\frac{1}{2L}\|\nabla F(X)\|^2 \d t\\
    &\le L\|X_k-X\|^2+F(X)-F(X^\ast).
  \end{align*}
  For the first inequality, we used the Cauchy-Schwarz inequality, for the third inequality we used the smoothness of $F$ on the first term, and for the fourth inequality, Property \ref{p3}.
  \par Hence, taking expectation of both sides, we obtain the desired result.\\
\end{proof}
\begin{corollaryA}\label{cor51}
  Under the same assumptions as Theorem \ref{th5}, the following holds:
  \[
  \E_{X_k\sim\rho_k }[F(X_k)]-F(X^\ast)\le LW_2^2(\rho_k,\nu)+2\left(\E_{X\sim\nu }[F(X)]-F(X^\ast)\right).
  \]
  The same statement holds with $X_k\sim \phi_k$.
\end{corollaryA}
An important feature of this theorem is that the square of the 2-Wasserstein metric appears. Thanks to this and Talagrand's inequality, we can directly use the results from sampling (e.g., Corollaries \ref{mcor11} and \ref{mcor21})
\par The approximation error can be bounded thanks to the following theorem from \citet{RRT2017}.
\begin{theoremA}[\citet{RRT2017}, Proposition 11]\label{th6}
  Under Assumptions \ref{as1} and \ref{as3}, for all $\gamma\ge \frac{2}{M}$
  \[
  \E_{X\sim\nu }[F(X)]-F(X^\ast)\le \frac{d}{2\gamma}\log{\left(\frac{\e L}{M}\left(\frac{b\gamma}{d}+1\right)\right)}.
  \]
\end{theoremA}
\begin{corollaryA}\label{cor61}
  Under the same assumptions as Theorem \ref{th6}, for all $\epsilon>0$, if we additionally require $\gamma\ge \frac{4d}{\epsilon}\log{\left(\frac{\e L}{M}\right)}\vee \frac{8db}{\epsilon^2}$, then
  \[
  \E_{X\sim\nu }[F(X)]-F(X^\ast)\le \frac{\epsilon}{4}.
  \]
\end{corollaryA}
\begin{proof}
  Since
  \[
  \frac{d}{2\gamma}\log{\left(\frac{\e L}{M}\left(\frac{b\gamma}{d}+1\right)\right)}=\frac{d}{2\gamma}\log{\frac{\e L}{M}}+\frac{d}{2\gamma}\log{\left(\frac{b\gamma}{d}+1\right)},
  \]
  it suffices to have $\frac{d}{2\gamma}\log{\frac{\e L}{M}}\le \frac{\epsilon}{8}$ and $\frac{d}{2\gamma}\log{\left(\frac{b\gamma}{d}+1\right)}\le \frac{\epsilon}{8}$. Furthermore, since for all $x\ge 0$
  \[
  \frac{\log{(x+1)}}{x}\le \frac{1}{\sqrt{x+1}}\le \frac{1}{\sqrt{x}}
  \]
  holds, we only need to require
  $\frac{d}{2\gamma}\log{\frac{\e L}{M}}\le \frac{\epsilon}{8}$ and $\frac{b}{2}\frac{1}{\sqrt{b\gamma/d}}\le \frac{\epsilon}{8}$. Solving these two inequalities according to $\gamma$ leads to the desired result.\\
\end{proof}
\begin{remarkA}
  The lower bound $\frac{4d}{\epsilon}\log{\left(\frac{\e L}{M}\right)}\vee \frac{8db}{\epsilon^2}$ is only calculated to acquire a concrete condition on $\gamma$. A more involved analysis could find a better lower bound.
\end{remarkA}
\subsubsection{Explicit Formulation of the Log-Sobolev Constant}
In this subsection, we give an explicit formulation of the Log-Sobolev constant of $\d \nu\propto \e^{-\gamma F}\d x$ in function of $\gamma$ for two cases: under Assumptions \ref{as1} and \ref{as3}, and under Assumptions \ref{as1}, \ref{as3} and \ref{as5} to \ref{as7}. The second case is roughly the first combined with the weak Morse condition.
\par When we only assume dissipativity and smoothness, we can obtain a Log-Sobolev constant whose inverse exponentially depends on the inverse temperature $\gamma$. This employs the following result from \citet{RRT2017}.
\begin{propA}[\citet{RRT2017}, Proposition 9]\label{p01}
  Under Assumptions \ref{as1} and \ref{as3}, for all $\gamma \ge \frac{2}{M}$, $\nu$ satisfies Log-Sobolev inequality with a constant $\alpha$ such that
  \[
  \frac{1}{\alpha}\le \frac{2M^2+2L^2}{M^2L\gamma}+\frac{1}{\lambda_\ast}\left(\frac{6L(d+\gamma)}{M}+2\right),
  \]
  where
  \[
  \frac{1}{\lambda_\ast}\le \frac{1}{M\gamma(d+b\gamma)}+\frac{2C_\ast(d+b\gamma)}{M\gamma}\exp{\left(\frac{2}{M}(L+B_\ast)(b\gamma+d)+\gamma(A_\ast+B_\ast)\right)}.
  \]
  Here, $A_\ast=\max_i\left\{|f_i(0)|\right\}$, $B_\ast=\max_i\left\{|\nabla f_i(0)|\right\}$ and $C_\ast$ is a universal constant that does not depend on $F$.
\end{propA}
From this, we immediately have the following property.
\begin{propA}\label{p1}
  Under Assumptions \ref{as1} and \ref{as3}, for all $\gamma \ge \frac{2}{M}$, we can take a Log-Sobolev constant $\alpha$ of $\nu$ which can be written with constants $C_1$ and $C_2>0$ independent of $\gamma$ as follows:
  \[
  \alpha=\gamma C_1 \e^{-C_2\gamma},
  \]
  where
  \[
  C_1=\left(\frac{2M^2+2L^2}{M^2L}+\left(\frac{6Ld}{M}+2\right)\left(\frac{1}{Md}+\frac{2C_\ast d}{M}\e^{\frac{2d}{M}(L+B_\ast)}\right)\right)^{-1},
  \]
  and
  \[
  C_2=\frac{2b}{M}(L+B_\ast)+(A_\ast+B_\ast)+b+1.
  \]
\end{propA}
\begin{proof}
  From Proposition \ref{p01},
  \[
  \frac{\gamma}{\alpha}\le \frac{2M^2+2L^2}{M^2L}+\frac{\gamma}{\lambda_\ast}\left(\frac{6L(d+\gamma)}{M}+2\right),
  \]
  and
  \[
  \frac{\gamma}{\lambda_\ast}\le \frac{1}{M(d+b\gamma)}+\frac{2C_\ast(d+b\gamma)}{M}\exp{\left(\frac{2}{M}(L+B_\ast)(b\gamma+d)+\gamma(A_\ast+B_\ast)\right)}.
  \]
  Roughly bounding these inequalities, we obtain
  \begin{align*}
    \frac{\gamma}{\lambda_\ast}\le &\  \frac{1}{Md}+\frac{2C_\ast(d+b\gamma)}{M}\e^{\frac{2d}{M}(L+B_\ast)} \e^{\left(\frac{2b}{M}(L+B_\ast)+(A_\ast+B_\ast)\right)\gamma}\\
    \le &\ \frac{1}{Md}+\frac{2C_\ast d}{M}\e^{\frac{2d}{M}(L+B_\ast)} \e^{\left(\frac{2b}{M}(L+B_\ast)+(A_\ast+B_\ast)+b\right)\gamma}\\
    \le &\ \left(\frac{1}{Md}+\frac{2C_\ast d}{M}\e^{\frac{2d}{M}(L+B_\ast)}\right) \e^{\left(\frac{2b}{M}(L+B_\ast)+(A_\ast+B_\ast)+b\right)\gamma},
  \end{align*}
  where for the second inequality we used $d+b\gamma\ge d\e^{b\gamma}$ for all $\gamma>0$ when $d\ge 1$. Thus,
  \begin{align*}
    \frac{\gamma}{\alpha}\le&\  \frac{2M^2+2L^2}{M^2L}+\frac{\gamma}{\lambda_\ast}\left(\frac{6L(d+\gamma)}{M}+2\right)\\
    \le &\  \frac{2M^2+2L^2}{M^2L}\e^{\left(\frac{2b}{M}(L+B_\ast)+(A_\ast+B_\ast)+b+1\right)\gamma}+\frac{\gamma}{\lambda_\ast}\left(\frac{6Ld}{M}+2\right)\e^{\gamma}\\
    \le &\ \left(\frac{2M^2+2L^2}{M^2L}+\left(\frac{6Ld}{M}+2\right)\left(\frac{1}{Md}+\frac{2C_\ast d}{M}\e^{\frac{2d}{M}(L+B_\ast)}\right)\right)\e^{\left(\frac{2b}{M}(L+B_\ast)+(A_\ast+B_\ast)+b+1\right)\gamma}.
  \end{align*}
  Finally, taking into account that a lower bound of a Log-Sobolev constant automatically satisfies the Log-Sobolev inequality, we obtain the desired result.\\
\end{proof}
On the other hand, under the additional condition of weak Morse, Lipschitzness of $\nabla^2 F$ and other minor assumptions, we can obtain a far better Log-Sobolev constant whose inverse depends only linearly on $\gamma$ as follows. This is a straightforward adaptation of Li and Erdogdu's result \citep{LE2020}. We provide a proof in Appendix \ref{apd}.
\begin{propA}\label{p2}
  Under Assumptions \ref{as1}, \ref{as3} and \ref{as5} to \ref{as7}, with $\gamma\ge 1$ such that
  \begin{align*}
    \gamma&\ge  C_\gamma\vcentcolon=\max\left(1,\left(\frac{24dL}{C_F^2}\right)^2\frac{4d{L'}^2}{{\lambda^\dagger}^2},4{L'}^2\left(\frac{24dL}{C_F^2}\right)^6\right),
  \end{align*}
  where $C_F$ is defined in Lemma \ref{la5}, $\nu$ satisfies the Log-Sobolev inequality with constant $\alpha$ such that
  \[
  \frac{1}{\alpha}=\frac{\gamma}{C_3},
  \]
  where
  \[
  C_3\vcentcolon=\left(\frac{2M^2+8L^2}{M^2L}+\left(\frac{6L(d+1))}{M}+2\right)\frac{35}{{\lambda^\dagger}}\right).
  \]
\end{propA}
\subsection{Main Proof}
\begin{theoremA}[Theorem \ref{mth3} restated]\label{th11}
  Using SVRG-LD or SARAH-LD, under Assumptions \ref{as1} to \ref{as3}, $0<\eta<\frac{\alpha}{16\sqrt{6}L^2m\gamma}$, $\gamma\ge \frac{4d}{\epsilon}\log{\left(\frac{\e L}{M}\right)}\vee \frac{8db}{\epsilon^2}\vee 1\vee \frac{2}{M}$ and $B\ge m$, if we take $B=m=\sqrt{n}$ and the largest permissible step size $\eta=\frac{\alpha}{16\sqrt{6}L^2\sqrt{n}\gamma}\wedge \frac{3}{1792}\frac{\alpha^2\epsilon}{L^2d\gamma}$, the gradient complexity to reach a precision of
  \[
  \E_{X_k}[F(X_k)]-F(X^\ast)\le \epsilon
  \]
  is
  \[
  \tilde{O}\left(\left(n+\frac{n^\frac{1}{2}}{\epsilon}\cdot \frac{d L}{\alpha}\right)\frac{\gamma^2 L^2}{\alpha^2}\right),
  \]
  where $\alpha$ is a function of $\gamma$.
\end{theoremA}
\begin{proof}
  It is sufficient to consider the case of SVRG-LD with $X_k\sim\rho_k$. From Corollary \ref{cor51}, the sufficient condition for
  \[
  \E_{X_k}[F(X_k)]-F(X^\ast)\le \epsilon
  \]
  is $LW_2^2(\rho_k,\nu)\le \epsilon/2$ and $\E_{X\sim\nu }[F(X)]-F(X^\ast)\le \epsilon/4$. From Corollary \ref{cor61}, the latter condition is satisfied when $\gamma\ge \frac{4d}{\epsilon}\log{\left(\frac{\e L}{M}\right)}\vee \frac{8bd}{\epsilon^2}\vee 1\vee \frac{2}{M}$. Moreover, concerning the former, from Talagrand's inequality
  \[
  W_2^2(\rho_k,\nu)\le \frac{2}{\alpha}H_\nu(\rho_k),
  \]
  it suffices to have
  \[
  H_\nu(\rho_k)\le \frac{\alpha\epsilon}{4L}.
  \]
  Thus, from Corollaries \ref{cor21} and \ref{cor41}, under the same conditions, we obtain a gradient complexity of
  \[
  \tilde{O}\left(\left(n+\frac{n^{\frac{1}{2}}}{\epsilon}\cdot \frac{d L}{\alpha}\right)\frac{\gamma^2 L^2}{\alpha^2}\right).
  \]
\end{proof}
This leads to the following corollaries.
\begin{corollaryA}[Corollary \ref{mcor31} restated]\label{cor111}
  Under the same assumptions as Theorem \ref{th11}, taking
   \[
  \gamma=i(\epsilon)\vcentcolon=\frac{4d}{\epsilon}\log{\left(\frac{\e L}{M}\right)}\vee \frac{8db}{\epsilon^2}\vee 1\vee \frac{2}{M},
  \]
  we obtain a gradient complexity of
  \[
  \tilde{O}\left(\left(n+\frac{n^\frac{1}{2}}{\epsilon}\cdot \frac{d L}{C_1 i(\epsilon)}\e^{C_2i(\epsilon)}\right) L^2\e^{2C_2i(\epsilon)}\right)
  \]
  since $\alpha=\gamma C_1\e^{-C_2\gamma}$ (Property \ref{p1}).
\end{corollaryA}
\begin{proof}
  The proof follows from Property \ref{p1}.\\
\end{proof}
\begin{corollaryA}[Corollary \ref{mcor32} restated]\label{cor112}
  Under the same assumptions as Theorem 11 and Assumptions \ref{as5} to \ref{as7}, taking
  \[
  \gamma=j(\epsilon)\vcentcolon=\frac{4d}{\epsilon}\log{\left(\frac{\e L}{M}\right)}\vee \frac{8db}{\epsilon^2}\vee 1\vee \frac{2}{M}\vee  C_\gamma,
  \]
  where $C_\gamma$ is a constant independent of $\epsilon$ defined in Property \ref{p2}, we obtain a gradient complexity of
  \[
  \tilde{O}\left(\left(n+\frac{n^\frac{1}{2}}{\epsilon}\cdot \frac{d L}{C_3}j(\epsilon)\right)C_3^2j(\epsilon)^4 L^2\right)
  \]
  since $\alpha=C_3/\gamma$ (Property \ref{p2}).
\end{corollaryA}
\begin{proof}
  The proof follows from Property \ref{p2}.\\
\end{proof}

\section{Proof of Property \ref{p2}}\label{apd}

\subsection{Overview and Main Result}
In this appendix, we prove Property \ref{p2} which is only a slight adaptation of Theorem 3.4 from \citet{LE2020}, which builds its foundation from prior work such as \citet{CGW2010} and \citet{MS2014}. We show that with additional weak Morse, and smoothness assumptions to dissipativity, we can obtain a Log-Sobolev constant of $\d\nu\propto\e^{-\gamma F}\d x$ whose inverse only depends linearly on the inverse temperature parameter. Property \ref{p2} is reminded below in a more precise form.
\begin{theorem}
  Under Assumptions \ref{as1}, \ref{as3} and \ref{as5} to \ref{as7}, with $a>0$ and $\gamma\ge 1$ such that
\[
    a^2\ge \frac{24dL}{C_F^2},
\]
and
\[
\gamma\ge \max\left(1,a^2\frac{4d{L'}^2}{{\lambda^\dagger}^2},4{L'}^2a^6\right),
\]
  $\nu$ satisfies the Log-Sobolev inequality with constant $\alpha$ such that
  \[
  \frac{1}{\alpha}=\left(\frac{2M^2+8L^2}{M^2L}+\left(\frac{6L(d+1))}{M}+2\right)\frac{35}{{\lambda^\dagger}}\right)\gamma.
  \]
\end{theorem}
This theorem shows that the strict saddle node assumption is almost sufficient to obtain in the Euclidean space for dissipative distributions a Log-Sobolev constant whose inverse does not exponentially depend on the inverse temperature, which was the case without this assumption.
\subsection{Preliminaries}
We first clarify some definitions.
\begin{definitionA}
  We say a probability measure $\nu$ satisfies the Poincaré inequality with a constant $\kappa$ if for all smooth $g:\mathbb{R}^d\to \mathbb{R}$,
  \[
  \E_\nu[g^2]-\E_\nu[g]^2\le \frac{1}{\kappa}\E_\nu[\|\nabla g\|^2].
  \]
\end{definitionA}
\begin{definitionA}
  A probability measure $\nu$ on $\mathbb{R}^d$ restricted on a set $\mathcal{Z}\subset\mathbb{R}^d$ is defined as
  \[
  \nu|_\mathcal{Z}\vcentcolon=\frac{\nu(x)}{\int_\mathcal{Z}\nu(y)dy}\mathds{1}_\mathcal{Z}(x).
  \]
\end{definitionA}
\begin{definitionA}
  We define the following sets:
  \begin{align*}
    \mathcal{B}\vcentcolon =&\left\{x\in\mathbb{R}^d\mid\ d(x,\mathcal{S})^2< \frac{a^2}{\gamma}\right\},\\
    \mathcal{U}\vcentcolon =&\left\{x\in\mathbb{R}^d\mid\ d(x,\mathcal{X})^2<\frac{a^2}{\gamma}\right\},\\
    \mathcal{A}\vcentcolon =& \left\{x\in\mathbb{R}^d\mid\ d(x,\mathcal{S}\cup\mathcal{X})^2\ge \frac{a^2}{4\gamma}\right\},
  \end{align*}
  where $\mathcal{X}$ is the set of global minima and $\mathcal{S}$ is the set of stationary points except the global minima. Note that $\mathcal{B}\cup \mathcal{U}\cup\mathcal{A}=\mathbb{R}^d$.
  \par Here, the distance from a point $x\in\mathbb{R}^d$ and a set $\mathcal{Z}\subset\mathbb{R}^d$ is defined as
  \[
  d(x,\mathcal{Z})\vcentcolon=\inf_{z\in\mathcal{Z}}\|x-z\|.
  \]
\end{definitionA}
In this appendix, we only consider the following generator $\mathcal{L}$.
\begin{definitionA}\label{da3}
  We define $\mathcal{L}$ such that
  \[
  \mathcal{L}f\vcentcolon=\langle-\nabla f, \nabla F\rangle+\frac{1}{\gamma}\Delta f,\ \ \forall f\in C^2(\mathbb{R}^d)
  \]
  which is the generator of the gradient Langevin Dynamics \eqref{eq21}.
\end{definitionA}
We will need some lemmas proved by \citet{LE2020}.
\begin{lemmaA}
  [\citet{LE2020}]\label{la1} Under Assumptions \ref{as1} and \ref{as6}, suppose $y\in\mathbb{R}^d$ is a stationary point of $F$. Then, with
  \[
  H(x)\vcentcolon=\nabla^2 F(0)\cdot x
  \]
defined in the coordinate centered at $y$, we obtain for all $x\in\mathbb{R}^d$,
\[
\|\nabla F-H(x)\|\le L'\|x\|^2.
\]
\end{lemmaA}
\begin{proof}
  From the mean value theorem, there exist a $\hat{x}$ on the line between 0 and $x$ such that
  \[
  \nabla F(x)=\nabla F(0)+\nabla^2 F(\hat{x})\cdot x=\nabla^2 F(\hat{x})\cdot x,
  \]
  where for the last equality, we used that $0$ was a stationary point.
  Therefore, we obtain
  \begin{align*}
    \|\nabla F(x)-H(x)\|=&\ \|\nabla^2 F(\hat{x})\cdot x-\nabla^2 F(0)\cdot x\|\\
    \le&\ \|\nabla^2 F(\hat{x})-\nabla^2 F(0)\|\|x\|\\
    \le&\ L'\|\hat{x}\|\|x\|\\
    \le&\ L'\|x\|^2,
  \end{align*}
where for the second inequality we used the $L'$-Lipschitzness of $\nabla^2 F$, and for the last inequality we used $\|\hat{x}\|\le \|x\|$.\\
\end{proof}
\begin{lemmaA}[\citet{LE2020}, Proposition E.5]\label{la2}
   Let $W_t$ and $\tilde{W}_t$ be weak solutions on some filtered probability space of the following one dimensional SDE's:
   \begin{align*}
     \d W_t&= \Phi(W_t)\d t+\sigma \d B_t,\\
     \d \tilde{W}_t&= \tilde{\Phi}(\tilde{W}_t)\d t+\sigma \d B_t,
   \end{align*}
   where $W_0=\tilde{W}_0$ a.s. and $\sigma>0$ is a constant. We further assume that for all $T\ge 0$,
   \[
   \int_0^T|\Phi (W_t)|+|\tilde{\Phi}(\tilde{W}_t)|\d t<\infty,\ a.s.
   \]
   If $\Phi (W_t)\ge \tilde{\Phi}(\tilde{W}_t)$ for all $x\in \mathbb{R}$, then $W_t\ge \tilde{W}_t$ a.s.
\end{lemmaA}
\begin{lemmaA}[\citet{LE2020}, Corollary D.6]\label{la3}
  Consider the following Cox-Ingersoll-Ross process defined as
  \[
  \d W_t=\left(2{\lambda^\dagger} W_t+\frac{1}{2\gamma}\right)\d t+\frac{2}{\sqrt{\gamma}}\sqrt{W_t}\d B_t,\ \ W_0=w_0\ge 0,
  \]
  where ${\lambda^\dagger}>0$, $\gamma>0$ and $\{B_t\}_{t\ge0}$ is a standard one dimensional Brownian motion. Then for its unique strong solution $W_t$, we have the following density function:
  \[
  f(w;t)=2^{-\frac{5}{4}}\left(\frac{w}{w_0}\right)^{-\frac{1}{4}}\frac{{\lambda^\dagger} \gamma}{\e^{\frac{{\lambda^\dagger} t}{2}}\sinh({\lambda^\dagger} t)}\exp\left(\frac{{\lambda^\dagger}\gamma(w\e^{-2{\lambda^\dagger} t}-\frac{w_0}{2})}{1-\e^{-2{\lambda^\dagger} t}}\right)I_{-\frac{1}{2}}\left(\frac{{\lambda^\dagger}\gamma}{\sinh({\lambda^\dagger} t)}\sqrt{\frac{ww_0}{2}}\right)
  \]
  for $w>0$ and $f(w;t)=0$ for $w=0$, where $I_{-\frac{1}{2}}$ is the modified Bessel function of the first kind of degree $-\frac{1}{2}$. Thus, $W_t>0$ a.s.
\end{lemmaA}
\begin{lemmaA}[\citet{LE2020}, Lemma C.7]\label{la4}
  For the density function $f(w;t)$ defined in Lemma \ref{la3}, we have for $w\le R$ and $t\ge 0$,
  \[
  f(w;t)\le C\e^{-2{\lambda^\dagger} t},
  \]
  where $C\vcentcolon=C(R,{\lambda^\dagger},\gamma)>0$ is a constant independent of $t$ and $w_0$.
\end{lemmaA}
Finally, the next two theorems will be highly useful to establish the Poincaré inequality with an explicit constant.
\begin{theoremA}[\citet{BBCG2008}, Theorem 1.4 adapted]\label{tha1}
  Suppose $\nu|_\mathcal{Z}$ ($\mathcal{Z}\subset\mathbb{R}^d$) satisfies the Poincaré inequality with constant $\kappa_\mathcal{Z}$ and there exists a Lyapunov function $V\in C^2(\mathcal{Z}')$, where $\mathcal{Z}\subset\mathcal{Z}'$. That is, $V\ge 1$ and there exist constants $\theta>0$ and $b\ge 0$ such that
  \[
  \mathcal{L}V=\langle -\nabla F,\nabla V\rangle+ \frac{1}{\gamma}\Delta V\le -\theta V+b\mathds{1}_\mathcal{Z}.
  \]
  Then $\nu|_{\mathcal{Z}'}$ satisfies the Poincaré inequality with constant
  \[
  \kappa=\frac{\theta}{1+b/\kappa_\mathcal{Z}}.
  \]
\end{theoremA}
\begin{theoremA}[\citet{LE2020}, Lemma B.14 adapted]\label{tha1bis}
  Set the following neighbourhood of saddle points
  \begin{align*}
    \mathcal{B}_r&=\{x\in \mathbb{R}^d\mid d(x,\mathcal{S})<r\}.
  \end{align*}
  Let $r>\tilde{r}>0$ and suppose $\nu\mid_{\mathcal{B}_{\tilde{r}}}$ satisfies the Poincaré inequality with constant $\tilde{\kappa}$ and there exist a Lyapunov function $1\le V\in C^2(\mathcal{B})$ and a constant $\theta>0$ such that
  \[
  \mathcal{L}V\le -\theta V.
  \]
  Then, $\nu$ satisfies the Poincaré inequality with constant $\kappa$ such that
  \[
  \frac{1}{\kappa}=\frac{4}{\theta}+\left(\frac{4}{\theta\gamma(r-\tilde{r})^2}+2\right)\frac{1}{\tilde\kappa}.
  \]
\end{theoremA}
From these theorems, we will be able to find a Poincaré constant for $\nu$ consecutively from $\mathcal{U}$ to $\mathcal{U}\cup\mathcal{A}$ and then to $\mathbb{R}^d$.
\subsection{Lyapunov Function for $\mathcal{B}$}
The following theorem gives a sufficient condition to find a Lyapunov function for $\mathcal{B}$ in the sense of Theorem \ref{tha1bis}. This is actually a combination of Bovier and Den Hollander's Theorem 7.15 \citep{BD2016} and Wainwright's Theorem 2.13 \citep{W2019}. See \citet{LE2020} for details.
\begin{theoremA}[\citet{LE2020}, Proposition 9.5]\label{tha2}
  If there exist constants $c_1>0$ and $c_2>0$ such that
  \[
  P\left(\tau_{\mathcal{B}^c}\ge t\mid X_0=x\right)\le c_1\e^{-c_2 t},\ \ \forall t\ge 0,\forall x\in \mathcal{B},
  \]
  where $\tau_{\mathcal{B}^c}\vcentcolon=\inf\{t\ge 0\mid X_t\notin \mathcal{B} \}$, then by defining $V(x)\vcentcolon=\E[\e^{c_2\tau_{\mathcal{B}^c}/2}\mid X_0=x]$, the following holds:
  \[
  \mathcal{L}V\le -\frac{c_2}{2}V.
  \]
\end{theoremA}
\par It is thus enough to establish an exponentially decaying tail bound for $\tau_{\mathcal{B}^c}$ as shown in the next theorem.
\begin{theoremA}[\citet{LE2020}, Proposition 9.6 adapted]\label{tha3}
  Let $\{X_t\}_{t\ge0}$ be the Langevin diffusion defined in \eqref{eq21}. Under Assumptions \ref{as1}, \ref{as5} and \ref{as6}, with $a>0$ and $\gamma>0$ such that
  \[
  \gamma\ge \max\left(a^2,4{L'}^2a^6\right),
  \]
  the following holds:
  \[
  P\left(\tau_{\mathcal{B}^c}\ge t\mid X_0=x\right)\le c_1\e^{-{\lambda^\dagger} t/2},\ \ \forall t\ge 0,\forall x\in \mathcal{B},
  \]
  where $c_1\vcentcolon=c_1(a,\gamma,{\lambda^\dagger})$ is a constant independent of $t$ and $x$. Hence, $V(x)\vcentcolon=\E[\e^{\theta\tau_{\mathcal{B}^c}}\mid X_0=x]$ is a Lyapunov function on $\mathcal{B}$ in the sense of Theorem \ref{tha1bis} with parameter $\theta=\frac{{\lambda^\dagger}}{4}$.
\end{theoremA}
\begin{proof}
  For each $y\in \mathcal{S}$, we define $v_y$ as the unit eigenvector of $\nabla^2 F(y)$ that corresponds to the minimum eigenvalue of $\nabla^2 F(y)$. From Assumption \ref{as5}, we have that $\left\langle v_y,\nabla^2 F(y)v_y\right\rangle\le -{\lambda^\dagger}$. Now, let us fix a $x\in\mathcal{B}$ and take a $y\in \mathcal{S}$ such that $\|x-y\|^2< \frac{a^2}{\gamma}$. In the remainder of this proof, we will work in coordinates centered at this $y$. Without loss of generality, we can thus set $y=0$.
  \par Let $r(x)\vcentcolon=\|x\|$. Then
\[
\nabla r(x)=\frac{x}{\|x\|}.
\]
\par We also define $P:\mathcal{B}\to \tilde{\mathcal{B}}$ where $\tilde{\mathcal{B}}\vcentcolon=\left\{tv_0\mid  |t|< \frac{a^2}{\gamma}\right\}$ such that
\[
Px\vcentcolon=\langle v_0,x\rangle v_0,
\]
and $\tilde{r}(x)\vcentcolon=|\langle v_0,x\rangle|$.\footnote{Note that, $\tilde{r}(x)$ is not differentiable for $x$ such that $\tilde{r}(x)=0$. For these points, we redefine $\nabla\tilde{r}(x)$ and $\Delta\tilde{r}(x)$ to be some constant $C_r>0$, but this case can be ignored as shown later.} As a result,
\[
\nabla \tilde{r}(x)=\frac{Px}{\|Px\|}=\mathrm{sign}(\langle v_0,x\rangle)v_0.
\]
\par Using Itô's formula, we obtain
\begin{align*}
  \d\left(\frac{1}{2}\tilde{r}(X_t)^2\right)=&\ \left(\langle-\nabla F(X_t),\tilde{r}(X_t)\nabla\tilde{r}(X_t)\rangle+\frac{1}{\gamma}\left(\|\nabla \tilde{r}(X_t)\|^2+\tilde{r}(X_t)\Delta\tilde{r}(X_t)\right)\right)\d t\\
  &+\frac{2}{\gamma}\langle \tilde{r}(X_t)\nabla \tilde{r}(X_t),\d W_t\rangle.
\end{align*}
Since $\nabla\tilde{r}(x)$ is a unit vector, we can consider $\langle \nabla \tilde{r}(X_t),dW_t\rangle$ as a standard one-dimensional Brownian motion independent of $X_t$ that we denote as $\d B_t$.
\par Next, considering
\[
H(x)\vcentcolon=\nabla^2 F(0)\cdot x,
\]
it follows from Lemma \ref{la1} that
\[
\|\nabla F(x)-H(x)\|\le L'\|x\|^2.
\]
\par Therefore,
\begin{align*}
  \langle -\nabla F(X_t),\nabla \tilde{r}(X_t) \rangle= &\ \langle -H(X_t),\nabla \tilde{r}(X_t) \rangle -\langle F(X_t)-\nabla H(X_t),\nabla \tilde{r}(X_t) \rangle\\
  \ge &\  \langle -H(X_t),\nabla \tilde{r}(X_t) \rangle -\|H(X_t)-\nabla F(X_t)\|\|\nabla \tilde{r}(X_t)\|\\
  \ge &\  \langle -H(X_t),\nabla \tilde{r}(X_t) \rangle -\|H(X_t)-\nabla F(X_t)\|\\
  \ge &\  \langle -H(X_t),\nabla \tilde{r}(X_t) \rangle -L'\|X_t\|^2,
\end{align*}
where for the second inequality, we used that $\nabla\tilde{r}(x)$ is a unit vector.
\par Using the definition of $H(x)$, we can further write
\[
\langle -H(x),\nabla\tilde{r}(x)\rangle=-\left\langle \nabla^2 F(0)\cdot x,\frac{Px}{\|Px\|}\right\rangle\ge {\lambda^\dagger}\tilde{r}(x),
\]
where for the inequality, we used that $v_0$ is an eigenvector of  $\nabla^2 F(0)$ and Assumption \ref{as5}.
\par Since $\Delta \tilde{r}(x)\ge 0$ we have that $\|\nabla\tilde{r}(x)\|^2+\tilde{r}(x)\Delta\tilde{r}(x)\ge 1.$
Therefore, from Lemma \ref{la2}, $\frac{1}{2}\tilde{r}(X_t)^2$ is lower bounded by the stochastic process $\frac{1}{2}\left(r_t^{(1)}\right)^2$ defined as
\begin{align*}
  \d\left(\frac{1}{2}\left(r_t^{(1)}\right)^2\right)=&\ \left( {\lambda^\dagger}\left(r_t^{(1)}\right)^2-L'\|X_t\|^2r_t^{(1)}+\frac{1}{\gamma}\right)\d t+\sqrt{\frac{2}{\gamma}}r_t^{(1)}\d B_t.
\end{align*}
Since we are only concerned with $X_t\in \mathcal{B}$, the following holds:
 \[
 \|X_t\|^2\le \frac{a^2}{\gamma},
 \]
  and
 \[
 r_t^{(1)}\le \tilde{r}(X_t)\le\sqrt{ \frac{a^2}{\gamma}}.
 \]
 We can again use Lemma \ref{la2} to obtain a lower bound of $r_t^{(1)}$ defined as
\begin{align*}
  \d\left(\frac{1}{2}\left(r_t^{(2)}\right)^2\right)=&\ \left( {\lambda^\dagger}\left(r_t^{(2)}\right)^2-L'\frac{a^3}{\gamma^{3/2}}+\frac{1}{\gamma}\right)\d t+\sqrt{\frac{2}{\gamma}}r_t^{(2)}\d B_t\\
  =&\ \left( {\lambda^\dagger}\left(r_t^{(2)}\right)^2+\frac{1}{\gamma}\left(1-L'\frac{a^3}{\gamma^{1/2}}\right)\right)\d t+\sqrt{\frac{2}{\gamma}}r_t^{(2)}\d B_t.
\end{align*}
Since $\gamma\ge 4L'^2a^6$, $1-L'\frac{a^3}{\gamma^{1/2}}\ge \frac{1}{2}$. This gives us a further lower bound $r_t^{(3)}$ defined as
\begin{align*}
  \d\left(\frac{1}{2}\left(r_t^{(3)}\right)^2\right)=&\ \left( 2{\lambda^\dagger}\frac{1}{2}\left(r_t^{(3)}\right)^2+\frac{1}{2\gamma}\right)\d t+\sqrt{\frac{2}{\gamma}}r_t^{(3)}\d B_t.
\end{align*}
This is a Cox-Ingersoll-Ross process with $W_t=\frac{1}{2}\left(r_t^{(3)}\right)^2$. Consequently, from Lemmas \ref{la3} and \ref{la4}, when $x\le\frac{a^2}{2\gamma}$, the density function $f(w;t)$ satisfies
\begin{align*}
f(w;t)\le C\e^{-{\lambda^\dagger} t/2},
\end{align*}
for all $t\ge 0$ and some constant $C\vcentcolon=C(a,\gamma,{\lambda^\dagger})$ independent of $t$ and $x$. Furthermore, since $\tilde{r}(X_t)$ is lower-bounded by $r_t^{(3)}$ which is almost surely positive by Lemma \ref{la3}, the case $\tilde{r}(X_t)=0$ can be ignored.
\par As a result, we obtain
\begin{align*}
  P\left(\tau_{\mathcal{B}^c}\ge t\mid X_0=x\right)= &\ P\left(\sup_{s\in[0,t]}\frac{1}{2}r(X_s)^2\le \frac{a^2}{2\gamma}\mid X_0=x\right)\\
  \le&\ P\left(\frac{1}{2}r(X_t)^2\le \frac{a^2}{2\gamma}\mid X_0=x\right)\\
  \le&\ P\left(\frac{1}{2}\left(r_t^{(3)}\right)^2\le \frac{a^2}{2\gamma}\mid r_0^{(3)}=\tilde{r}(x)\right)\\
  =&\ \int_0^{\frac{a^2}{2\gamma}}f(w;t)dw\\
  \le&\ \frac{a^2}{2\gamma}C\e^{-{\lambda^\dagger} t/2},
\end{align*}
where for the second equality, we used Lemma \ref{la3} and for the last inequality we used Lemma \ref{la4}. This gives the desired result.\\
\end{proof}
\subsection{Lyapunov Function for $\mathcal{A}$}
In this section, we prepare statements for the Lyapunov function on $\mathcal{A}$.
\begin{lemmaA}[\citet{LE2020}, Lemma 9.7 adapted]\label{la5}
  Under Assumptions \ref{as1},\ref{as3}, \ref{as5} and \ref{as6}, with $\mathcal{C}\vcentcolon=\mathcal{S}\cup\mathcal{X}$, there exists a constant $0<C_F\le 1$ such that
  \[
  \|\nabla F(x)\|\ge C_Fd(x,\mathcal{C}),
  \]
  where
  \[
  C_F\vcentcolon=\min\left(1,\frac{{\lambda^\dagger}}{2},\inf_{x:d(x,\mathcal{C})>\frac{{\lambda^\dagger}}{4dL'}}\frac{\|\nabla F(x)\|}{d(x,\mathcal{C})}\right).
  \]
\end{lemmaA}
\begin{proof}
  First, observe that when $F$ is $(M,b)$-dissipative, we have
  \begin{align*}
    \frac{1}{M}\|\nabla F(x)\|^2+\frac{M}{2}\|x\|^2\ge M\|x\|^2-b,
  \end{align*}
  which leads to
  \[
  \frac{\|\nabla F(x)\|}{\|x\|}\ge \sqrt{\frac{M^2}{2}-Mb/\|x\|^2}.
  \]
  We obtain thus
  \begin{align}
    \liminf_{\|x\|\to\infty}\frac{\|\nabla F(x)\|}{\|x\|}\ge \sqrt{\frac{M^2}{2}}.\label{eqd1}
  \end{align}
  Therefore,
  \[
  \inf_{x:d(x,\mathcal{C})>\frac{{\lambda^\dagger}}{4dL'}}\frac{\|\nabla F(x)\|}{d(x,\mathcal{C})}>0,
  \]
  since by equation \eqref{eqd1} $\|\nabla F(x)\|/d(x,\mathcal{C})>0$ holds outside a compact set of $x$ around the origin and since in this compact set and away from stationary points there exist an $x$ that minimizes $\|\nabla F(x)\|/d(x,\mathcal{C})$ and that cannot be 0 as we are outside $\mathcal{C}$. As a result, we just have to consider the case when $d(x,\mathcal{C})\le \frac{{\lambda^\dagger}}{4dL'}$.
  \par Let $y$ be a stationary point such that $\|x-y\|<2d(x,\mathcal{C})\le \frac{{\lambda^\dagger}}{2dL'}$. Since $\lambda_i$ is $d$-Lipschitz, we have that
  \[
  |\lambda_i\left(\nabla^2 F(x)\right)-\lambda_i\left(\nabla^2 F(y)\right)|\le d|\nabla^2 F(x)-\nabla^2 F(y)|\le dL'\|x-y\|\le \frac{{\lambda^\dagger}}{2}.
  \]
  Thus, we obtain
  \[
  |\lambda_i\left(\nabla^2 F(x)\right)|\ge \frac{\lambda^\dagger}{2}.
  \]
  \par Now, let us define $\gamma(t)\vcentcolon=tx+(1-t)y\ (t\in[0,1])$. We obtain
  \begin{align*}
    \left\|\frac{\d }{\d t}\nabla F(\gamma(t))\right\|=\left\|\nabla^2 F(\gamma (t))\cdot (x-y)\right\|\ge \min_{i\in \{1,\ldots,d\}}|\lambda_i\left(\nabla^2 F(x)\right)|\|x-y\|\ge \frac{{\lambda^\dagger}}{2}\|x-y\|.
  \end{align*}
  \par We obtain thus the desired result with $C_F>0$ by taking the minimum of the two constants.\\
\end{proof}
\begin{lemmaA}[\citet{LE2020}, Lemma 9.8 adapted]\label{la6}
  Under Assumptions \ref{as1}, \ref{as3}, \ref{as5} and \ref{as6}, for $a>0$ and $\gamma>0$ such that
  \[
  a^2\ge \frac{24dL}{C_F^2},
  \]
  the following holds:
  \[
  \frac{\Delta F(x)}{2}-\frac{\gamma}{4}\|\nabla F(x)\|^2\le -dL,\ \forall x\in \mathbb{R}^d: d(x,\mathcal{C})\ge \frac{a^2}{4\gamma}.
  \]
\end{lemmaA}
\begin{proof}
  From Lemma \ref{la5} and smoothness of $F$, we have
  \[
  \frac{\Delta F(x)}{2}-\frac{\gamma}{4}\|\nabla F(x)\|^2\le \frac{dL}{2}-\frac{\gamma}{4}C_F^2d(x,\mathcal{C}).
  \]
  \par Since $d(x,\mathcal{C})\ge \frac{a^2}{\gamma}$ and $a^2\ge \frac{24dL}{C_F^2}$, we obtain
  \begin{align*}
    \frac{\Delta F(x)}{2}-\frac{\gamma}{4}\|\nabla F(x)\|^2\le&\  \frac{dL}{2}-\frac{\gamma}{4}C_F^2d(x,\mathcal{C})\\
    \le &\ \frac{dL}{2}-\frac{\gamma}{4}C_F^2\frac{a^2}{\gamma}\\
    \le &\ -dL.
  \end{align*}
\end{proof}
\subsection{Poincaré Inequality}
In this section, we establish the Poincaré inequality for $\nu$ using Theorem \ref{tha1} and \ref{tha1bis}. It is easy to find a Poincaré constant for $\nu|_{\mathcal{U}}$ which is our starting point.
\begin{lemmaA}[\citet{LE2020}, Lemma 9.9 adapted]\label{la7}
  Under Assumptions \ref{as1}, \ref{as3} and \ref{as5} to \ref{as7}, with $a>0$ and $\gamma>0$ such that
  \[
  \gamma\ge a^2\frac{4d{L'}^2}{{\lambda^\dagger}^2},
  \]
  $\nu|_\mathcal{U}$ satisfies the Poincaré inequality with constant $\kappa_\mathcal{U}=\frac{{\lambda^\dagger}}{2}$.
\end{lemmaA}
\begin{proof}
  Let $x^\ast$ be the global minimum. Then $\mathcal{U}=\{x\in\mathbb{R}^d\mid \|x-x^\ast\|^2< \frac{a^2}{\gamma}\}$. Using the same idea as Lemma \ref{la5}, we have
  \[
\min_{i\in\{1,\ldots,n\}}\lambda_i\left(\nabla^2 F(x)\right)\ge {\lambda^\dagger}-dL'\|x-x^\ast\|\ge \frac{\lambda^\dagger}{2},
  \]
  where we used  $\gamma\ge a^2\frac{4d{L'}^2}{{\lambda^\dagger}^2}$ and $\|x-x^\ast\|\le \sqrt{{a^2}{\gamma}}$.
  \par This implies for all $x\in \mathcal{U}$,
  \[
  \nabla^2 F(x)\ge \frac{{\lambda^\dagger}}{2}I_{d\times d},
  \]
  where $I_{d\times d}$ is the $d\times d$ unit matrix. Therefore, $\nu|_\mathcal{U}$ satisfies the Poincaré inequality with constant $\kappa_\mathcal{U}=\frac{{\lambda^\dagger}}{2}$.\\
\end{proof}
Next, we show that $\nu|_{\mathcal{U}\cup\mathcal{A}}$ satisfies the Poincaré inequality.
\begin{lemmaA}[\citet{LE2020}, Lemma 9.11 adapted]\label{la8}
  Under Assumptions \ref{as1}, \ref{as3} and \ref{as5} to \ref{as7}, with $a>0$ and $\gamma>0$ such that
  \[
    a^2\ge \frac{24dL}{C_F^2},
  \]
  and
  \[
    \gamma\ge a^2\frac{4d{L'}^2}{{\lambda^\dagger}^2},
  \]
 $\nu|_{\mathcal{U}\cup \mathcal{A}}$ satisfies the Poincaré inequality with constant
  \[
  \kappa_{\mathcal{U}\cup \mathcal{A}}=\frac{1}{1+3/(2\kappa_\mathcal{U})}.
  \]
\end{lemmaA}
\begin{proof}
  Let us choose the candidate Lyapunov function $V_1(x)=\e^{\frac{\gamma}{2}F(x)}$. Then,
  \[
  \frac{\mathcal{L}V_1}{V_1}=\frac{1}{2}\Delta F-\frac{\gamma}{4}\|\nabla F\|^2.
  \]
  From Lemma \ref{la6}, for all $x\in \mathcal{A}$ we have
  \[
  \frac{\mathcal{L}V_1}{V_1}\le -dL.
  \]
  On the other hand, for all $x\in \mathcal{U}$ we obtain
  \[
  \frac{\mathcal{L}V_1}{V_1}=\frac{1}{2}\Delta F\le \frac{1}{2}dL.
  \]
  This leads to
  \[
    \frac{\mathcal{L}V_1}{V_1}\le -dL+\frac{3dL}{2}\mathds{1}_\mathcal{U}
  \]
  for all $x\in \mathcal{U}\cup \mathcal{A}$. Since the assumptions of Theorem \ref{tha1} are satisfied, we conclude that $\nu|_{\mathcal{U}\cup \mathcal{A}}$ satisfies the Poincaré inequality with a constant
  \[
  \kappa_{\mathcal{U}\cup \mathcal{A}}=\frac{dL}{1+3dL/(2\kappa_\mathcal{U})}.
  \]
  Since $L\ge 1$ and $d\ge 1$, we can replace this value by its lower bound
  \[
  \kappa_{\mathcal{U}\cup \mathcal{A}}=\frac{1}{1+3/(2\kappa_\mathcal{U})}.
  \]
\end{proof}
Finally, we can establish the Poincaré inequality for $\nu$.
\begin{lemmaA}[\citet{LE2020}, Proposition 9.12 adapted]\label{la9}
  Under Assumptions \ref{as1}, \ref{as3} and \ref{as5} to \ref{as7}, with $a>0$ and $\gamma>0$ such that
  \[
    a^2\ge \frac{24dL}{C_F^2},
    \]
    and
    \[
    \gamma\ge \max\left(a^2\frac{4d{L'}^2}{{\lambda^\dagger}^2},4{L'}^2a^6\right),
    \]
  $\nu$ satisfies the Poincaré inequality with constant
  \[
  \kappa=\frac{{\lambda^\dagger}}{35}.
  \]
\end{lemmaA}
\begin{proof}
  Let us select the candidate Lyapunov function
  \[
  V_2(x)=\E[\e^{{\lambda^\dagger}\tau_{\mathcal{B}^c}/4}|X_0=x],
  \]
  where $\tau_{\mathcal{B}^c}=\inf\{t\ge 0|X_t\notin \mathcal{B} \}$.
  \par From Theorem \ref{tha2} and  \ref{tha3}, we have that $V_2$ satisfies the Lyapunov condition with
  \[
  \frac{\mathcal{L}V_2}{V_2}\le -\frac{{\lambda^\dagger}}{4}.
  \]
  Now, using Theorem \ref{tha1bis} with $\theta=\frac{{\lambda^\dagger}}{4}$, $\tilde{r}=\frac{1}{2}r=\frac{a}{2\sqrt{\gamma}}$, we conclude that $\nu$ satisfies the Poincaré inequality with a constant $\kappa$ such that
  \[
  \frac{1}{\kappa}=\frac{16}{\lambda^\dagger}+\left(\frac{16}{\lambda^\dagger\gamma}\frac{4\gamma}{a^2}+2\right)\frac{1}{\kappa_{\mathcal{U}\cup \mathcal{A}}}.
  \]
  \par Since $C_F\le 1$, $d\ge 1$ and $L\ge 1$, we have $\frac{1}{a^2}\le \frac{C_F^2}{24dL}\le \frac{1}{24}$, which leads to
  \[
  \frac{1}{\kappa}\le\frac{16}{\lambda^\dagger}+\left(\frac{8}{3\lambda^\dagger}+2\right)\frac{1}{\kappa_{\mathcal{U}\cup \mathcal{A}}}.
  \]
  Plugging $1/\kappa_{\mathcal{U}\cup \mathcal{A}}=1+3/(2\kappa_{\mathcal{U}})=1+3/\lambda^\dagger$ from Lemma \ref{la7} and \ref{la8}, we obtain
  \begin{align*}
    \frac{1}{\kappa}&\le\frac{16}{\lambda^\dagger}+\left(\frac{8}{3\lambda^\dagger}+2\right)\left(1+\frac{3}{\lambda^\dagger}\right)\\
    &\le \frac{35}{\lambda^\dagger},
  \end{align*}
  where we used $\frac{1}{\lambda^\dagger}\ge 1$ in the last inequality.
\end{proof}
\subsection{Log-Sobolev Inequality}
Finally, we can establish the Log-Sobolev inequality thanks to the following theorem.
\begin{theoremA}[\citet{CGW2010}]\label{tha4}
  Suppose the following conditions hold for the generator defined in Definition \ref{da3}.
  \begin{enumerate}
    \item There exist constants $\theta>0$ and $b>0$ and a $C^2$ function $V:\mathbb{R}^d\to [1,\infty)$ such that for all $x\in\mathbb{R}^d$
    \[
    \frac{\gamma\mathcal{L}V(x)}{V(x)}\le -\theta\|x\|^2+b.
    \]
    \item $\nu$ satisfies the Poincaré inequality with a constant $\kappa$.
    \item There exists some constant $K> 0$, such that $\nabla^2 F\succeq -LI_{d\times d}$.
  \end{enumerate}
  Then, $\nu$ satisfies the Log-Sobolev inequality with a constant $\alpha$ such that
  \[
  \frac{1}{\alpha}=C_1+(C_2+2)\frac{1}{\kappa},
  \]
  where
  \[
  C_1=\frac{2\gamma L}{\theta}+\frac{2}{\gamma K},
  \]
  and\[
  C_2=\frac{2\gamma L}{\theta}\left(b+\theta\int_{\mathbb{R}^d}\|x\|^2\d\nu\right).
  \]
\end{theoremA}
\begin{theoremA}\label{tha5}
  Under Assumptions \ref{as1}, \ref{as3} and \ref{as5} to \ref{as7}, with $a>0$ and $\gamma\ge 1$ such that
  \[
    a^2\ge \frac{24dL}{C_F^2},
    \]
    and
    \[
    \gamma\ge \max\left(1,a^2\frac{4d{L'}^2}{{\lambda^\dagger}^2},4{L'}^2a^6\right),
    \]
  $\nu$ satisfies the Log-Sobolev inequality with constant $\alpha$ such that
  \[
  \frac{1}{\alpha}=\left(\frac{2M^2+8L^2}{M^2L}+\left(\frac{6L(d+1))}{M}+2\right)\frac{35}{{\lambda^\dagger}}\right)\gamma.
  \]
\end{theoremA}
\begin{proof}
  Let us consider the candidate Lyapunov function $V(x)=\e^{M\gamma\|x\|^2/4}$. Then from $V\ge 1$ and Assumption \ref{as3}, we obtain
  \begin{align*}
    \gamma\mathcal{L}V(x)=&\ \left(\frac{M\gamma d}{2}+\frac{M^2\gamma^2}{4}\|x\|^2-\frac{M\gamma^2}{2}\langle x, \nabla F(x)\rangle\right)V(x)\\
    \le &\ \left(\frac{M\gamma(d+b\gamma)}{2}-\frac{M^2\gamma^2}{4}\|x\|^2\right)V(x).
  \end{align*}
   Under
   \[
    a^2\ge \frac{24dL}{C_F^2},
    \]
    and
    \[
    \gamma\ge \max\left(a^2\frac{4d{L'}^2}{{\lambda^\dagger}^2},4{L'}^2a^6\right),
    \]
  $\nu$ satisfies the Poincaré inequality with a constant $\frac{\lambda\ast}{35}$. Moreover, since $F$ is $L$-smooth, $\nabla^2 F\succeq-LI_{d\times d}$. Therefore, all the conditions of Theorem \ref{tha4} are satisfied.
  \par We conclude that from Theorem \ref{tha4}, $\nu$ satisfies the Log-Sobolev inequality with a constant $\alpha$ such that
  \[
  \frac{1}{\alpha}\le C_1+\left(C_2+2\right)\frac{35}{{\lambda^\dagger}},
  \]
  where constants $C_1$ and $C_2$ can be calculated as
  \[
  C_1=\frac{2M^2+8L^2}{M^2L\gamma},
  \]
  and
  \[
  C_2\le \frac{6L(d+\gamma)}{M}
  \]
  from \citet{RRT2017}.
  We can replace this value by a simple upper bound which gives us
  \[
  \frac{1}{\alpha}\le \left(\frac{2M^2+8L^2}{M^2L}+\left(\frac{6L(d+1))}{M}+2\right)\frac{35}{{\lambda^\dagger}}\right)\gamma
  \]
  since $\gamma\ge 1$.\\
\end{proof}
\begin{remarkA}
  Once we obtain the Poincaré constant, they are several ways to construct the Log-Sobolev constant. Another approach is possible, maybe simpler, by proceeding as \citet{LE2020} did in their analysis. Even though their method is interesting, this should not seriously change our main point since we just wanted to show that a polynomial dependence of the Log-Solev constant on the inverse temperature was achievable under certain additional conditions.
\end{remarkA}

\section{Analysis of an annealing scheme}\label{ape}

In this Appendix, we prove the global convergence of SVRG-LD and SARAH-LD combined with an annealed scheme.
\subsection{Algorithm}
In the context of optimization, we can use Algorithm \hyperlink{al1}{1} by setting a $\gamma$ huge enough so that the stationary distribution concentrates on the global minimizer of $F$. On the other hand, we can also introduce to SVRG-LD and SARAH-LD an increasing inverse temperature and a decreasing step size as follows.\\\ \\
\begin{algorithm}[H]\hypertarget{al3}{}
  input: batch size $B$, epoch length $m$\\
  annealing schedule: step size $\eta_s>0$ and inverse temperature $\gamma_s\ge 1$\\
  initialization: $X_0=0$, $ X^{(0)}=X_0$\\
  \ForEach{$s=0,1,\ldots,(K/m)$}
  {
  $v_{sm}=\nabla F(X^{(s)})$\\
  randomly draw $\epsilon_{sm} \sim N(0,I_{d\times d})$\\
  $X_{sm+1}=X_{sm}-\eta_s v_{sm}+\sqrt{2\eta_s/\gamma_s}\epsilon_{sm}$\\
  \ForEach{$l=1,\ldots,m-1$}
  {
  $k=sm+l$\\
  randomly pick a subset $I_k$ from $\{1,\ldots,n\}$ of size $|I_k|=B$\\
  randomly draw $\epsilon_k \sim N(0,I_{d\times d})$\\
  \uIf{SVRG-LD}{
  $v_k=\frac{1}{B}\sum_{i_k\in I_k}(\nabla f_{i_k}(X_k)-\nabla f_{i_k}(X^{(s)}))+v_{sm}$
  }
  \ElseIf{SARAH-LD}{
  $v_k=\frac{1}{B}\sum_{i_k\in I_k}\left(\nabla f_{i_k}(X_k)-\nabla f_{i_k}(X_{k-1})\right)+v_{k-1}$
  }
  $X_{k+1}=X_k-\eta_s v_k+\sqrt{2\eta_s/\gamma_s}\epsilon_k$\\
  }
  $X^{(s+1)}=X_{(s+1)m}$\\
  }
  \caption{SVRG-LD / SARAH-LD with annealing}
\end{algorithm}
\begin{definition}
  We define $\psi_k$ as the distribution of $X_k$ generated at the kth step of Algorithm \hyperlink{al3}{2}.
\end{definition}
\subsection{Preparation for the Proof}
Let us first establish some special notations to keep the proof clear and simple.
\begin{notationA}
  We define $\nu_{\gamma_k}$ as the stationary Gibbs distribution of SDE \eqref{eq21} when the inverse temperature parameter is set at $\gamma_k$, namely,
  \[
  \nu_{\gamma_k}\vcentcolon= \e^{-\gamma_k F}/Z_{\gamma_k},
  \]
  where $Z_{\gamma_k}$ is the normalizing constant, and $\alpha_k$ as the Log-Sobolev constant of $\nu_{\gamma_k}$ under Assumptions \ref{as1} and \ref{as3}. We also abbreviate the KL divergence between the distribution $\psi_{sm+r}$ of the random variable $X_{sm+r}$ generated by Algorithm 3 and the Gibbs distribution $\nu_{\gamma_s}$ as follows, where $s\in \mathbb{N}\cup\{0\}$ and $r=1,\ldots,m$:
  \[
   H_{sm+r}\vcentcolon =H_{\nu_{\gamma_s}}(\psi_{sm+r}).
  \]
  Moreover, $H_0\vcentcolon=H_{\nu_{\gamma_0}}(\psi_{0})$.
\end{notationA}
We will also need the following technical lemma.
\begin{lemmaA}\label{l8}
  For all $s\in \mathbb{N}\cup\{0\}$, $\sigma\ge 3$ and $\mu> 2$,
  \[
  \left(\frac{2}{3}\right)^{\frac{2}{\mu}}(s+1)^{1-\frac{2}{\mu}}\sigma^{-\frac{2}{\mu}}  \le \sum_{i=0}^s(i+\sigma)^{-\frac{2}{\mu}},
  \]
  where $C_\mu$ is a constant independent of $s$ and $\sigma$.
\end{lemmaA}
\begin{proof}
 By a simple argument of area under the curve $y=x^{-\frac{2}{\mu}}$,
\[
\sum_{i=0}^s(i+\sigma)^{-\frac{2}{\mu}}\ge \int_{\sigma}^{s+\sigma+1}x^{-\frac{2}{\mu}}\d x.
\]
According to the mean value theorem for integrals, there exist a constant $c_s\in [\sigma,s+\sigma+1]$ such that
\[
   \sum_{i=0}^s(i+\sigma)^{-\frac{2}{\mu}}\ge \int_{\sigma}^{s+\sigma+1}x^{-\frac{2}{\mu}}\d x=c_s^{-\frac{2}{\mu}}(s+1).
\]
We have also
\begin{align*}
  c_s^{-\frac{2}{\mu}}\ge &\ (s+1+\sigma)^{-\frac{2}{\mu}}\\
  =&\ (s+1)^{-\frac{2}{\mu}}(1+\frac{\sigma}{s+1})^{-\frac{2}{\mu}}\\
  \ge&\ (s+1)^{-\frac{2}{\mu}}(1+\sigma)^{-\frac{2}{\mu}}\\
  =&\  (s+1)^{-\frac{2}{\mu}}\sigma^{-\frac{2}{\mu}}\left(\frac{1+\sigma}{\sigma}\right)^{-\frac{2}{\mu}}\\
  \ge &\ (s+1)^{-\frac{2}{\mu}}\sigma^{-\frac{2}{\mu}}\left(\frac{2}{3}\right)^{\frac{2}{\mu}}.
\end{align*}
In the last inequality, we used $\sigma\ge 2$. This implies the inequality of the statement.\\
\end{proof}
Now, considering that we only change the step size and the inverse temperature parameter at the beginning of every inner loop, all statements proved in Appendix \ref{apa} (Lemmas \ref{l3} and \ref{l4}) and in Appendix \ref{apb} (Lemmas \ref{l5}, \ref{l6} and \ref{l7}) that consider only the inner loop hold for Algorithm \hyperlink{al3}{2} as well.
\par Moreover, let us consider the annealing schedule
\begin{align}
  \eta_s=&\ \bar{\eta}(s+\sigma)^{-\frac{1}{\mu}},\label{eq52} \\
  \gamma_s=&\ \bar{\gamma}\log\left\{g(s+\sigma)^\frac{1}{\mu}\right\},\label{eq53}
\end{align}
where we suppose $\bar{\eta}>0$, $\bar{\gamma}>0$, $\sigma\ge 3$, $g\ge \e$ and $\mu>2$. This annealing schedule is chosen on the one hand so that $\sum_{i=0}^s\frac{\alpha_i}{\gamma_i}\eta_i$ is explicitly computable, and on the other hand because \citet{CHS1987} showed that the annealed continuous time GLD
\[
\d X_t^{\footnotesize{\mathrm{Ann}}}=\nabla F(X_t^{\footnotesize{\mathrm{Ann}}})\d t+\sqrt{T(t)}\d B_t
\]
could find the global minimum with the annealing schedule $T(t)\propto \frac{1}{\log t}$, which corresponds to equation \eqref{eq53}.
\par Then, the following theorem holds under this annealing schedule.
\begin{theoremA}\label{th7}
  With the annealing schedule \eqref{eq52} and \eqref{eq53}, under Assumptions \ref{as1} and \ref{as3}, $0<\bar{\eta}<\frac{C_1}{16\sqrt{6}g L^2m}$,  $\bar{\gamma}=\frac{1}{C_2}$, $\mu>2$, $g\ge \e$, and $B\ge m$, for all $k=sm+r$ where $s\in \mathbb{N}\cup \{0\}$ and $r=0,\ldots,m-1$, the following holds in the update of Algorithm \hyperlink{al3}{2}:
  \begin{align*}
    H_{\nu_{\gamma_s}}(\psi_{sm+r+1})\le&\ \e^{-\frac{3\alpha_s}{2\gamma_s}\eta_s}\left(1+\frac{32\gamma_s L^4\eta_s^3}{\alpha_s}\right)H_{\nu_{\gamma_s}}(\psi_{sm+r})\\
    &+\e^{-\frac{3\alpha_s}{2\gamma_s}\eta_s}\sum_{i=0}^{r-1}\frac{128\gamma_s L^4\eta_s^3}{\alpha_s} \e^{-\frac{\alpha_s m}{\gamma_s}\eta_s} H_{\nu_{\gamma_s}}(\psi_{sm+i})\\
    &+56\eta_s^2 dL^2.
  \end{align*}
  Here, $C=\frac{(n-B)}{B(n-1)}$.
\end{theoremA}
\begin{proof}
  From Property \ref{p1}, $\nu_{\gamma_s}$ satisfies Log-Sobolev inequality with a constant $\alpha_s$ such that $\frac{\alpha_s}{\gamma_s}=C_1\e^{-C_{2} \gamma_s}$. It thus suffices to notice that under $0<\bar{\eta}<\frac{C_1}{16\sqrt{6}g L^2m}$ and $\bar{\gamma}=\frac{1}{C_2}$, we have for all $s\in\mathbb{N}\cup\{0\}$,
  \begin{align*}
    \eta_s=&\ \bar{\eta}(s+\sigma)^{-\frac{1}{\mu}}\\
    \le&\ \frac{C_1}{16\sqrt{6}g (s+\sigma)^{\frac{1}{\mu}}L^2m}\\
    =& \frac{\alpha_s}{16\sqrt{6}\gamma_s L^2m}.
  \end{align*}
  In the inequality, we used the fact that with $\bar{\gamma}=\frac{1}{C_2}$, \begin{align*}
    \frac{\alpha_s}{\gamma_s}=&\ C_1\e^{-C_2\gamma_s}\\
    =&\ C_1\e^{-C_2\bar{\gamma}\log\left\{g (s+\sigma)^{\frac{1}{\mu}}\right\}}\\
    =&\ \frac{C_1}{g(s+\sigma)^{\frac{1}{\mu}}}.
  \end{align*}
  Therefore, all the assumptions of Theorem \ref{th1} and \ref{th3} are satisfied. From the proof of each theorem, we immediately obtain the inequality of the statement from equations \eqref{eq37} and \eqref{eq45}.\\
\end{proof}
The problem with changing the inverse temperature parameter of each inner loop is that we cannot immediately give an upper bound for each $H_{k}$ as Theorem \ref{th2} and \ref{th4}. The main challenge resides in linking $H_{\nu_{\gamma_s}}(\psi_{sm})$ and $H_{\nu_{\gamma_{s-1}}}(\psi_{sm})$, which corresponds to the shift of optimization trajectory in the space of measures generated by the change of inverse temperature parameter at the beginning of each inner loop. The following lemma suggests that a small enough difference between two consecutive inverse temperatures will solve this issue.
\begin{lemmaA}\label{l9}
  Under Assumptions \ref{as1}, \ref{as3} and $F\ge 0$, for all $s\in \mathbb{N}$ and $\gamma_0\ge \frac{2}{M}$,
  \[
  H_{\nu_{\gamma_{s}}}(\psi_{sm})\le\left(1+\Delta\gamma_s\frac{2L}{\alpha_{s-1}}\right)H_{\nu_{\gamma_{s-1}}}(\psi_{sm})+\Delta\gamma_s\left(\chi+F(X^\ast)\right).
  \]
  Here, $\Delta\gamma_s\vcentcolon=\gamma_s-\gamma_{s-1}$, $\chi\vcentcolon=\max_{\gamma\ge 1}\left\{\frac{d}{\gamma}\log{\left(\frac{\e L}{M}\left(\frac{b\gamma}{d}+1\right)\right)}\right\}$ and $X^\ast$ is the global minimum of $F$.
\end{lemmaA}
\begin{proof}
 \begin{align*}
   H_{\nu_{\gamma_{s}}}(\psi_{sm})=&\ H_{\nu_{\gamma_{s-1}}}(\psi_{sm})+ H_{\nu_{\gamma_{s}}}(\psi_{sm})-H_{\nu_{\gamma_{s-1}}}(\psi_{sm})\\
   =&\ H_{\nu_{\gamma_{s-1}}}(\psi_{sm})+ \int \psi_{sm}\log\frac{\psi_{sm}}{\nu_{\gamma_{s}}}\d z -\int \psi_{sm}\log\frac{\psi_{sm}}{\nu_{\gamma_{s-1}}}\d z\\
   =&\ H_{\nu_{\gamma_{s-1}}}(\psi_{sm})+ \int \psi_{sm}\log\frac{\nu_{\gamma_{s-1}}}{\nu_{\gamma_{s}}}\d z\\
   =&\ H_{\nu_{\gamma_{s-1}}}(\psi_{sm})+ \int \psi_{sm}\log\frac{\e^{-\gamma_{s-1}F}/Z_{\gamma_{s-1}}}{\e^{-\gamma_{s}F}/Z_{\gamma_{s}}}\d z\\
   =&\ H_{\nu_{\gamma_{s-1}}}(\psi_{sm})+ \int \psi_{sm}(\gamma_{s}-\gamma_{s-1})F\d z+\log\frac{Z_{\gamma_{s}}}{Z_{\gamma_{s-1}}}.
 \end{align*}
 Here, as $\gamma_s\ge \gamma_{s-1}$and $F\ge 0$, we have that
 \[
 -\gamma_s F\le -\gamma_{s-1}F,
 \]
 which means
 \[
 Z_{\gamma_{s}}\le Z_{\gamma_{s-1}}.
 \]
 Thus,
 \[
 H_{\nu_{\gamma_{s}}}(\psi_{sm})\le H_{\nu_{\gamma_{s-1}}}(\psi_{sm})+\Delta\gamma_s \E_{X\sim \psi_{sm}}[F(X)].
 \]
 Now, from Corollary \ref{cor51} and Theorem \ref{th6}, we know that
 \begin{align*}
   \E_{X\sim \psi_{sm}}[F(X)]=&\  \E_{X\sim \psi_{sm}}[F(X)]-F(X^\ast)+F(X^\ast)\\
   \le &\ LW_2^2(\psi_k,\nu_{\gamma_{s-1}})+2\left(\E_{X\sim\nu_{\gamma_{s-1}} }[F(X)]-F(X^\ast)\right)+F(X^\ast)\\
   \le &\ LW_2^2(\psi_k,\nu_{\gamma_{s-1}})+\frac{d}{\gamma_{s-1}}\log{\left(\frac{\e L}{M}\left(\frac{b\gamma_{s-1}}{d}+1\right)\right)}+F(X^\ast)\\
   \le &\ \frac{2L}{\alpha_{s-1}}H_{\nu_{\gamma_{s-1}}}(\psi_{sm})+\chi+F(X^\ast).
 \end{align*}
 We used Corollary \ref{cor51} at the first inequality, Theorem \ref{th6} at the second inequality and Talagrand's inequality at the last inequality. This gives the desired result.\\
\end{proof}
 Since the logarithmic function $\log x$ is strictly increasing while its derivative decreases according to $x$, we can find an adequate bound of $\sigma$ to assure that $\Delta\gamma_s$ is small enough. As a reminder, we set $\gamma_s=\bar{\gamma}\log\left\{ g(s+\sigma)^\frac{1}{\mu}\right\}$ and $\eta_s=\bar{\eta}(s+\sigma)^\frac{1}{\mu}$.
 \begin{lemmaA}\label{l10}
   With the annealing \eqref{eq52} and \eqref{eq53}, when $\alpha_s=\gamma_sC_1\e^{-C_2\gamma_s}$,
   \[\sigma\ge 3 \vee\left(\frac{8Lg^2}{C_1^2\bar{\eta}}\right)^\frac{\mu}{\mu-3}\vee\left(\frac{2}{\mu C_2L^2\bar{\eta}^2}\right)^\frac{\mu}{\mu-2},
   \]
   $\bar{\gamma}=\frac{1}{C_2}$, $\mu> 3$ and $g\ge \e$, we have
   \begin{align}\label{eq54}
     \Delta\gamma_s\frac{2L}{\alpha_{s-1}}\le \frac{\alpha_s\eta_s}{2\gamma_s},
   \end{align}
   and
   \begin{align}\label{eq55}
     \Delta\gamma_s\le \eta_s^2L^2\le \frac{1}{4}
   \end{align}
   for all $s\in\mathbb{N}\cup\{0\}$.
 \end{lemmaA}
 \begin{proof}
   First of all, by the mean value theorem, there exists a $c\in [s-1,s]$ such that,
   \[
   \Delta\gamma_s=\frac{\bar{\gamma}/\mu}{c+\sigma}.
   \]
   Thus,
   \[
   \Delta\gamma_s=\frac{\bar{\gamma}/\mu}{c+\sigma}\le \frac{\bar{\gamma}/\mu}{s-1+\sigma}=\frac{1}{\mu C_2}\frac{1}{(s-1+\sigma)}.
   \]
   Therefore, in order to satisfy inequality \eqref{eq54}, it suffices to have
   \begin{align*}
     \frac{1}{\mu C_2}\frac{1}{(s-1+\sigma)}\le &\ \frac{\alpha_{s-1}}{2L}\frac{\alpha_s\eta_s}{2\gamma_s}\\
     =&\ \frac{\gamma_{s-1}C_1\e^{-C_2\gamma_{s-1}}}{2L}\frac{1}{2}C_1\e^{-C_2\gamma_s}\eta_s.
  \end{align*}
  A sufficient condition to this is
  \begin{align*}
    \frac{1}{\mu C_2}\frac{1}{(s-1+\sigma)}\le&\ \frac{C_1C_2^{-1}\log\left\{g(s-1+\sigma)^\frac{1}{\mu}\right\}}{2Lg(s+\sigma)^\frac{1}{\mu}}\frac{\bar{\eta}C_1}{2g(s+\sigma)^\frac{2}{\mu}},
  \end{align*}
   which gives
\begin{align*}
  \frac{4Lg^2}{C_1^2\bar{\eta}}\le&\ \frac{s+\sigma-1}{(s+\sigma)^\frac{3}{\mu}}\log\left\{g(s-1+\sigma)\right\}\\
  =&\ (s+\sigma)^{1-\frac{3}{\mu}}\frac{s+\sigma-1}{s+\sigma}\log\left\{g(s-1+\sigma)\right\}\\
  =&\ (s+\sigma)^{1-\frac{3}{\mu}}\left(1-\frac{1}{s+\sigma}\right)\log\left\{g(s-1+\sigma)\right\}.
\end{align*}
As $\log\left\{g(s-1+\sigma)\right\}\ge 1$ and $1-\frac{1}{s+\sigma}\ge \frac{1}{2}$ when $g\ge \e$, $s\ge 0$ and $\sigma\ge 2$, it suffices to have the following inequality satisfied when $s=0$:
\[
\frac{8L}{C_1^2\bar{\eta}}\le (s+\sigma)^{1-\frac{3}{\mu}}.
\]
From this, we obtain $\sigma\ge \left(\frac{8Lg^2}{C_1^2\bar{\eta}}\right)^\frac{\mu}{\mu-3}$.\par
 Likewise, in order to satisfy inequality \eqref{eq55}, it suffices to have
 \begin{align*}
   \frac{1}{\mu C_2}\frac{1}{(s-1+\sigma)}\le &\ \frac{\bar{\eta}^2L^2}{(s+\sigma)^\frac{2}{\mu}},
 \end{align*}
 which gives
 \[
 \frac{1}{\mu C_2L^2\bar{\eta}^2}\le \frac{s+\sigma-1}{(s+\sigma)^\frac{2}{\mu}}.
 \]
 It thus suffices to have the following inequality satisfied when $s=0$:
 \[
 \frac{2}{\mu C_2L^2\bar{\eta}^2}\le (s+\sigma)^{1-\frac{2}{\mu}}.
 \]
 This gives $\sigma\ge \left(\frac{2}{\mu C_2L^2\bar{\eta}^2}\right)^\frac{\mu}{\mu-2}$.\par
 The last inequality $\eta_s^2L^2\le \frac{1}{2}$ is immediately satisfied with $\eta_s\le \bar{\eta}\le \frac{1}{4L}$.\\
\end{proof}
\subsection{Main Proof}
We are now ready to prove the main results. We first evaluate how $H_k$ decreases compared with the previous step.
\begin{theoremA}\label{th8}
  With the annealing schedule \eqref{eq52} and \eqref{eq53}, under Assumptions \ref{as1}, \ref{as3} and $F\ge 0$, $0<\bar{\eta}<\frac{C_1}{16\sqrt{6}gL^2m}$, $\bar{\gamma}=\frac{1}{C_2}$, $B\ge m$, $\mu>3$, $g\ge \e$ and
   \[\sigma\ge 3 \vee\left(\frac{8Lg^2}{C_1^2\bar{\eta}}\right)^\frac{\mu}{\mu-3}\vee\left(\frac{2}{\mu C_2L^2\bar{\eta}^2}\right)^\frac{\mu}{\mu-2},
  \]
  for all $k=sm+r$ where $s\in \mathbb{N}\cup \{0\}$ and $r=0,\ldots,m-1$, the following holds in the update of Algorithm \hyperlink{al3}{2}:
  \begin{align*}
    H_{sm+r+1}\le&\ \e^{-\frac{\alpha_s}{\gamma_s}\eta_s}\left(1+\frac{\alpha_s}{4\gamma_s}\eta_s\right)H_{sm+r}\\
    &+\e^{-\frac{\alpha_s}{\gamma_s}\eta_s}\sum_{i=0}^{r-1}\frac{\alpha_s}{4m\gamma_s}\eta_s \e^{-\frac{\alpha_s m}{\gamma_s}\eta_s} H_{sm+i}\\
    &+\eta_s^2 dL^2E.
  \end{align*}
  Here, $E=56+2\chi+2F(X^\ast)$
\end{theoremA}
\begin{proof}
  When $r=0$, from Theorem \ref{th7}, we have
  \begin{align*}
    H_{\nu_{\gamma_s}}(\psi_{sm+1})\le&\ \e^{-\frac{3\alpha_s}{2\gamma_s}\eta_s}\left(1+\frac{32\gamma_s L^4\eta_s^3}{\alpha_s}\right)H_{\nu_{\gamma_s}}(\psi_{sm})\\
    &+56\eta_s^2 dL^2.
  \end{align*}
  Under
  \[\sigma\ge 3 \vee\left(\frac{8Lg^2}{C_1^2\bar{\eta}}\right)^\frac{\mu}{\mu-3}\vee\left(\frac{2}{\mu C_2L^2\bar{\eta}^2}\right)^\frac{\mu}{\mu-2},
 \]
 we can derive the following bound:
 \begin{align*}
   H_{\nu_{\gamma_s}}(\psi_{sm+1})\le&\ \e^{-\frac{3\alpha_s}{2\gamma_s}\eta_s}\left(1+\frac{32\gamma_s L^4\eta_s^3}{\alpha_s}\right)\left(1+\Delta\gamma_s\frac{2L}{\alpha_{s-1}}\right)H_{\nu_{\gamma_{s-1}}}(\psi_{sm})\\
   &+\e^{-\frac{3\alpha_s}{2\gamma_s}\eta_s}\left(1+\frac{32\gamma_s L^4\eta_s^3}{\alpha_s}\right)\Delta\gamma_s\left(\chi+F(X^\ast)\right)\\
   &+56\eta_s^2 dL^2\\
   \le&\ \e^{-\frac{3\alpha_s}{2\gamma_s}\eta_s}\left(1+\frac{32\gamma_s L^4\eta_s^3}{\alpha_s}\right)\left(1+\frac{\alpha_s}{2\gamma_s}\eta_s\right)H_{\nu_{\gamma_{s-1}}}(\psi_{sm})\\
   &+\left(1+2 L^2\eta_s^2\right)\eta_s^2L^2\left(\chi+F(X^\ast)\right)\\
   &+56\eta_s^2 dL^2\\
   \le&\ \e^{-\frac{3\alpha_s}{2\gamma_s}\eta_s}\left(1+\frac{32\gamma_s L^4\eta_s^3}{\alpha_s}\right)\e^{\frac{\alpha_s}{2\gamma_s}\eta_s}H_{\nu_{\gamma_{s-1}}}(\psi_{sm})\\
   &+\left(1+1\right)\eta_s^2L^2\left(\chi+F(X^\ast)\right)\\
   &+56\eta_s^2 dL^2\\
   \le&\ \e^{-\frac{\alpha_s}{\gamma_s}\eta_s}\left(1+\frac{32\gamma_s L^4\eta_s^3}{\alpha_s}\right)H_{\nu_{\gamma_{s-1}}}(\psi_{sm})\\
   &+\eta_s^2 dL^2\left(56+2\chi+2F(X^\ast)\right).
 \end{align*}
 We used Lemma \ref{l9} at the first inequality, Lemma \ref{l10} and $\eta_s\le \frac{\alpha_s}{16\gamma_sL^2m}$ at the second inequality and $\eta_s^2L^2\le \frac{1}{2}$ at the last inequality.\par
 When $r\ge 1$, from Theorem \ref{th7}, we have
 \begin{align*}
   H_{\nu_{\gamma_s}}(\psi_{sm+r+1})\le&\ \e^{-\frac{3\alpha_s}{2\gamma_s}\eta_s}\left(1+\frac{32\gamma_s L^4\eta_s^3}{\alpha_s}\right)H_{\nu_{\gamma_s}}(\psi_{sm+r})\\
   &+\e^{-\frac{3\alpha_s}{2\gamma_s}\eta_s}\sum_{i=0}^{r-1}\frac{128\gamma_s L^4\eta_s^3}{\alpha_s} \e^{-\frac{\alpha_s m}{\gamma_s}\eta_s} H_{\nu_{\gamma_s}}(\psi_{sm+i})\\
   &+56\eta_s^2 dL^2.
 \end{align*}
 Under
 \[\sigma\ge 3 \vee\left(\frac{8Lg^2}{C_1^2\bar{\eta}}\right)^\frac{\mu}{\mu-3}\vee\left(\frac{2}{\mu C_2L^2\bar{\eta}^2}\right)^\frac{\mu}{\mu-2},
\]
\begin{align*}
  H_{\nu_{\gamma_s}}(\psi_{sm+r+1})\le&\ \e^{-\frac{3\alpha_s}{2\gamma_s}\eta_s}\left(1+\frac{32\gamma_s L^4\eta_s^3}{\alpha_s}\right)H_{\nu_{\gamma_s}}(\psi_{sm+r})\\
  &+\e^{-\frac{3\alpha_s}{2\gamma_s}\eta_s}\sum_{i=1}^{r-1}\frac{128\gamma_s L^4\eta_s^3}{\alpha_s} \e^{-\frac{\alpha_s m}{\gamma_s}\eta_s} H_{\nu_{\gamma_s}}(\psi_{sm+i})\\
  &+\e^{-\frac{3\alpha_s}{2\gamma_s}\eta_s}\frac{128\gamma_s L^4\eta_s^3}{\alpha_s} \e^{-\frac{\alpha_s m}{\gamma_s}\eta_s} H_{\nu_{\gamma_s}}(\psi_{sm})\\
  &+56\eta_s^2 dL^2\\
  \le&\ \e^{-\frac{3\alpha_s}{2\gamma_s}\eta_s}\left(1+\frac{32\gamma_s L^4\eta_s^3}{\alpha_s}\right)H_{\nu_{\gamma_s}}(\psi_{sm+r})\\
  &+\e^{-\frac{3\alpha_s}{2\gamma_s}\eta_s}\sum_{i=1}^{r-1}\frac{128\gamma_s L^4\eta_s^3}{\alpha_s} \e^{-\frac{\alpha_s m}{\gamma_s}\eta_s} H_{\nu_{\gamma_s}}(\psi_{sm+i})\\
  &+\e^{-\frac{3\alpha_s}{2\gamma_s}\eta_s}\frac{128\gamma_s L^4\eta_s^3}{\alpha_s} \e^{-\frac{\alpha_s m}{\gamma_s}\eta_s} \left(1+\Delta\gamma_s\frac{2L}{\alpha_{s-1}}\right)H_{\nu_{\gamma_{s-1}}}(\psi_{sm})\\
  &+\e^{-\frac{3\alpha_s}{2\gamma_s}\eta_s}\frac{128\gamma_s L^4\eta_s^3}{\alpha_s}\e^{-\frac{\alpha_s m}{\gamma_s}\eta_s}\Delta\gamma_s\left(\chi+F(X^\ast)\right)\\
  &+56\eta_s^2 dL^2\\
  \le&\ \e^{-\frac{3\alpha_s}{2\gamma_s}\eta_s}\left(1+\frac{32\gamma_s L^4\eta_s^3}{\alpha_s}\right)H_{\nu_{\gamma_s}}(\psi_{sm+r})\\
  &+\e^{-\frac{3\alpha_s}{2\gamma_s}\eta_s}\sum_{i=1}^{r-1}\frac{128\gamma_s L^4\eta_s^3}{\alpha_s} \e^{-\frac{\alpha_s m}{\gamma_s}\eta_s} H_{\nu_{\gamma_s}}(\psi_{sm+i})\\
  &+\e^{-\frac{\alpha_s}{\gamma_s}\eta_s}\frac{128\gamma_s L^4\eta_s^3}{\alpha_s} \e^{-\frac{\alpha_s m}{\gamma_s}\eta_s} H_{\nu_{\gamma_{s-1}}}(\psi_{sm})\\
  &+2\eta_s^2L^2\Delta\gamma_s\left(\chi+F(X^\ast)\right)+56\eta_s^2 dL^2\\
  \le&\ \e^{-\frac{\alpha_s}{\gamma_s}\eta_s}\left(1+\frac{32\gamma_s L^4\eta_s^3}{\alpha_s}\right)H_{\nu_{\gamma_s}}(\psi_{sm+r})\\
  &+\e^{-\frac{\alpha_s}{\gamma_s}\eta_s}\sum_{i=0}^{r-1}\frac{128\gamma_s L^4\eta_s^3}{\alpha_s} \e^{-\frac{\alpha_s m}{\gamma_s}\eta_s} H_{sm+i}\\
  &+\eta_s^2dL^2\left(56+2\chi+2F(X^\ast)\right).
\end{align*}
 Therefore, for all $r=0,\ldots,m-1$,
 \begin{align*}
   H_{sm+r+1} \le&\ \e^{-\frac{\alpha_s}{\gamma_s}\eta_s}\left(1+\frac{32\gamma_s L^4\eta_s^3}{\alpha_s}\right)H_{sm+r}\\
  &+\e^{-\frac{\alpha_s}{\gamma_s}\eta_s}\sum_{i=0}^{r-1}\frac{128\gamma_s L^4\eta_s^3}{\alpha_s} \e^{-\frac{\alpha_s m}{\gamma_s}\eta_s} H_{sm+i}\\
  &+\eta_s^2dL^2\left(56+2\chi+2F(X^\ast)\right)\\
  \le&\ \e^{-\frac{\alpha_s}{\gamma_s}\eta_s}\left(1+\frac{\alpha_s }{4\gamma_s}\eta_s\right)H_{sm+r}\\
 &+\e^{-\frac{\alpha_s}{\gamma_s}\eta_s}\sum_{i=0}^{r-1}\frac{\alpha_s }{4m\gamma_s}\eta_s \e^{-\frac{\alpha_s m}{\gamma_s}\eta_s} H_{sm+i}\\
 &+\eta_s^2dL^2E.
 \end{align*}
 In the last inequality, we used $\eta_s\le \frac{\alpha_s}{16\sqrt{2}L^2m\gamma_s}$.\\
\end{proof}
\begin{theoremA}\label{th9}
  With the annealing schedule \eqref{eq52} and \eqref{eq53}, under Assumptions \ref{as1}, \ref{as3} and $F\ge 0$, $0<\bar{\eta}<\frac{C_1}{16\sqrt{6}gL^2m}$, $B\ge m$, $\mu> 3$, $\bar{\gamma}=\frac{1}{C_2}$, $g\ge \e$ and
   \[\sigma\ge 3 \vee\left(\frac{8Lg^2}{C_1^2\bar{\eta}}\right)^\frac{\mu}{\mu-3}\vee\left(\frac{2}{\mu C_2L^2\bar{\eta}^2}\right)^\frac{\mu}{\mu-2},
  \]
  for all $k=sm$ where $s\in \mathbb{N}$, the following holds in the update of Algorithm 3:
 \begin{align*}
   H_{k}\le&\ \e^{-\frac{C_1\bar{\eta}}{2g}\left(\frac{2}{3}\right)^{\frac{2}{\mu}}k^{1-\frac{2}{\mu}}m^{\frac{2}{\mu}}\sigma^{-\frac{2}{\mu}}}H_{0}+\frac{8}{3}\bar{\eta}dL^2EgC_1^{-1}k^\frac{2}{\mu}m^{-\frac{2}{\mu}}\sigma^\frac{2}{\mu},
 \end{align*}
 where $E=56+2\chi+2F(X^\ast)$.
\end{theoremA}
\begin{proof}
  First of all, we will prove by mathematical induction that in each inner loop the following inequality holds for all $r=0,\ldots,m-1$:
  \[
  H_{sm+r+1}\le \e^{-\frac{\alpha_s}{2\gamma_s}\eta_s(r+1)}H_{sm}+\eta_s^2dL^2E\left(\sum_{i=0}^r\e^{-\frac{\alpha_s}{2\gamma_s}\eta_si}\right)\ \ \ \ \ \ \ \ldots (\ast\ast)\hypertarget{astast}{}
  \]
  When $r=0$, from Theorem \ref{th8}, we have
  \begin{align*}
    H_{sm+1}\le&\ \e^{-\frac{\alpha_s}{\gamma_s}\eta_s}\left(1+\frac{\alpha_s}{4\gamma_s}\eta_s\right)H_{sm}+\eta_s^2 dL^2E\\
    \le&\ \e^{-\frac{\alpha_s}{\gamma_s}\eta_s}\left(1+\frac{\alpha_s}{2\gamma_s}\eta_s\right)H_{sm}+\eta_s^2 dL^2E\\
    \le&\ \e^{-\frac{\alpha_s}{\gamma_s}\eta_s}\e^{\frac{\alpha_s}{2\gamma_s}\eta_s}H_{sm}+\eta_s^2 dL^2E\\
    \le&\ \e^{-\frac{\alpha_s}{2\gamma_s}\eta_s}H_{sm}+\eta_s^2 dL^2E.
  \end{align*}
  Thus, \hyperlink{astast}{$(\ast\ast)$} holds for $r=0$.
  \par Now, let us suppose that \hyperlink{astast}{$(\ast\ast)$} is true for all $r\le l$. Then, when $r=l+1$ from Theorem \ref{th8}, we have
  \begin{align*}
    H_{sm+l+2}\le&\ \e^{-\frac{\alpha_s}{\gamma_s}\eta_s}\left(1+\frac{\alpha_s}{4\gamma_s}\eta_s\right)H_{sm+l+1}+\e^{-\frac{\alpha_s}{\gamma_s}\eta_s}\sum_{i=0}^{l}\frac{\alpha_s}{4m\gamma_s}\eta_s \e^{-\frac{\alpha_s m}{\gamma_s}\eta_s} H_{sm+i}\\
    &+\eta_s^2 dL^2E\\
    \le&\ \e^{-\frac{\alpha_s}{\gamma_s}\eta_s}\left(1+\frac{\alpha_s}{4\gamma_s}\eta_s\right)\left(\e^{-\frac{\alpha_s}{2\gamma_s}\eta_s(l+1)}H_{sm}+\eta_s^2dL^2E\left(\sum_{j=0}^l\e^{-\frac{\alpha_s}{2\gamma_s}\eta_sj}\right)\right)\\
    &+\e^{-\frac{\alpha_s}{\gamma_s}\eta_s}\sum_{i=0}^{l}\frac{\alpha_s}{4m\gamma_s}\eta_s \e^{-\frac{\alpha_s m}{\gamma_s}\eta_s} \left(\e^{-\frac{\alpha_s}{2\gamma_s}\eta_si}H_{sm}+\eta_s^2dL^2E\left(\sum_{j=0}^{i-1}\e^{-\frac{\alpha_s}{2\gamma_s}\eta_sj}\right)\right)\\
    &+\eta_s^2 dL^2E\\
    \le&\ \e^{-\frac{\alpha_s}{\gamma_s}\eta_s}\left(1+\frac{\alpha_s}{4\gamma_s}\eta_s\right)\left(\e^{-\frac{\alpha_s}{2\gamma_s}\eta_s(l+1)}H_{sm}+\eta_s^2dL^2E\left(\sum_{j=0}^l\e^{-\frac{\alpha_s}{2\gamma_s}\eta_sj}\right)\right)\\
    &+\e^{-\frac{\alpha_s}{\gamma_s}\eta_s}\sum_{i=0}^{l}\frac{\alpha_s}{4m\gamma_s}\eta_s \left(\e^{-\frac{\alpha_s\eta_s}{2\gamma_s}(l+1)}H_{sm}+\eta_s^2dL^2E\left(\sum_{j=0}^{l}\e^{-\frac{\alpha_s}{2\gamma_s}\eta_sj}\right)\right)\\
    &+\eta_s^2 dL^2E\\
    \le&\ \e^{-\frac{\alpha_s}{2\gamma_s}\eta_s(l+1)}\e^{-\frac{\alpha_s}{\gamma_s}\eta_s}\left(1+\frac{\alpha_s}{4\gamma_s}\eta_s+\sum_{i=0}^{l}\frac{\alpha_s}{4m\gamma_s}\eta_s\right)H_{sm}\\
    &+\eta_s^2 dL^2E\e^{-\frac{\alpha_s}{\gamma_s}\eta_s}\left(1+\frac{\alpha_s}{4\gamma_s}\eta_s+\sum_{i=0}^{l}\frac{\alpha_s}{4m\gamma_s}\eta_s\right)\left(\sum_{j=0}^{l}\e^{-\frac{\alpha_s}{2\gamma_s}\eta_sj}\right)\\
    &+\eta_s^2 dL^2E.
  \end{align*}
  This further implies,
  \begin{align*}
  H_{sm+l+2}\le&\ \e^{-\frac{\alpha_s}{2\gamma_s}\eta_s(l+1)}\e^{-\frac{\alpha_s}{\gamma_s}\eta_s}\e^{\frac{\alpha_s}{2\gamma_s}\eta_s}H_{sm}\\
  &+\eta_s^2 dL^2E\e^{-\frac{\alpha_s}{\gamma_s}\eta_s}\e^{\frac{\alpha_s}{2\gamma_s}\eta_s}\left(\sum_{j=0}^{l}\e^{-\frac{\alpha_s}{2\gamma_s}\eta_sj}\right)+\eta_s^2 dL^2E\\
  \le&\ \e^{-\frac{\alpha_s}{2\gamma_s}\eta_s(l+2)}H_{sm}+\eta_s^2dL^2E\left(\sum_{i=0}^{l+1}\e^{-\frac{\alpha_s}{2\gamma_s}\eta_si}\right).
\end{align*}
  Therefore, \hyperlink{astast}{$(\ast\ast)$} holds for all inner loop and $r=0,\ldots,m-1$.\par
  Especially, when $r=m-1$, we obtain
  \[
  H_{(s+1)m}\le \e^{-\frac{\alpha_s}{2\gamma_s}\eta_sm}H_{sm}+\eta_s^2dL^2E\left(\sum_{i=0}^{m-1}\e^{-\frac{\alpha_s}{2\gamma_s}\eta_si}\right).
  \]
  Consecutively using this inequality, we obtain
  \begin{align*}
    H_{(s+1)m}\le&\ \e^{-\frac{\alpha_s}{2\gamma_s}\eta_sm}H_{sm}+\eta_s^2dL^2E\left(\sum_{i=0}^{m-1}\e^{-\frac{\alpha_s}{2\gamma_s}\eta_si}\right)\\
    \le&\ \e^{-\frac{\alpha_s}{2\gamma_s}\eta_sm}\left(\e^{-\frac{\alpha_{s-1}}{2\gamma_{s-1}}\eta_{s-1}m}H_{(s-1)m}+\eta_{s-1}^2dL^2E\left(\sum_{i=0}^{m-1}\e^{-\frac{\alpha_{s-1}}{2\gamma_{s-1}}\eta_{s-1}i}\right)\right)\\
    &+\eta_s^2dL^2E\left(\sum_{i=0}^{m-1}\e^{-\frac{\alpha_s}{2\gamma_s}\eta_si}\right)\\
    =&\ \e^{-\frac{m}{2}\left(\frac{\alpha_s}{\gamma_s}\eta_s+\frac{\alpha_{s-1}}{\gamma_{s-1}}\eta_{s-1}\right)}H_{(s-1)m}\\
    &+dL^2E\left(\eta_s^2\sum_{i=0}^{m-1}\e^{-\frac{\alpha_s}{2\gamma_s}\eta_si}+\eta_{s-1}^2\e^{-\frac{\alpha_s}{2\gamma_s}\eta_sm}\sum_{i=0}^{m-1}\e^{-\frac{\alpha_{s-1}}{2\gamma_{s-1}}\eta_{s-1}i}\right)\\
    \ldots&\\
    \le &\ \e^{-\frac{m}{2}\sum_{i=0}^s\frac{\alpha_i}{\gamma_i}\eta_i}H_{0}\\
    &+dL^2E\sum_{i=0}^s\left(\eta_i^2\e^{-\frac{m}{2}\sum_{j=i+1}^s\frac{\alpha_j}{\gamma_j}\eta_j}\sum_{j=0}^{m-1}\e^{-\frac{\alpha_i}{2\gamma_i}\eta_ij}\right),
  \end{align*}
  which implies
  \begin{align*}
    H_{(s+1)m}\le&\ \e^{-\frac{m}{2}\sum_{i=0}^s\frac{\alpha_i}{\gamma_i}\eta_i}H_{0}+dL^2E\sum_{i=0}^s\left(\eta_i^2\e^{-\frac{m}{2}\sum_{j=i+1}^s\frac{\alpha_s}{\gamma_s}\eta_s}\sum_{j=0}^{m-1}\e^{-\frac{\alpha_s}{2\gamma_s}\eta_sj}\right)\\
    \le&\ \e^{-\frac{m}{2}\sum_{i=0}^s\frac{\alpha_i}{\gamma_i}\eta_i}H_{0}+\bar{\eta}^2dL^2E\sum_{i=0}^\infty\e^{-\frac{\alpha_s}{2\gamma_s}\eta_si}\\
    = &\ \e^{-\frac{m}{2}\sum_{i=0}^s\frac{\alpha_i}{\gamma_i}\eta_i}H_{0}+\bar{\eta}^2dL^2E\left(1-\e^{-\frac{\alpha_s}{2\gamma_s}\eta_s}\right)^{-1}\\
    \le &\ \e^{-\frac{m}{2}\sum_{i=0}^s\frac{\alpha_i}{\gamma_i}\eta_i}H_{0}+\bar{\eta}^2dL^2E \left(\frac{3}{4}\frac{\alpha_s}{2\gamma_s}\eta_s\right)^{-1}\\
    = &\ \e^{-\frac{mC_1\bar{\eta}}{2g}\sum_{i=0}^s(i+\sigma)^{-\frac{2}{\mu}}}H_{0}+\frac{8}{3}\bar{\eta}dL^2EgC_1^{-1}(s+\sigma)^\frac{2}{\mu}\\
    \le &\ \e^{-\frac{C1\bar{\eta}}{2g}\left(\frac{2}{3}\right)^\frac{2}{\mu}m(s+1)^{1-\frac{2}{\mu}}\sigma^{-\frac{2}{\mu}}}H_{0}+\frac{8}{3}\bar{\eta}dL^2EgC_1^{-1}(s+1)^\frac{2}{\mu}\sigma^\frac{2}{\mu}.
  \end{align*}
  For the first inequality, we used $\frac{\alpha_i}{\gamma_i}\eta_i\ge\frac{\alpha_s}{\gamma_s}\eta_s$ for all $i\le s$, for the third inequality, we used $1-\e^{-c}\ge \frac{3}{4}c$ holds for all $0< c= \frac{\alpha_s}{2\gamma_s}\eta_s\le \frac{1}{4}$, and for the last inequality, we used Lemma \ref{l8}.
  \par Setting $k=(s+1)m$, we obtain
  \begin{align*}
    H_{k}\le&\ \e^{-\frac{C_1\bar{\eta}}{2g}\left(\frac{2}{3}\right)^{\frac{2}{\mu}}k^{1-\frac{2}{\mu}}m^{\frac{2}{\mu}}\sigma^{-\frac{2}{\mu}}}H_{0}\\
    &+\frac{8}{3}\bar{\eta}dL^2EgC_1^{-1}k^\frac{2}{\mu}m^{-\frac{2}{\mu}}\sigma^\frac{2}{\mu}.
  \end{align*}
\end{proof}
 Finally, we obtain the following global convergence guarantee for Algorithm \hyperlink{al3}{2}.
\begin{theoremA}\label{th10}
  Using Algorithm \hyperlink{al3}{2} with the annealing schedule $\eta_s=\bar{\eta}(s+\sigma)^{-\frac{1}{\mu}}$ and $\gamma_s= \bar{\gamma}\log\left\{g(s+\sigma)^\frac{1}{\mu}\right\}$, under Assumptions \ref{as1}, \ref{as3} and $F\ge 0$, $0<\bar{\eta}<\frac{C_1}{16\sqrt{6}gL^2m}$, $B\ge m$, $\epsilon=O\left(\frac{LH_0}{C_1C_2^{-1}}\right)$, $\mu\ge 13$, $g= \e^{\frac{h(\epsilon)\vee 2M\vee \bar{\gamma}}{\bar{\gamma}}}$, $\bar{\gamma}=\frac{1}{C_2}$ and
  \[\sigma= 3 \vee\left(\frac{8Lg^2}{C_1^2\bar{\eta}}\right)^\frac{\mu}{\mu-3}\vee\left(\frac{2}{\mu C_2L^2\bar{\eta}^2}\right)^\frac{\mu}{\mu-2},
 \]
 where
 \[
 h(\epsilon)\vcentcolon= \frac{4d}{\epsilon}\log{\left(\frac{\e L}{M}\right)}\vee \frac{8bd}{\epsilon^2}\vee 1,
 \]
  if we take $B=m=\sqrt{n}$, the largest permissible step size according to the value of $\sigma$, the gradient complexity to reach a precision of
  \[
  \E_{X_k\sim\rho_k }[F(X_k)]-F(X^\ast)\le \epsilon
  \]
  is
  \[
  \tilde{O}\left(GC_1+GC_2+GC_3\right).
  \]
  where
  \begin{align*}
    GC_1=&\ ng^\frac{2\mu}{\mu-2}C_1^\frac{-2\mu}{\mu-2}L^\frac{2\mu}{\mu-2}+n^{\frac{1}{2}-\frac{5}{2(\mu-5)}}\epsilon^{-\frac{\mu}{\mu-5}}g^\frac{3\mu}{\mu-5}C_1^{-\frac{3\mu}{\mu-5}}C_2^{\frac{\mu}{\mu-5}}(dE)^\frac{\mu}{\mu-5}L^\frac{3\mu}{\mu-5},\\
    GC_2=&\ n^{\frac{1}{2}+\frac{\mu^2-3\mu+6}{2(\mu-2)(\mu-3)}}\left(\frac{gL}{C_1}\right)^\frac{2\mu^2}{(\mu-2)(\mu-3)}\\
    &+n^{\frac{1}{2}-{\frac{5(\mu-3)}{2(\mu^2-11\mu+15)}}}\epsilon^{-\frac{\mu(\mu-1)}{\mu^2-11\mu+15}}\left(\frac{gL}{C_1}\right)^{\frac{(3\mu^2-7\mu+6)\mu}{(\mu-3)(\mu^2-11\mu+15)}}(dE)^\frac{\mu(\mu-1)}{\mu^2-11\mu+15},\\
    GC_3=&\ n^{\frac{1}{2}+\frac{\mu^2+4}{2(\mu-2)^2}}\left(\frac{gL}{C_1}\right)^\frac{2\mu^2}{(\mu-2)^2}C_2^{-\frac{2\mu}{(\mu-2)^2}}\\
    &+n^{\frac{1}{2}-\frac{5(\mu-2)}{2(\mu^2-13\mu+10)}}\epsilon^{-\frac{\mu(\mu+2)}{\mu^2-13\mu+10}}\left(\frac{gL}{C_1}\right)^{\frac{(3\mu-4)\mu}{\mu^2-13\mu+10}}(dE)^\frac{\mu(\mu+2)}{\mu^2-13\mu+10}C_2^\frac{(\mu^2-12\mu-6)\mu}{(\mu^2-13\mu+10)(\mu-2)},\\
    E=&\ 56+2\max_{\gamma\ge1}\left(\frac{d}{\gamma}\log\left(\frac{\e L}{M}\left(\frac{b\gamma}{d}+1\right)\right)\right)+2F^\ast,
  \end{align*}
  and $C_1$ and $C_2$ are defined in Property \ref{p1}.
\end{theoremA}
\begin{proof}
  Let us take $k=(s+1)m$, where $s\in \mathbb{N}\cup\{0\}$. From Corollary \ref{cor51}, the sufficient condition for
  \[
  \E_{X_k\sim\psi_k }[F(X_k)]-F(X^\ast)\le \epsilon
  \]
  is $LW_2^2(\psi_k,\nu_{\gamma_s})\le \epsilon/2$ and $\E_{X\sim\nu_{\gamma_s} }[F(X)]-F(X^\ast)\le \epsilon/4$. From $g\ge \e^\frac{2M}{\bar{\gamma}}$, which implies $\gamma_s\ge \frac{2}{M}$, and from Corollary \ref{cor61}, the latter condition is satisfied when $\gamma_s\ge \frac{4d}{\epsilon}\log{\left(\frac{\e L}{M}\right)}\vee \frac{8bd}{\epsilon^2}\vee 1$. Let us define
  \[
  h(\epsilon)\vcentcolon= \frac{4d}{\epsilon}\log{\left(\frac{\e L}{M}\right)}\vee \frac{8bd}{\epsilon^2}\vee 1.
  \]
  Then, as $\gamma_s=\bar{\gamma}\log{\left\{g(s+\sigma)^\frac{1}{\mu}\right\}}$ and $s+\sigma\ge \e$, a sufficient condition is
  \[
  \bar{\gamma}\log{g}\ge h(\epsilon).
  \]
  This is satisfied with
  \[
  g= \e^{\frac{h(\epsilon)\vee 2M\vee\bar{\gamma}}{\bar{\gamma}}}\ge \e^{\frac{h(\epsilon)}{\bar{\gamma}}}.
 \]
  Moreover, concerning the former condition, from Talagrand's inequality
  \[
  W_2^2(\rho_k,\nu_{\gamma_s})\le \frac{2}{\alpha_s}H_\nu(\rho_k),
  \]
  it suffices to have
  \[
  H_{\nu_{\gamma_s}}(\rho_k)\le \frac{\alpha_s\epsilon}{4L}=\frac{\epsilon}{4L}\frac{C_1C_2^{-1}\log{g(s+\sigma)^\frac{1}{\mu}}}{g(s+\sigma)^\frac{1}{\mu}}.
  \]
  As $g\ge \e$, $s+\sigma\ge 1$ and $(s+\sigma)\le (s+1)\sigma$, we obtain a simpler sufficient condition which is
  \[
  H_{\nu_{\gamma_s}}(\rho_k)\le \frac{\epsilon C_1C_2^{-1}}{4Lk^\frac{1}{\mu}m^{\frac{-1}{\mu}}\sigma^\frac{1}{\mu}g}.
  \]
  Therefore, from Theorem \ref{th9}, it is enough to take $\bar{\eta}$ and $k$ such that
  \begin{equation}\label{eq56}
\frac{8}{3}\bar{\eta}dL^2EgC_1^{-1} k^{\frac{2}{\mu}}m^{-\frac{2}{\mu}}\sigma^{\frac{2}{\mu}}\le \frac{\epsilon C_1C_2^{-1}}{8Lk^\frac{1}{\mu}m^{\frac{-1}{\mu}}\sigma^\frac{1}{\mu}g},
  \end{equation}
  and
  \begin{equation}\label{eq57}
    k^{1-\frac{2}{\mu}}\ge 2gC_1^{-1}\left(\frac{3}{2}\right)^\frac{2}{\mu}m^{-\frac{2}{\mu}}\sigma^{\frac{2}{\mu}}\bar{\eta}^{-1}\log\left(\frac{8Lk^\frac{1}{\mu}m^{\frac{-1}{\mu}}\sigma^\frac{1}{\mu}H_0}{\epsilon C_1C_2^{-1}}\right).
  \end{equation}
Concerning the first inequality \eqref{eq56}, we obtain
\begin{equation}\label{eq58}
\bar{\eta}\sigma^{\frac{3}{\mu}}\le \frac{3C_1^2C_2^{-1}}{64dL^3Eg^2}\epsilon k^{-\frac{3}{\mu}}m^\frac{3}{\mu}.
\end{equation}
On the other hand, since $(s+1)^\frac{1}{\mu}\ge 1$ and $\sigma^\frac{1}{\mu}\ge 1$, as long as $\epsilon=O\left(\frac{LH_0}{C_1C_2^{-1}}\right)$, we can consider the following condition for the second inequality \eqref{eq57}:
\begin{equation}\label{eq59}
  k^{1-\frac{2}{\mu}}\ge\tilde\Theta\left(2gC_1^{-1}\left(\frac{3}{2}\right)^\frac{2}{\mu}m^{-\frac{2}{\mu}}\sigma^{\frac{2}{\mu}}\bar{\eta}^{-1}\right)=\tilde\Theta\left(gC_1^{-1}m^{-\frac{2}{\mu}}\sigma^{\frac{2}{\mu}}\bar{\eta}^{-1}\right).
\end{equation}
(I) When
\[
\sigma=3 \vee\left(\frac{8Lg^2}{C_1^2\bar{\eta}}\right)^\frac{\mu}{\mu-3}\vee\left(\frac{2}{\mu C_2L^2\bar{\eta}^2}\right)^\frac{\mu}{\mu-2}=3,
\]
 equation \eqref{eq58} becomes
\[
\bar{\eta}\le \left(\frac{3C_1^2C_2^{-1}3^\frac{-3}{\mu}}{64dL^3E g^2}\right)\epsilon k^{-\frac{3}{\mu}}m^\frac{3}{\mu}.
\]
On the other hand, by plugging $\sigma=3$ to \eqref{eq59}, we obtain
\begin{align*}
  k^{1-\frac{2}{\mu}}\ge&\  \tilde\Theta\left(gC_1^{-1}m^{-\frac{2}{\mu}}\sigma^{\frac{2}{\mu}}\bar{\eta}^{-1}\right)\\
  \ge&\ \tilde\Theta\left(gC_1^{-1}m^{-\frac{2}{\mu}}\bar{\eta}^{-1}\right).
\end{align*}
If inequality \eqref{eq58} is stronger than $0<\bar{\eta}<\frac{C_1}{16\sqrt{6}L^2m}$, then
\[
k^{1-\frac{2}{\mu}}\ge \tilde\Theta\left(\frac{g^3dEL^3}{C_1^3C_2^{-1}}k^\frac{3}{\mu}m^{-\frac{5}{\mu}}\epsilon^{-1}\right).
\]
This leads to
\[
k\ge\tilde\Theta\left(\left(\frac{g^3dEL^3}{C_1^3C_2^{-1}}\right)^\frac{\mu}{\mu-5}m^{-\frac{5}{\mu-5}}\epsilon^{-\frac{\mu}{\mu-5}}\right).
\]
From this, if we take the largest permissible step size and the smallest permissible $\sigma$, the gradient complexity can be calculated with an optimal order when $B=m=n^\frac{1}{2}$ as
\begin{align*}
  \tilde{\Theta}\left(kB+\frac{k}{m}n\right)=&\ \tilde\Theta(k\sqrt{n})\\
  =&\tilde\Theta\left(n^{\frac{1}{2}-\frac{5}{2(\mu-5)}}\epsilon^{-\frac{\mu}{\mu-5}}g^\frac{3\mu}{\mu-5}C_1^{-\frac{3\mu}{\mu-5}}C_2^{\frac{\mu}{\mu-5}}(dE)^\frac{\mu}{\mu-5}L^\frac{3\mu}{\mu-5}\right).
\end{align*}
If inequality \eqref{eq58} is weaker than $0<\bar{\eta}<\frac{C_1}{16\sqrt{6}gL^2m}$, then
\[
k^{1-\frac{2}{\mu}}\ge\tilde\Theta\left(g^2C_1^{-2}L^
{2}m^{1-\frac{2}{\mu}}\right).
\]
This leads to
\[
k\ge\tilde\Theta\left(\left(g^2C_1^{-2}L^
{2}\right)^\frac{\mu}{\mu-2}m\right).
\]
From this, if we take the largest permissible step size and the smallest permissible $\sigma$, the gradient complexity can be calculated with an optimal order when $B=m=n^\frac{1}{2}$ as
\begin{align*}
  \tilde\Theta\left(kB+\frac{k}{m}n\right)=&\ \tilde\Theta(k\sqrt{n})\\
  =&\ \tilde\Theta\left(ng^\frac{2\mu}{\mu-2}C_1^\frac{-2\mu}{\mu-2}L^\frac{2\mu}{\mu-2}\right).
\end{align*}
Therefore, we obtain the following gradient complexity for this case:
\begin{equation}\label{eq510}
  \tilde{\Theta}\left(ng^\frac{2\mu}{\mu-2}C_1^\frac{-2\mu}{\mu-2}L^\frac{2\mu}{\mu-2}+n^{\frac{1}{2}-\frac{5}{2(\mu-5)}}\epsilon^{-\frac{\mu}{\mu-5}}g^\frac{3\mu}{\mu-5}C_1^{-\frac{3\mu}{\mu-5}}C_2^{\frac{\mu}{\mu-5}}(dE)^\frac{\mu}{\mu-5}L^\frac{3\mu}{\mu-5}\right).
\end{equation}
(II) When
\[
\sigma=3 \vee\left(\frac{8Lg^2}{C_1^2\bar{\eta}}\right)^\frac{\mu}{\mu-3}\vee\left(\frac{2}{\mu C_2L^2\bar{\eta}^2}\right)^\frac{\mu}{\mu-2}=\left(\frac{8Lg^2}{C_1^2\bar{\eta}}\right)^\frac{\mu}{\mu-3},
\]
 equation \eqref{eq58} becomes
\[
\bar{\eta}\le \left(\frac{3C_1^2C_2^{-1}}{64dL^3Eg^2}\right)^\frac{\mu-3}{\mu-6}\left(\frac{8Lg^2}{C_1^2}\right)^\frac{-3}{\mu-6}\epsilon^{\frac{\mu-3}{\mu-6}}k^{-\frac{3(\mu-3)}{\mu(\mu-6)}}m^\frac{3(\mu-3)}{\mu(\mu-6)}.
\]
On the other hand, by plugging $\sigma= \left(\frac{8Lg^2}{C_1^2\bar{\eta}}\right)^\frac{\mu}{\mu-3}$ to \eqref{eq59}, we obtain
\begin{align*}
  k^{1-\frac{2}{\mu}}\ge&\ \tilde\Theta\left(gC_1^{-1}m^{-\frac{2}{\mu}}\sigma^{\frac{2}{\mu}}\bar{\eta}^{-1}\right)\\
  \ge&\  \tilde\Theta\left(g^\frac{\mu+1}{\mu-3}C_1^{-\frac{\mu+1}{\mu-3}}L^\frac{2}{\mu-3}m^{-\frac{2}{\mu}}\bar{\eta}^{-\frac{\mu-1}{\mu-3}}\right).
\end{align*}
If inequality \eqref{eq58} is stronger than $0<\bar{\eta}<\frac{C_1}{16\sqrt{6}gL^2m}$, then
\[
k^{1-\frac{2}{\mu}}\ge\tilde\Theta\left(\left(\frac{C_1^2C_2^{-1}}{dEL^3g^2}\right)^{-\frac{\mu-1}{\mu-6}}\left(\frac{Lg^2}{C_1^2}\right)^{\frac{3(\mu-1)}{(\mu-6)(\mu-3)}}\frac{g^\frac{\mu+1}{\mu-3}L^\frac{2}{\mu-3}}{C_1^{\frac{\mu+1}{\mu-3}}}\epsilon^{-\frac{\mu-1}{\mu-6}}k^{\frac{3(\mu-1)}{\mu(\mu-6)}}m^{-\frac{5(\mu-3)}{\mu(\mu-6)}}\right).
\]
This leads to
\[
k\ge\tilde\Theta\left(\frac{\left(\left(\frac{C_1^2C_2^{-1}}{dEL^3g^2}\right)^{-\frac{\mu-1}{\mu-6}}\left(\frac{Lg^2}{C_1^2}\right)^{\frac{3(\mu-1)}{(\mu-6)(\mu-3)}}g^\frac{\mu+1}{\mu-3}C_1^{-\frac{\mu+1}{\mu-3}}L^\frac{2}{\mu-3}\right)^{\frac{\mu(\mu-6)}{\mu^2-11\mu+15}}}{m^{\frac{5(\mu-3)}{\mu^2-11\mu+15}}\epsilon^{\frac{\mu(\mu-1)}{\mu^2-11\mu+15}}}\right).
\]
From this, if we take the largest permissible step size and the smallest permissible $\sigma$, the gradient complexity can be calculated with an optimal order when $B=m=n^\frac{1}{2}$ as
\begin{align*}
  \tilde\Theta\left(kB+\frac{k}{m}n\right)=&\ \tilde\Theta(k\sqrt{n})\\
  =&\ \tilde\Theta\left(n^{\frac{1}{2}-{\frac{5(\mu-3)}{2(\mu^2-11\mu+15)}}}\epsilon^{-\frac{\mu(\mu-1)}{\mu^2-11\mu+15}}\left(\frac{gL}{C_1}\right)^{\frac{(3\mu^2-7\mu+6)\mu}{(\mu-3)(\mu^2-11\mu+15)}}(dE)^\frac{\mu(\mu-1)}{\mu^2-11\mu+15}\right).
\end{align*}
If inequality \eqref{eq58} is weaker than $0<\bar{\eta}<\frac{C_1}{16\sqrt{6}gL^2m}$, then
\[
k^{1-\frac{2}{\mu}}\ge \tilde\Theta\left(\left(\frac{gL}{C_1}\right)^\frac{2\mu}{\mu-3}m^{\frac{\mu^2-3\mu+6}{\mu(\mu-3)}}\right).
\]
This leads to
\[
k\ge\tilde\Theta\left(\left(\frac{gL}{C_1}\right)^\frac{2\mu^2}{(\mu-2)(\mu-3)}m^{\frac{\mu^2-3\mu+6}{(\mu-2)(\mu-3)}}\right).
\]
From this, if we take the largest permissible step size and the smallest permissible $\sigma$, the gradient complexity can be calculated with $B=m=n^\frac{1}{2}$ as
\begin{align*}
  \tilde\Theta\left(kB+\frac{k}{m}n\right)=&\ \tilde\Theta(k\sqrt{n})\\
  =&\ \tilde\Theta\left(n^{\frac{1}{2}+\frac{\mu^2-3\mu+6}{2(\mu-2)(\mu-3)}}\left(\frac{gL}{C_1}\right)^\frac{2\mu^2}{(\mu-2)(\mu-3)}\right).
\end{align*}
Therefore, we obtain the following gradient complexity for this case:
\begin{equation}\label{eq512}
  \tilde{\Theta}\left(n^{\frac{2\mu^2-8\mu+12}{2(\mu-2)(\mu-3)}}\left(\frac{gL}{C_1}\right)^\frac{2\mu^2}{(\mu-2)(\mu-3)}+\frac{n^{\frac{1}{2}-{\frac{5(\mu-3)}{2(\mu^2-11\mu+15)}}}}{\epsilon^{\frac{\mu(\mu-1)}{\mu^2-11\mu+15}}}\left(\frac{gL}{C_1}\right)^{\frac{(3\mu^2-7\mu+6)\mu}{(\mu-3)(\mu^2-11\mu+15)}}(dE)^\frac{\mu(\mu-1)}{\mu^2-11\mu+15}\right).
\end{equation}
(III) When
\[
\sigma=3 \vee\left(\frac{8Lg^2}{C_1^2\bar{\eta}}\right)^\frac{\mu}{\mu-3}\vee\left(\frac{2}{\mu C_2L^2\bar{\eta}^2}\right)^\frac{\mu}{\mu-2}=\left(\frac{2}{\mu C_2L^2\bar{\eta}^2}\right)^\frac{\mu}{\mu-2},
\]
 equation \eqref{eq58} becomes
\[
\bar{\eta}\le \left(\frac{3C_1^2C_2^{-1}}{64dL^3Eg^2}\right)^\frac{\mu-2}{\mu-8}\left(\frac{2}{\mu C_2L^2}\right)^\frac{-3}{\mu-8}\epsilon^\frac{\mu-2}{\mu-8}k^{-\frac{3(\mu-2)}{\mu(\mu-8)}}m^{\frac{3(\mu-2)}{\mu(\mu-8)}}.
\]
On the other hand, by plugging $\sigma= \left(\frac{2}{\mu C_2L^2\bar{\eta}^2}\right)^\frac{\mu}{\mu-2}$ to \eqref{eq59}, we obtain
\begin{align*}
  k^{1-\frac{2}{\mu}}\ge&\  \tilde\Theta\left(gC_1^{-1}m^{-\frac{2}{\mu}}\sigma^{\frac{2}{\mu}}\bar{\eta}^{-1}\right)\\
  \ge&\ \tilde\Theta\left(gC_1^{-1}(C_2L^2)^{-\frac{2}{\mu-2}}m^{-\frac{2}{\mu}}\bar{\eta}^{-\frac{\mu+2}{\mu-2}}\right).
\end{align*}
If inequality \eqref{eq58} is stronger than $0<\bar{\eta}<\frac{C_1}{16\sqrt{6}gL^2m}$, then
\[
k^{1-\frac{2}{\mu}}\ge\tilde\Theta\left(gC_1^{-1}(C_2L^2)^\frac{-2}{\mu-2}\left(\frac{C_1^2C_2^{-1}}{dEL^3g^2}\right)^{-\frac{\mu+2}{\mu-8}}(C_2L^2)^\frac{-3(\mu+2)}{(\mu-8)(\mu-2)}k^\frac{3(\mu+2)}{\mu(\mu-8)}m^{-\frac{5(\mu-2)}{\mu(\mu-8)}}\epsilon^\frac{\mu+2}{\mu-8}\right).
\]
This leads to
\[
k\ge\tilde\Theta\left(\frac{\left(gC_1^{-1}(C_2L^2)^\frac{-2}{\mu-2}\left(\frac{C_1^2C_2^{-1}}{dEL^3g^2}\right)^{-\frac{\mu+2}{\mu-8}}(C_2L^2)^\frac{-3(\mu+2)}{(\mu-8)(\mu-2)}\right)^{\frac{\mu(\mu-8)}{\mu^2-13\mu+10}}}{m^{\frac{5(\mu-2)}{\mu^2-13\mu+10}}\epsilon^{\frac{\mu(\mu+2)}{\mu^2-13\mu+10}}}\right).
\]
From this, if we take the largest permissible step size and the smallest permissible $\sigma$, the gradient complexity can be calculated with an optimal order when $B=m=n^\frac{1}{2}$ as
\begin{align*}
  \tilde\Theta\left(kB+\frac{k}{m}n\right)=&\ \tilde\Theta(k\sqrt{n})\\
  =&\ \tilde\Theta\left(\frac{n^{\frac{1}{2}-\frac{5(\mu-2)}{2(\mu^2-13\mu+10)}}}{\epsilon^{\frac{\mu(\mu+2)}{\mu^2-13\mu+10}}}\left(\frac{gL}{C_1}\right)^{\frac{(3\mu-4)\mu}{\mu^2-13\mu+10}}(dE)^\frac{\mu(\mu+2)}{\mu^2-13\mu+10}C_2^\frac{(\mu^2-12\mu-6)\mu}{(\mu^2-13\mu+10)(\mu-2)}\right).
\end{align*}
If inequality \eqref{eq58} is weaker than $0<\bar{\eta}<\frac{C_1}{16\sqrt{6}gL^2m}$, then
\[
k^{1-\frac{2}{\mu}}\ge\tilde\Theta\left(gC_1^{-1}(C_2L^2)^{-\frac{2}{\mu-2}}\left(\frac{gL^2}{C_1}\right)^\frac{\mu+2}{\mu-2}m^{\frac{\mu^2+4}{\mu(\mu-2)}}\right).
\]
This leads to
\[
k\ge\tilde\Theta\left(\left(gC_1^{-1}(C_2L^2)^{-\frac{2}{\mu-2}}\left(\frac{gL^2}{C_1}\right)^\frac{\mu+2}{\mu-2}\right)^{\frac{\mu}{\mu-2}}m^{\frac{\mu^2+4}{(\mu-2)^2}}\right).
\]
From this, if we take the largest permissible step size and the smallest permissible $\sigma$, the gradient complexity can be calculated with $B=m=n^\frac{1}{2}$ as
\begin{align*}
  \tilde\Theta\left(kB+\frac{k}{m}n\right)=&\ \tilde\Theta(k\sqrt{n})\\
  =&\ \tilde\Theta\left(n^{\frac{1}{2}+\frac{\mu^2+4}{2(\mu-2)^2}}\left(\frac{gL}{C_1}\right)^\frac{2\mu^2}{(\mu-2)^2}C_2^{-\frac{2\mu}{(\mu-2)^2}}\right).
\end{align*}
Therefore, we obtain the following gradient complexity for this case:
\begin{equation}\label{eq513}
  \scriptstyle{\tilde{\Theta}\left(n^{\frac{\mu^2-2\mu+4}{(\mu-2)^2}}\left(\frac{gL}{C_1C_2^\frac{1}{\mu}}\right)^\frac{2\mu^2}{(\mu-2)^2}+\frac{n^{\frac{\mu^2-18\mu+20}{2(\mu^2-13\mu+10)}}}{\epsilon^{\frac{\mu(\mu+2)}{\mu^2-13\mu+10}}}\left(\frac{gL}{C_1}\right)^{\frac{(3\mu-4)\mu}{\mu^2-13\mu+10}}(dE)^\frac{\mu(\mu+2)}{\mu^2-13\mu+10}C_2^\frac{(\mu^2-12\mu-6)\mu}{(\mu^2-13\mu+10)(\mu-2)}\right).}
\end{equation}
The statement of the theorem is obtained by grouping \eqref{eq510}, \eqref{eq512} and \eqref{eq513}.\\
\end{proof}
Now, looking at the gradient complexity of Theorem \ref{th10}, we remark that the dependence on $n$ is slightly improved for finite values of $\mu$ compared with $\mu\to\infty$. However, taking into account that $g=\e^{C_2h(\epsilon)}$ most of the time, the dependence on $\epsilon$ becomes worse as the exponent of $g$ of the first term is greater than 2, and that of the second term is greater than 3 for all $GC_i\ (i=1,2,3)$. Since this influence cannot be ignored in this case, we conclude that the best value of $\mu$ is $\mu=\infty$ in our analysis, i.e., the method that keeps $\eta$ and $\gamma$ constants.

\end{document}